%% file: main.tex
\documentclass[final]{cvpr}

\usepackage{times}
\usepackage{epsfig}
\usepackage{graphicx}
\usepackage{amsmath}
\usepackage{amssymb}
\usepackage{comment}
\usepackage{amsmath}
\usepackage{tabularx}
\usepackage{graphbox}
\usepackage{caption}

\usepackage{pifont}
\usepackage{adjustbox}

\include{math_commands}

\usepackage[pagebackref=true,breaklinks=true,colorlinks,bookmarks=false]{hyperref}

\newcommand{\ours}{\textsc{DatasetGAN}}

\def\cvprPaperID{6508} 
\def\confYear{CVPR 2021}

\begin{document}

\title{\vspace{-5.5mm}DatasetGAN: Efficient Labeled Data Factory with Minimal Human Effort}

\author{Yuxuan Zhang$^{1,5}$\thanks{authors contributed equally} \quad Huan Ling$^{1,2,3, \ast}$  \quad Jun Gao$^{1,2,3}$  \quad Kangxue Yin $^{1}$  \\
	\quad \quad Jean-Francois Lafleche$^{1}$   \quad 	Adela Barriuso$^{4}$   \quad 	Antonio Torralba$^{4}$   \quad Sanja Fidler$^{1,2,3}$ \\
	\quad \small{NVIDIA\textsuperscript{1} \quad University of Toronto\textsuperscript{2} \quad Vector Institute\textsuperscript{3} \quad MIT\textsuperscript{4} \quad University of Waterloo\textsuperscript{5}} \vspace{3pt}\\
	\quad \quad \texttt{\scriptsize y2536zha@uwaterloo.ca, \{huling, jung, kangxuey, jlafleche\}@nvidia.com}   \\
\texttt{\scriptsize  adela.barriuso@gmail.com, torralba@mit.edu, sfidler@nvidia.com}
	}


\twocolumn[{
\renewcommand\twocolumn[1][]{#1}
\maketitle
\begin{center}
\vspace{-9mm}
    \centering
  	\captionsetup{type=figure}
	\hspace{-2mm}\includegraphics[width=0.83\linewidth, trim=50 0 49 0,clip]{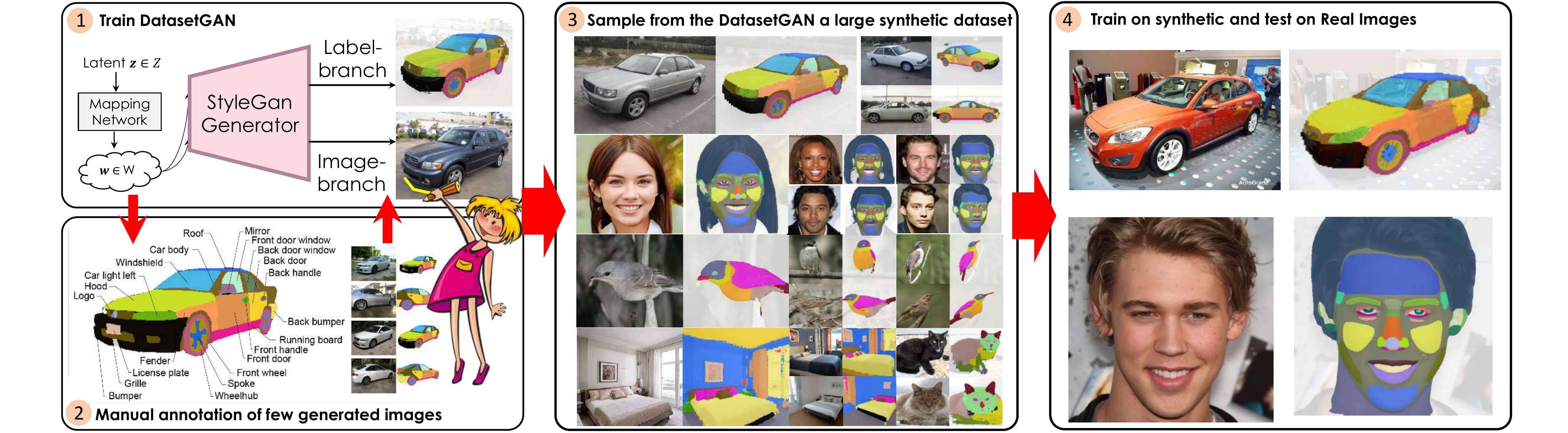} \\[-3.5mm]

	\captionof{figure}{\footnotesize {\bf {\ours}} synthesizes image-annotation pairs, and can produce large high-quality datasets with detailed pixel-wise labels. Figure illustrates the 4 steps. \emph{(1 \& 2).} Leverage StyleGAN and annotate only a handful of synthesized images. Train a highly effective branch to generate labels.   \emph{(3).} Generate a huge synthetic dataset of annotated images authomatically. \emph{(4).} Train your favorite approach with the synthetic dataset and test on real images.  }
	\label{fig:teaser}

\end{center}
}]


\maketitle
\pagestyle{empty}  
\thispagestyle{empty} 

\begin{abstract}

We introduce \emph{DatasetGAN}: an automatic procedure to generate massive datasets of high-quality semantically segmented images requiring minimal human effort. Current deep networks are extremely data-hungry, benefiting from training on large-scale datasets, which are time consuming to annotate. Our method relies on the power of recent GANs to generate realistic images. We show how the GAN latent code can be decoded to produce a semantic segmentation of the image. Training the decoder only needs a few labeled examples to generalize to the rest of the latent space, resulting in an infinite annotated dataset \emph{generator}! These generated datasets can then be used for training any computer vision architecture just as real datasets are. As only a few images need to be manually segmented, it becomes possible to annotate images in extreme detail and generate datasets with rich object and part segmentations. To showcase the power of our approach, we generated datasets for 7 image segmentation tasks which include pixel-level labels for 34 human face parts, and 32 car parts. Our approach outperforms all semi-supervised baselines significantly and is on par with fully supervised methods, which in some cases require as much as 100x more annotated data as our method.

\end{abstract}

\input{doc/intro.tex}
\input{doc/2_related.tex}

\input{doc/3_method.tex}

\input{doc/3a_annotation.tex}

\input{doc/4_exp.tex}

\input{doc/conc.tex}

{\small
\bibliographystyle{plain}
\bibliography{egbib}
}

\end{document}

%% file: math_commands.tex
\usepackage{amsmath,amsfonts,bm}

\def\eqref#1{equation~\ref{#1}}

\def\1{\bm{1}}

\DeclareMathAlphabet{\mathsfit}{\encodingdefault}{\sfdefault}{m}{sl}
\SetMathAlphabet{\mathsfit}{bold}{\encodingdefault}{\sfdefault}{bx}{n}



%% file: doc/intro.tex
\vspace{-4mm}
\section{Introduction}
\vspace{-2mm}


Curating image datasets with pixel-wise labels such as semantic or instance segmentation is very laborious (and expensive).  
Labeling a complex scene with 50 objects can take anywhere between 30 to 90 minutes -- clearly a bottleneck in achieving the scale of a dataset that we might desire. 
In this paper, we aim to synthesize large high quality labeled datasets by needing to label only a handful of examples.

Semi-supervised learning has been a popular approach in the quest of reducing the need for labeled data, by leveraging an additional large unlabeled dataset. The dominant approach trains a model on a labeled dataset using ground truth annotations while utilizing pseudo-labels~\cite{berthelot2019mixmatch,sohn2020fixmatch} and consistency regularization~\cite{sohn2020fixmatch,tarvainen2017mean} on the unlabeled examples. While most methods were showcased on classification tasks, recent work also showed success for the task of  semantic segmentation~\cite{mittal2019semi}.  On the other hand, contrastive learning aims to train feature extractors using contrastive (unsupervised) losses on sampled image pairs~\cite{oord2018representation, tian2019contrastive,chen2020big,misra2020self}, or image patches~\cite{ji2019invariant}. Once a  powerful image representation is trained using unsupervised losses alone, only a small subset of labeled images is typically required to train accurate predictors. In our work, we show that the latest state-of-the-art generative models of images learn extremely powerful latent representations that can be leveraged for complex pixel-wise tasks. 

We introduce \emph{DatasetGAN} which generates massive datasets of high-quality semantically segmented images requiring minimal human effort.  
Key to our approach is an observation that GANs trained to synthesize images must acquire rich semantic  knowledge in their ability to render diverse and realistic examples of objects. 
We exploit the feature space of a trained GAN and train a shallow decoder to produce a pixel-level labeling. Our key insight is that only a few labeled images are needed to train a successful decoder, leading to an infinite annotated dataset \emph{generator}. These generated datasets can then be used for training any computer vision architecture just as real datasets are. Since we only need to label a few examples, 
we annotate images in extreme detail and generate datasets with rich object and part segmentations. We generated datasets for 7 image segmentation tasks which include pixel-level labels for 34 human face parts, and 32 car parts. Our approach outperforms all semi-supervised baselines significantly and is on par with fully supervised methods, while in some cases requiring two orders of magnitude less annotated data. 

The ability of training successful computer vision models with as little as 16 labeled examples opens the door to exciting downstream applications. In our work, we showcase 3D reconstruction of animatable objects where we exploit the detailed part labels our method produces. 



%% file: doc/2_related.tex
\vspace{-2mm}
\section{Related Work}
\vspace{-1mm}


\vspace{-1mm}
\paragraph{Generative Models of Labeled Data:} Prior work on dataset synthesis has mainly focused on generative models of 3D scene graphs, utilizing graphics to render images and their labels~\cite{Metasim19,metasim20,fedsim20}. In our work, we focus on   
Generative Adversarial Networks (GANs)~\cite{goodfellow2014generative, stylegan, karras2017progressive, brock2018large} which synthesize high-quality images after training on a large dataset using adversarial objectives. 
Previous work utilized GANs to create synthetic datasets. 
In domain adaptation~\cite{murez2018image,zou2018unsupervised, vu2019advent, choi2019self}, several works aimed at translating a  labeled image dataset into another domain, in which image annotation is either expensive or missing entirely, by leveraging image-to-image translation techniques. 
A supervised computer vision model can then be trained on the translated dataset. These methods assume the existence of a large labeled  domain that can be leveraged for the new domain. In our work, we require only a handful of human-annotated images, and synthesize a much larger set. 
%

Recently,~\cite{zhang2020image} used StyleGAN~\cite{stylegan} as a multi-view image dataset generator for training an inverse graphics network to predict 3D shapes. The authors exploited the disentanglement between viewpoint and object identity in the StyleGAN's latent code. We go one step further and synthesize accurate semantic labels, by leveraging only a few human-provided examples. 
For the purpose of zero-shot image classification, GANs have also been used for synthesizing visual features of unseen classes from their semantic features~\cite{bucher2017generating, long2017zero, felix2018multi, sariyildiz2019gradient}.
To the best of our knowledge, {ours} is the first work in using GANs to directly synthesize a large dataset of images annotated to a high level of detail.

\vspace{-3mm}
\paragraph{Semi-Supervised Learning:}
%
Given a large set of unlabeled images and a small set of annotated images, semi-supervised approaches~\cite{luc2016semantic,souly2017semi, hung2018adversarial,mittal2019semi, ke2020guided} aim to learn  better  segmentation networks than with supervised data alone. Most of these methods treat the  segmentation network as a generator,  and train it adversarially with the small set of real annotations. In their case, the adversarial losses try to learn good segmentations from fake ones produced by the model, but they do not exploit generative modelling of  images themselves, as we do in our work. Pseudo-labels~\cite{berthelot2019mixmatch,sohn2020fixmatch} and consistency regularization~\cite{sohn2020fixmatch,tarvainen2017mean} have also recently been explored for semantic segmentation~\cite{mittal2019semi}, where the key ideas involve training on the small labeled dataset,  and re-training the model using a mix of real labeled data and highly confident predictions on unlabeled images. 
%
%
Different than existing semi-supervised methods, we utilize a GAN to synthesize both images and their pixel-wise labels. 

Concurrent work by~\cite{galeev2020learning} also translates GAN features  into semantic segmentation. However, their method relies on a decoder built with convolutional and residual blocks for projecting the internal layers of StyleGAN into a segmentation map. Our method directly interprets the disentangled feature vector for each pixel into its semantic label by a simple ensemble of MLP classifiers, which better utilizes the semantic knowledge in the StyleGAN's feature vectors. Furthermore, we use our approach to create large datasets of images annotated with high-detailed part labels and keypoints, which we hope will enable a wide variety of downstream applications not possible previously. 

In parallel work~\cite{segGAN21}, the authors explore an alternative direction in which the GAN, equipped with a segmentation branch, is also used as a semantic decoder at test time. A related idea was explored in~\cite{amodalVAE20}, where a VAE was used to decode amodal instance masks from partially visible masks. An encoder maps an image into a latent code using test-time optimization, which is then used to predict both the reconstructed image as well as semantic outputs. The semantic GAN is trained differently than ours, using adversarial losses. This method requires more training data than ours and is slower at test time, however, it has the appealing property of out of domain generalization. 

\vspace{-5.5mm}
\paragraph{Contrastive Learning:}
Contrastive methods learn a representation space for images with a contrastive loss for measuring similarity of sampled pairs~\cite{hadsell2006dimensionality}. Recent work on contrastive learning has shown promising results for image classification~\cite{oord2018representation,ji2019invariant, tian2019contrastive,bachman2019learning,he2020momentum,chen2020simple, chen2020big,grill2020bootstrap,misra2020self}.
With the learned self-supervised representation, the classification accuracy can be significantly improved by  fine-tuning on a small amount of labeled examples. 
Contrastive learning can also be applied to image segmentation by learning on pairs of image patches~\cite{ji2019invariant}.
Like ours, this line of work uses learned image representations to amortize the need for large labeled datasets. However, instead of using contrastive losses, we leverage the semantic knowledge in GAN's feature maps for fine-grained annotation synthesis.

%% file: doc/3_method.tex

\vspace{-1mm}
\section{Our Approach} 
\vspace{-2mm}
\label{sec:approach}

We now introduce {\ours} which synthesizes image-annotation pairs. We primarily focus on pixel-wise annotation tasks such as semantic segmentation and keypoint prediction, since they are typical examples of the most time consuming manual annotation tasks. 

The key insight of {\ours} is that generative models such as GANs that are trained to synthesize highly realistic images must acquire semantic knowledge in their high dimensional latent space. 
For example, the latent code in architectures like StyleGAN contains disentangled dimensions that control 3D properties such as viewpoint and object identity~\cite{stylegan, zhang2020image}.  
Interpolating between two latent codes have been shown to yield realistic generations~\cite{stylegan}, indicating that the GAN has also learned to semantically and geometrically align objects and their parts. {\ours}  aims to utilize these powerful properties of image GANs. Intuitively, if a human provides a labeling corresponding to one latent code, we expect to be able to effectively propagate this labeling across the GAN's latent space.

\begin{figure}[t!]
	\begin{tabular}{ccccc}
	\end{tabular}
	\begin{tabular}{c}
		\hspace{-2mm}\includegraphics[width=0.99\linewidth, trim=0 0 0 0,clip]{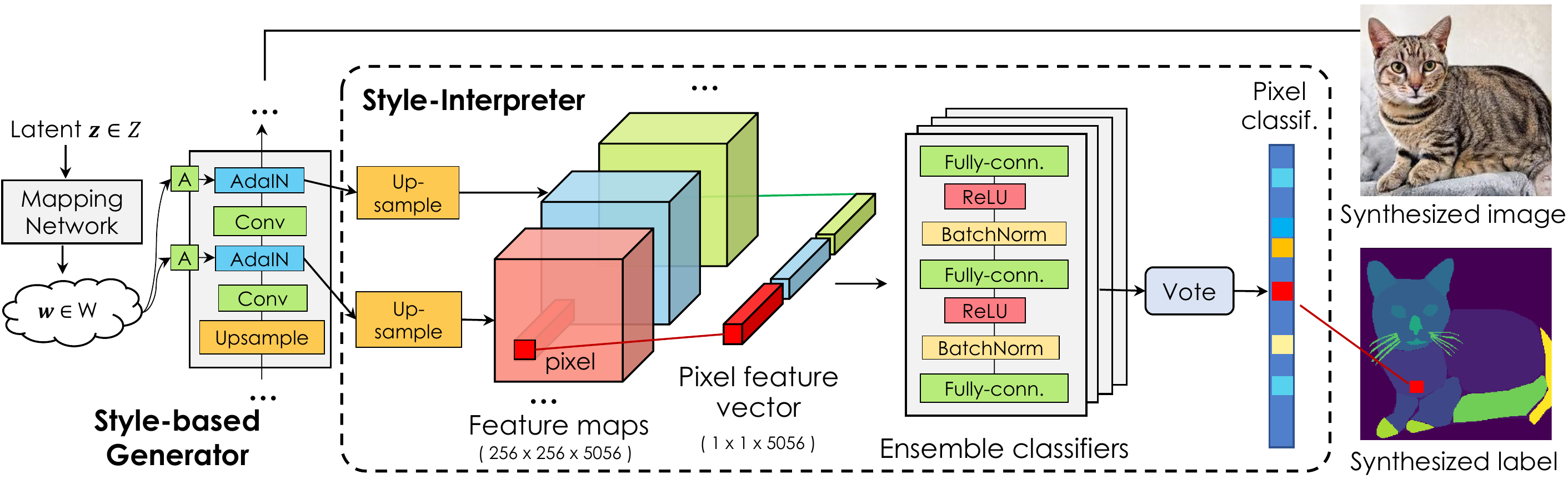} \\[-4mm]
	\end{tabular}
	\caption{\footnotesize Overall architecture of our {\ours}. We upsample the feature maps from StyleGAN to the highest resolution for constructing pixel-wise feature vectors for all pixels on the synthesized image. An ensemble of MLP classifiers is then trained for interpreting the semantic knowledge in the feature vector of a pixel into its part label.}
	\label{fig:model}
	\vspace{-4.5mm}
\end{figure}

Our {\ours}  is extremely simple, while extremely powerful. Specifically, we synthesize a small number of images by utilizing a GAN architecture, StyleGAN in our paper, and record their corresponding latent feature maps. A human annotator is asked to label these images with a desired set of labels. We then train a simple ensemble of MLP classifiers on top of the StyleGAN's pixel-wise feature vectors, which we refer to as the \emph{Style Interpreter}, to match the target human-provided labeling. Figure ~\ref{fig:model} provides a visualization. We observe that training the Style Interpreter requires only a few annotated examples for achieving good accuracy. 
When the Style Interpreter is trained, we use it as a label-generating branch in the StyleGAN architecture. By sampling latent codes $z$ and passing each through the entire architecture, we have an infinite dataset generator! 
These datasets can then be used for training any computer vision architecture just as real datasets are. 

We take advantage of the effectiveness of {\ours} in requiring only a few human-labeled images,  and devote efforts in annotating each individual image to a very high-detail pixel-wise labeling.  We create tiny datasets of up to 40 images containing extremely detailed part and keypoint annotations for a few classes, and utilize our {\ours} to synthesize much larger datasets. 
We believe that the community will find these datasets useful for a variety of exciting downstream applications. 

We briefly summarize StyleGAN in Sec~\ref{sec:stylegan}, and describe Style Interpreter in Sec~\ref{sec:interpreter}. We discuss dataset generation in Sec~\ref{sec:factory}, and detail our annotation efforts in Sec~\ref{sec:annotations}.


\vspace{-2mm}
\subsection{Prerequisites} 
\label{sec:stylegan}
\vspace{-1mm}

{\ours} uses StyleGAN as the generative backbone due to its impressive synthesis quality. 
The StyleGAN generator maps a latent code $z \in Z$ drawn from a normal distribution to a realistic image. Latent code $z$  is first mapped to an intermediate latent code $w \in W$ by a mapping function. $w$ is then transformed to $k$ vectors, $w^1,..., w^k$, through $k$ learned affine transformations. 
%
These $k$ transformed latent codes are injected as style information into $k/2$ synthesis blocks in a progressive fashion~\cite{karras2017progressive}. 
Specifically, each synthesis block consists of an upsampling ($\times2$) layer and two convolutional layers. Each convolutional layer is followed by an adaptive instance normalization (AdaIN) layer~\cite{huang2017arbitrary} which is controlled by its corresponding  $w^i$, a transformed latent code. We denote the output feature maps from the $k$ AdaIN layers as $\{S^0, S^1.., S^k\}$.

\begin{figure*}[t!]
\centering

\begin{minipage}{0.7\linewidth}
\begin{tabular}{c}
\hspace{-2mm}\includegraphics[width=.9\linewidth, height=5.5cm,trim=0 0 0 0,clip]{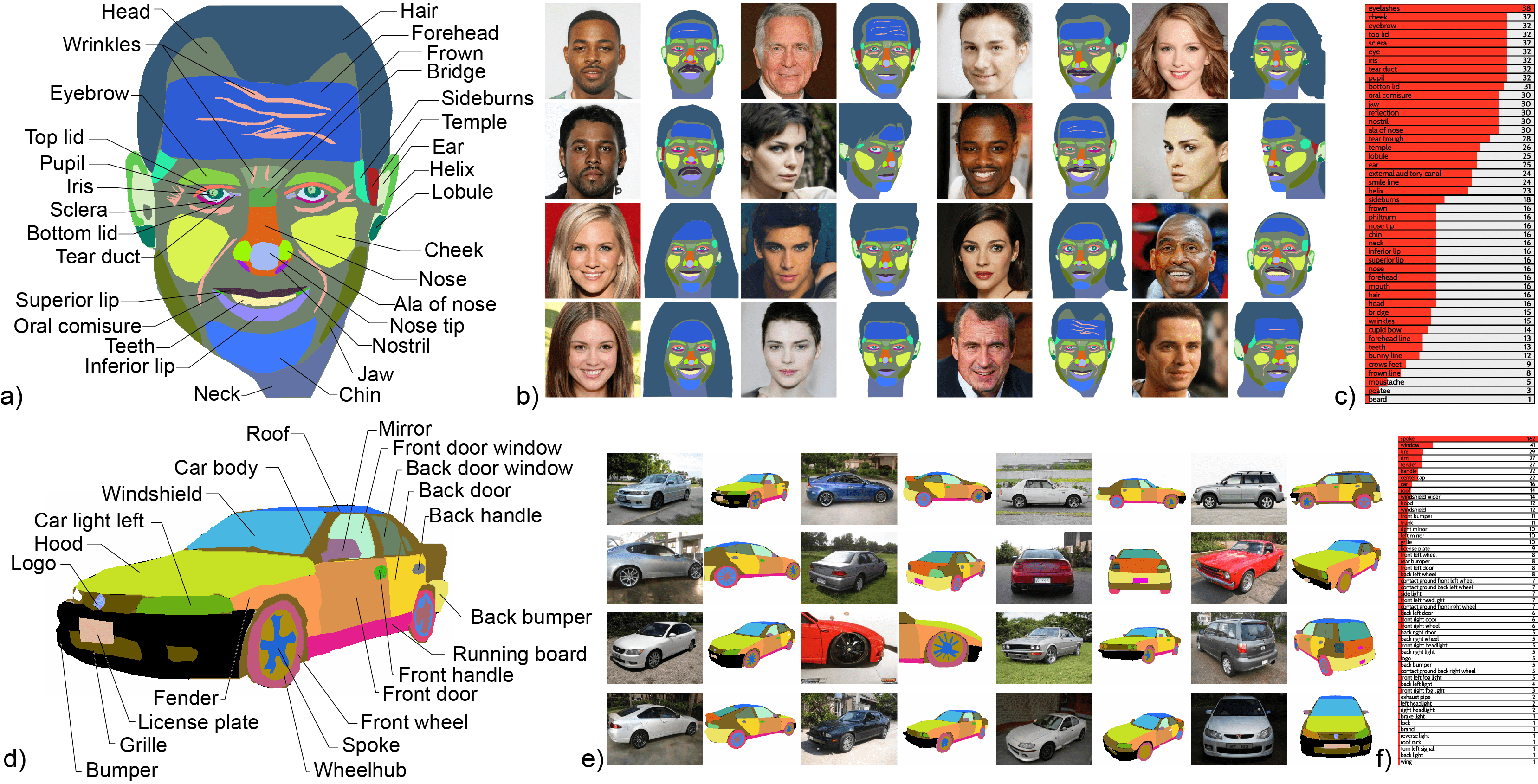} \\[-0.3mm]
\end{tabular}
\end{minipage}
\hspace{-5mm}
\begin{minipage}{0.28\linewidth}
\vspace{-3mm}
\caption{\footnotesize {\bf Small human-annotated face and car datasets. } Most datasets for semantic segmentation (MS-COCO~\cite{lin2014microsoft}, ADE~\cite{zhou2017scene}, Cityscapes~\cite{cordts2016cityscapes}) are too large for a user to be able to check every single training image. In this figure, we show all labeled training examples for face (a-c) and car (d-f) segmentation. a) shows an example of segmentation mask and associated labels, b) shows the full collection of training images (GAN samples), and c) shows the list of annotated parts and the number of instances in the dataset. As a fun fact, note that there are more labels in a single image than there are images in the dataset.}
\label{fig:dataset}
\vspace{-3mm}
\end{minipage}
\vspace{-5mm}
\end{figure*}

\vspace{-2mm}
\subsection{Style Interpreter} 
\label{sec:interpreter}
\vspace{-1mm}
We interpret StyleGAN as a ``rendering'' engine, and its latent codes as ``graphics" attributes that define what to render. 
We thus hypothesize that a flattened array of features that output a particular RGB pixel contains semantically meaningful information for rendering the pixel realistically. 
To this end, we upsample all feature maps $\{S^0, S^1.., S^k\}$ from AdaIN layers to the highest output resolution (resolution of $S^k$), and concatenate them to get a 3D feature tensor $S^* = (S^{0,*}, S^{1,*}.., S^{k,*})$.  Each pixel $i$ in the output image has its own feature vector $S^*_{i} = (S^{0,*}_{i}, S^{1,*}_{i}.., S^{k,*}_{i} )$, as shown in Figure~\ref{fig:model}. 
%
We use a three-layer MLP classifier on top of each feature vector to predict labels. We share weights across all pixels for simplicity. 

\vspace{-4mm}
\paragraph{Training:} We discuss annotation collection in Sec~\ref{sec:annotations}. Note that our goal here is to train the feature classifier -- the corresponding synthesized image is only used to collect annotations from a human labeler. 

Since feature vectors $S^*_{i}$ are of high dimensionality (5056), and the feature map has a high spatial resolution (1024 \time 1024 at most), we cannot easily consume all image feature vectors in a batch. We thus perform random sampling of feature vectors from each image, whereby we ensure that we sample at least once from each labeled region. We utilize a different loss for different tasks we consider. For semantic segmentation,  we train the classifier with cross-entropy loss.  For keypoint prediction, we build a Gaussian heatmap for each keypoint in the training set, and use the MLP functions to fit the heat value for each pixel. We do not backpropagate gradients to the StyleGAN backbone. 

To amortize the effect of random sampling, we train an ensemble of $N$ classifiers, $N=10$ in our paper. We use majority voting in each pixel at test time for semantic segmentation. 
For keypoint prediction, we average the $N$ heat values  predicted by each of  the $N$ classifiers for each pixel. 

\begin{figure*}
\vspace{-2mm}
\begin{center}
\includegraphics[width=0.44\linewidth, trim=20 0 15 13,clip]{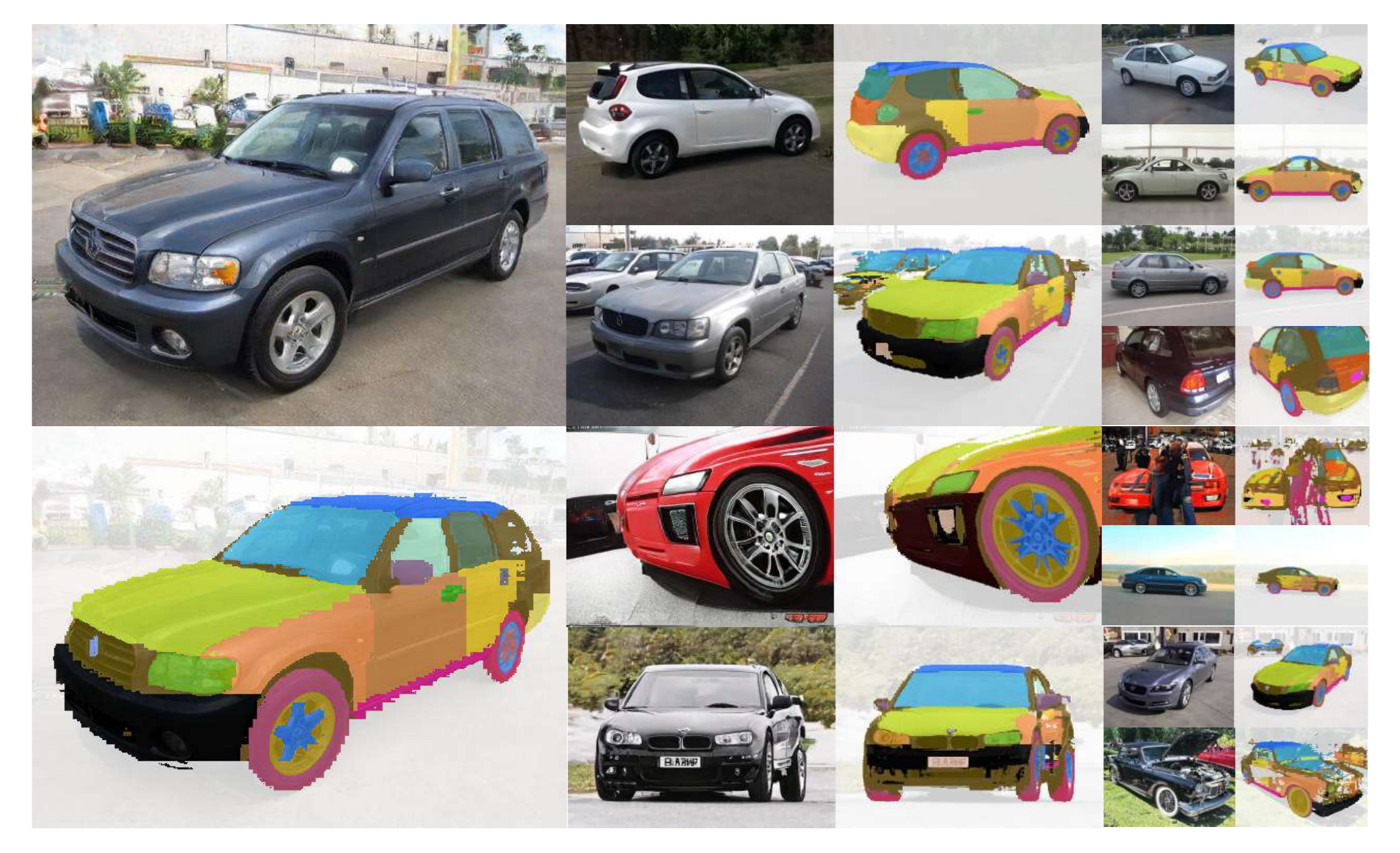}
\includegraphics[width=0.44\linewidth, trim=20 0 25 10,clip]{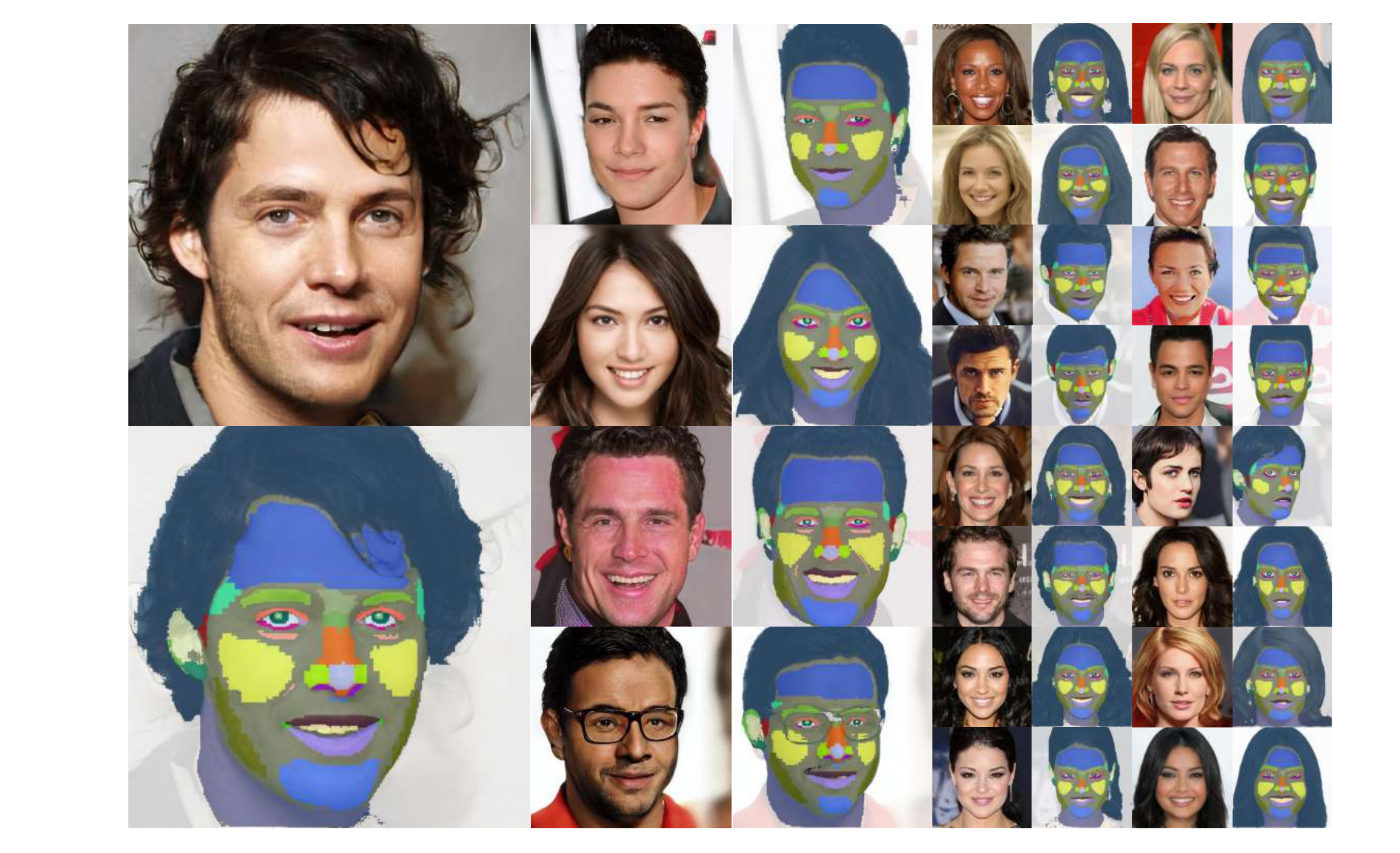}
\end{center}
\vspace{-8mm}
\caption{\footnotesize  Examples of synthesized images and labels from our {\ours} for faces and cars. StyleGAN backbone was trained on CelebA-HQ  (faces) on $1024 \times 1024$ resolution images, and  on LSUN CAR (cars) on $512\times384$ resolution images. {\ours} was trained on 16 annotated examples. }
\label{fig:vis_testing_cars}
\vspace{-2mm}
\end{figure*}

Our feature classifiers require remarkably few annotated images to make accurate predictions, shown in  Fig~\ref{fig:vis_testing_cars} and~\ref{fig:vis_testing2}, and validated in Experiments. 

\vspace{-2mm}
\subsection{DatasetGAN as a Labeled Data Factory} 
\label{sec:factory}
\vspace{-2mm}

Once trained, our Style Interpreter is used as a label-synthesis branch on top of the StyleGAN backbone, forming our {\ours}. We can therefore generate any desired number of  image-annotation pairs, which forms our synthetic dataset. 
Synthesizing an image-annotation pair requires a forward pass through StyleGAN, which takes 9s on average. While our experiments show that downstream performance keeps slightly increasing with every 10k of synthesized images, there is an associated cost and we use a dataset of 10k in size for most experiments. 

Naturally, StyleGAN also fails occasionally which introduces noise in the synthesized dataset.  We noticed that the StyleGAN's discriminator score is not a robust measure of failure and we found that utilizing our ensemble of classifiers to measure the uncertainty of a synthesized example is a more robust approach. 
We follow~\cite{Kuo2018CostSensitiveAL}, and use the Jensen-Shannon (JS) divergence~\cite{8579074,JSAL} as the uncertainty measure for a pixel. To calculate image uncertainty, we sum over all image pixels.
We filter out the top $10\%$ most uncertain images. We provide details in the Appendix. 

Random samples from five of our synthesized datasets for part segmentation of various object classes are shown in Fig~\ref{fig:vis_testing_cars} and Fig~\ref{fig:vis_testing2}. 
While not perfect (\eg, missing wrinkles), the quality of the synthesized labels is remarkable. 
Crowdsourcing labels on the same scale (10k images) for one  dataset would take over 3200 hours (134 days), and, we hypothesize, would be extremely noisy since annotating an image to that level of detail requires both skill and immense patience. In our case, human-annotation time for a dataset was roughly 5 hours, affording us to leverage a single skilled annotator. This is described next. 
\vspace{-1.5mm}

%% file: doc/3a_annotation.tex
\begin{figure*}
\vspace{-2mm}
\begin{center}
\addtolength{\tabcolsep}{2.3pt}
\begin{tabular}{ccc}
\includegraphics[height=3.8cm, trim=40 20 55 0,clip]{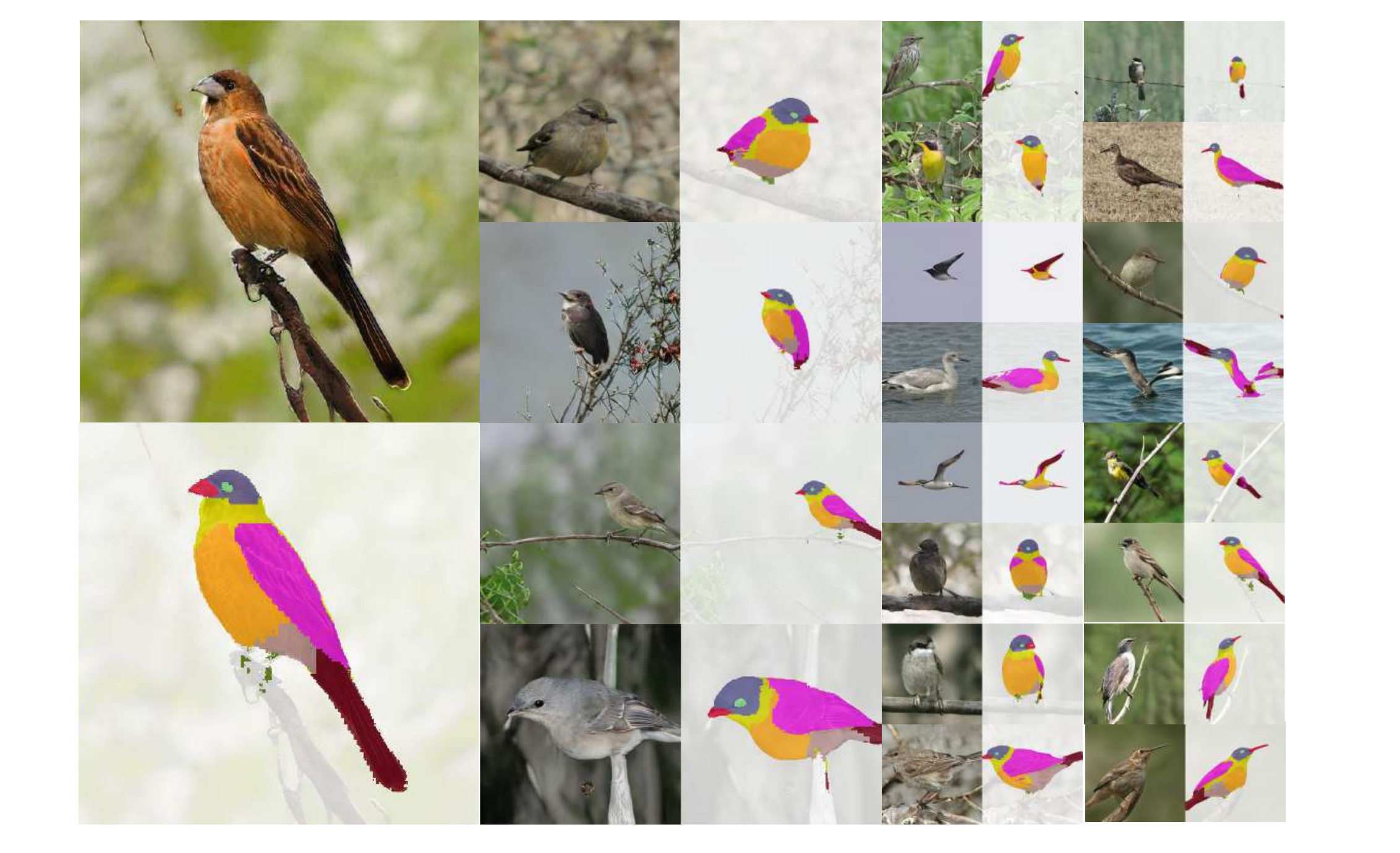} &
\includegraphics[height=3.8cm, trim=0 10 330 0,clip]{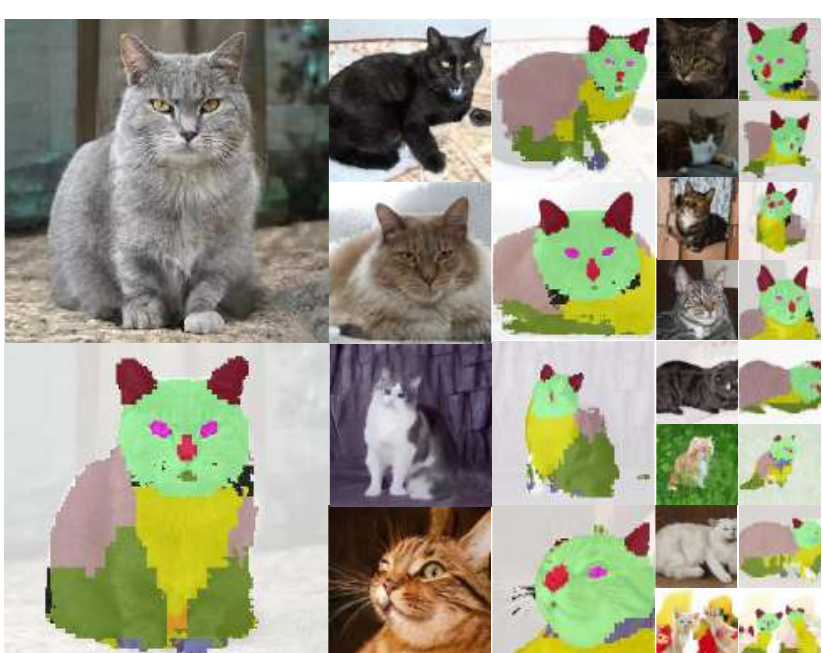} &
\includegraphics[height=3.8cm, trim=330 10 0 0,clip]{figs/stylegan_dataset_2/stylegan_dataset_bedroom_cat_compressed}
\end{tabular}
\end{center}
\vspace{-8mm}
\caption{\footnotesize  Examples of synthesized images and labels  from our {\ours} for birds, cats, bedrooms. StyleGAN was trained on NABirds ($1024 \times 1024$ images), LSUN CAT ($256 \times 256$), and  LSUN Bedroom ($256 \times 256$). 
{\ours} was trained on 30 annotated bird examples, 30 cats, and 40 bedrooms. }
\label{fig:vis_testing2}
\vspace{-4mm}
\end{figure*}

\vspace{-1.8mm}
\section{Collecting Fine-grained Annotations} 
\vspace{-2mm}
\label{sec:annotations}

Real and GAN generated images were annotated with LabelMe~\cite{labelme} by a single experienced annotator. For the GAN-generated images, which are used for training the Style Interpreter, there are 40 bedrooms (1109 polygons), 16 cars (605 polygons), 16 heads (950 polygons), 30 birds (443 polygons), and 30 cats (737 polygons). For each class, we manually defined a partonomy including as many details as it was possible. Fig.~\ref{fig:dataset} shows the partonomy and all the annotated images for two classes. Real images (from different datasets -- see Sec~\ref{sec:experiments}) are used for evaluation only.

{\bf Statistics of Annotating Real vs GAN images:} GANs produce images with different quality depending on the class. In the case of heads, the images are very realistic and the annotation resulted in a similar number of parts than when annotating real images (58 annotated parts on average for GAN images and 55 parts for real images). The annotation of each head, with all the parts takes 1159 seconds (to annotate only the head outline took 74 seconds on average). GANs trained on Birds and Cats result in slightly worst quality images. GAN bird images had 13.7 parts, while real birds were annotated with 17 parts. GAN Cats had 23.6 parts while real cats had 27.4 annotated parts on average. Despite of the slight decrease in the number of parts, the amount of detailed parts available for annotation in GAN generated images is remarkable. 

{\bf Annotation Times:} Birds, with all the parts, took 262 seconds to annotate, and Cats took 484 seconds, on average per image. GAN generated bedrooms are of high quality but contained fewer recognizable objects than real images. GAN generated images have resolution of 256x256 pixels, while the real images were higher resolution. GAN bedrooms had 37.8 annotated objects on average while real bedrooms had 47.8 annotated objects. One average, GAN bedroom images took 629 seconds to annotate while real images took 1583 seconds as they contain more details.

{\bf Limitations:} Since our approach relies on labeling GAN images, image quality sometimes interferes with labeling. Our annotator complained when annotating Birds. Synthesized bird legs are mostly invisible, blurry and unnatural, making annotation challenging. As shown in Fig~\ref{fig:vis_testing2}, our synthesized datasets barely generated the leg labels, which influences test-time performance for this part as a result. 

%% file: doc/4_exp.tex
\vspace{-3.5mm}
\section{Experiments}{}
\label{sec:experiments}
\vspace{-1.5mm}

We extensively evaluate our approach. First, we perform evaluation on part segmentation across five different categories: Car, Face, Bird, Cat, and Bedroom(scene). Furthermore, we also label two keypoint datasets (Car and Bird), and also evaluate keypoint detection performance supported by our approach.  We finally showcase a qualitative 3D application that leverages our synthesized data for Car, to achieve single-image 3D asset creation. 

{\bf StyleGAN models:} Each class requires a pretrained category-specific StyleGAN model. For Car, Cat, Face and Bedroom, we directly use the pretrained StyleGAN models from the official GitHub repository provided by StyleGAN authors.  For bird, we train our own StyleGAN models on NABirds~\cite{7298658}, which contains 48k images.  

\vspace{-2mm}
\subsection{Parts Segmentation} 
\label{sec:parts_result}

\paragraph{Our Generated Datasets: } 
Figures~\ref{fig:vis_testing_cars},~\ref{fig:vis_testing2} show samples of synthesized image-annotation pairs for all classes used in the paper. We notice that higher resolution StyleGAN models (Bird, Face) result in more accurate synthesized annotations.  We provide more examples in the Appendix.

\begin{figure}
\vspace{-2.5mm}
\begin{minipage}{0.62\linewidth}
\includegraphics[width=1\linewidth, trim=5 0 0 0,clip]{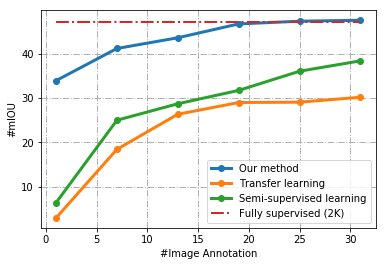}
\end{minipage}
\hspace{1.5mm}
\begin{minipage}{0.32\linewidth}
\vspace{1.5mm}
\caption{\footnotesize\textbf{Number of training examples vs. mIOU}. We compare to baselines on ADE-Car-12 testing set. The red dash line represents the fully supervised method which exploits 2.6k training examples from ADE20k. }	
\label{fig:num_vs_iou}
\end{minipage}
\vspace{-6.5mm}
\end{figure}

\begin{table*}[t!]
\vspace{-2mm}
\begin{center}
\begin{minipage}{0.71\linewidth}
 \begin{adjustbox}{width=1\linewidth}
{\small
\addtolength{\tabcolsep}{-6pt}
\begin{tabular}{|l|c|c|c|c|c|c|c|c|}
\hline
Testing Dataset & ADE-Car-12 & ADE-Car-5 &  Car-20 & CelebA-Mask-8 (Face)  &   Face-34 & Bird-11 & Cat-16 & Bedroom-19  \\
\hline 
Num of Training Images & 	16     & 16             &16        & 16               & 16         & 30       & 30       & 40  \\ 
\hline
Num of Classes  & 12           & 5             & 20 &  8 & 34  &  11 & 16 &   19 \\ 
\hline
\hline
Transfer-Learning& 24.85     &  44.92 & 33.91 $\pm$ 0.57  & 62.83 & 45.77 $\pm$ 1.51   & 21.33 $\pm$ 1.32    & 21.58 $\pm$ 0.61  & 22.52 $\pm$ 1.57   \\ 
\hline
Transfer-Learning (\textbf{*})& 29.71       &  47.22     & \ding{55} & 64.41  &  \ding{55}  & \ding{55} & \ding{55} & \ding{55}  \\ 
\hline
Semi-Supervised~\cite{mittal2019semi} & 28.68  & 45.07  & 44.51 $\pm$ 0.94  & 63.36 &48.17 $\pm$ 0.66  & 25.04 $\pm$ 0.29   &24.85 $\pm$ 0.35   & 30.15 $\pm$ 0.52   \\ 
\hline
Semi-Supervised~\cite{mittal2019semi} (\textbf{*}) & 34.82 & 48.76    & \ding{55} & 65.53 &  \ding{55}  & \ding{55} & \ding{55} & \ding{55}    \\ 
\hline
Ours  & \textbf{45.64}  &  \textbf{57.77} &  \textbf{62.33 $\pm$ 0.55} & \textbf{70.01}  & \textbf{53.46 $\pm$ 1.21}  & \textbf{36.76 $\pm$ 2.11} &  \textbf{31.26 $\pm$ 0.71 } & \textbf{36.83 $\pm$ 0.54}  \\ 
\hline

\end{tabular}\\
}
\end{adjustbox}
{\scriptsize \ding{55} means that the method does not apply to this setting due to missing labeled data in the domain.}
\end{minipage}
\hfill
\begin{minipage}{0.28\linewidth}
\vspace{-0mm}
\caption{\footnotesize {\bf Comparisons on Part Segmentation}.  \textbf{(*)} denotes In-domain experiment, where training and testing are conducted on the same dataset but a different split. Otherwise, training is conducted on our generated images. Note that In-domain setting does not apply to our approach, as we do not train StyleGAN on the provided datasets.} 
\label{tbl:all_results}
\end{minipage}
\end{center}
\vspace{-8mm}
\end{table*}

\begin{table*}[t!]
\vspace{-3mm}
\begin{minipage}{0.71\linewidth}
\begin{center}
 \begin{adjustbox}{width=1\linewidth}
{\scriptsize
\addtolength{\tabcolsep}{-3.5pt}
\begin{tabular}{l|c|c|c|c||c|c|c|c}
\hline
Testing Dataset &  \multicolumn{4}{c}{Car-20}  &   \multicolumn{4}{c}{CUB-Bird}  \\
\hline
Metric &  L2 Loss $\downarrow$ &  PCK th-15  $\uparrow$ &  PCK th-10  $\uparrow$    &  PCK th-5  $\uparrow$  & L2 Loss$\downarrow$	 &  PCK th-25  $\uparrow$   &  PCK th-15   $\uparrow$     &  PCK th-10  $\uparrow$  \\
\hline
Transfer Learn. &  4.4 $\times 10^{-4}$	 &  43.54  &  36.66     &  18.53 &  5.3 $\times 10^{-4}$&  23.17	 &  18.21   &  12.74    \\ 

Ours &  2.4 $\times 10^{-4}$	 &  79.91    &  67.14   &   35.17 &  4.3 $\times 10^{-4}$ & 60.61  &  46.36   &  32.00     \\ 
\hline
Fully Sup. &  \ding{55}	 &  \ding{55}   &  \ding{55}   &   \ding{55} &  3.2 $\times 10^{-4}$ & 77.54  & 65.00   &  53.73     \\ 

\hline

\end{tabular}   
}
\end{adjustbox}
\end{center}
\end{minipage}
\hfill
\begin{minipage}{0.28\linewidth}
\vspace{4mm}
\caption{\footnotesize {\bf Comparisons on Keypoint Detection}.  Our method leads to significantly better results than those obtained by baseline methods. }
\label{tbl:Keypoints_result}
\end{minipage}
\vspace{-6mm}
\end{table*}

\vspace{-2.5mm}
\paragraph{Parts Segmentation Network:}  
For simplicity, we exploit Deeplab-V3~\cite{chen2017deeplab}, with ImageNet pre-trained ResNet151~\cite{he2016deep} backbone, as the part segmentation network to be trained on our synthesized datasets. We let Deeplab-V3 output one probability distribution over all part labels for each pixel. While exploiting part hierarchies in the model is possible, we opted for the simplest approach here. We use Deeplab-V3 as the backbone for all the baseline models.

\vspace{-2.5mm}
\paragraph{Baselines:}   We compare our method to two types of baselines: Transfer-Learning (TL) and Semi-Supervised baseline. For the TL baseline, we initialize the network with the pre-trained weights on semantic segmentation of MS-COCO~\cite{Lin2014MicrosoftCC}, and finetune the last layer on our small human-annotated dataset in a supervised way. This baseline evaluates the standard practice in computer vision of pre-training on a large dataset  and only finetuning in-domain. It does not access unlabeled data on the target domain, but leverages a large labeled dataset from another domain. 
We adopt~\cite{mittal2019semi} as the state-of-the-art semi-supervised baseline, and use the same pre-trained backbone as in our approach. We train this method on our human-labeled images plus the unlabeled \emph{real} images that the StyleGAN is trained on.  To demonstrate the effectiveness of our method, we further compare to a fully supervised baseline which is trained on a large number of labeled real images. Further details are in Appendix.  We emphasize that all methods and our approach use the same segmentation network architecture, and the only difference is the training data and algorithm. 

\vspace{-2.5mm}
\paragraph{Test Datasets:}  
For cars, we evaluate our model on part segmentation at different level of details, to leverage the existing datasets for benchmarking. Car instances from ADE20K~\cite{zhou2016semantic,zhou2017scene} and PASCAL~\cite{Everingham10} have 12 and 5 part labels, respectively.  We split cars in ADE20K testing set into our validation and testing sets, which contains 50 and 250 images. 
We refer to cars from ADE20K as \textbf{ADE-Car-12} and further merge  labels from ADE-Car-12 into 5 classes (\textbf{ADE-Car-5}) according to the PASCAL annotation protocol.  We also exploit 900 cars from PASCAL for cross-domain testing purposes (no training), namely \textbf{PASCAL-Car-5}. For faces, we evaluate our model on CelebA-Mask-8~\cite{liu2015faceattributes}, which contains 30K images with 8 part categories. We exploit the first 500 images in testing set as validation set. 
Since there is no existing fine-detailed part segmentation datasets for cat, bird, and bedrooms and both ADE-Car-12 and CelebA-Mask-8 are relatively coarse compared to our annotation, we manually annotate 20 test images for each category to evaluate the performance on detailed part labeling (described as ``Real" in Sec~\ref{sec:annotations}). We refer to them as~\textbf{Car-20},~\textbf{Face-34},~\textbf{Bird-11},~\textbf{Cat-16},~\textbf{Bedroom-19}, respectively.  We select images for these small test datasets from Stanford Cars~\cite{KrauseStarkDengFei-Fei_3DRR2013}, Celeb-A mask~\cite{liu2015faceattributes}, CUB~\cite{cub}, and Kaggle Cat~\cite{kaggle_cat}, respectively. For bedrooms, we pick 20 images from the web. A summary of all test datasets is in Table~\ref{tbl:all_results}.  Since there is no validation set for our annotated datasets, we split testing images into five folds. We set each fold as validation and choose checkpoints accordingly. We report mean IOU and standard deviation.

\begin{table*}[h!]
\vspace{-2mm}
\begin{minipage}{.45\linewidth}
\begin{center}
{\footnotesize
\addtolength{\tabcolsep}{-0pt}
\begin{tabular}{|l|c|c|c|c|}
\hline
Generated Dataset Size & 3K & 5K & 10K & 20K \\ 
\hline
mIOU &  43.34 & 44.37  &   44.60 &  {\bf 45.04} \\
\hline
\end{tabular}   
\vspace{-3mm}
\caption{\footnotesize {\bf Ablation study of synthesized dataset size.} Here, Style-Interpreter is trained on 16 human-labeled images.  Results are reported on ADE-Car-12 test set. Performance is slowly saturating. } 
\label{tbl:number_data}
}
\end{center}
\end{minipage}
\hspace{2mm}
\begin{minipage}{.535\linewidth}
\begin{center}
{\footnotesize
\addtolength{\tabcolsep}{-0pt}
\begin{tabular}{|l|c|c|c|c|}
\hline
Filtering Ratio & 0\% & 5\% & 10\% & 20\% \\ 
\hline
mIOU &  44.60 & 44.89  &   \textbf{45.64}  & 45.18 \\
\hline
\end{tabular}   
\vspace{-3mm}
\caption{\footnotesize {\bf Ablation study of the filtering ratio.} We filter out the most uncertain synthesized Image-Annotation pairs.  Result shown are reported on ADE-Car-12 test set, using the generated dataset of size 10k. We use $10\%$ in other experiments.} 
\label{tbl:uncertainty_denoise}
}
\end{center}
\end{minipage}
\begin{minipage}{.48\linewidth}
\begin{center}
 \begin{adjustbox}{width=0.87\linewidth}
{\footnotesize
\addtolength{\tabcolsep}{-2.5pt}
\begin{tabular}{|l|c|c|c|}
\hline
Testing Dataset &  ADE-Car-5  & PASCAL-Car-5 \\
\hline
Num of Classes & 	5             & 5      \\
\hline
\hline
Deeplab-V3~\cite{chen2017deeplab}  (2600 labels)  &  59.41 (\textbf{*})             & 54.31  \\ 
\hline
Ours (25 labels)  &  57.71 &   55.65  \\ 
\hline
\end{tabular}   
}
\end{adjustbox}
\vspace{-3mm}
\caption{\footnotesize {\bf Comparisons to fully supervised methods for Part Segmentation. } \textbf{(*)} denotes In domain experiments. Deeplab-V3 is trained on ADE-CAR and our model is trained on our generated dataset.  } 
\label{tbl:coarse_car_cross_domain}
\end{center}
\end{minipage}
\hspace{3mm}
\begin{minipage}{.5\linewidth}
\begin{center}
 \begin{adjustbox}{width=0.91\linewidth}
{\footnotesize
\addtolength{\tabcolsep}{-2.5pt}
\begin{tabular}{|l|c|c|c|c|}
\hline
Number of Annotated Images  & 1  & 7  &  13 & 19 \\ 
\hline 
Random & /& 40.06 $\pm$ 1.32 & 42.44 & 44.41 \\
\hline
Active Learning &  / & 40.88   &  43.49 & 46.82 \\
\hline
Manual &  33.92 & 41.19 &  43.61 & 46.74 \\
\hline
\end{tabular}  
}
\end{adjustbox} 
\vspace{-3mm}
\caption{\footnotesize {\bf Data selection.} We compare different strategies for selecting StyleGAN images to be annotated manually.  mIoU is reported on ADE-Car-12 test set.  We compute  mean \& var over 5 random runs  with 1 \& 7 training examples.} 
\label{tbl:AL}
\end{center}
\end{minipage}
\vspace{-4mm}
\end{table*}

\vspace{-2.5mm}
\paragraph{Quantitative Comparison:}
\vspace{-2mm}
We first compare our approach to Transfer-Learning and Semi-supervised baselines in Tab~\ref{tbl:all_results}. 
We evaluate in both out-of-domain and in-domain settings, where baselines are trained on our annotated images (Sec.~\ref{sec:annotations}) or an equal number of randomly selected in-domain images (ADE cars). Note that our method falls in the out-of-domain setting, since we only train on our synthesized dataset and test on real images. Our method outperforms both Transfer-Learning and Semi-Supervised learning baselines on all classes by a large margin. Strikingly, on ADE-Car-12, our model outperforms the out-of-domain baselines by 20.79\% for Transfer-Learning and 16.96\% for Semi-supervised Learning, and outperforms two in-domain baselines by 15.93\% and 10.82\%, respectively. 

We further show the number of training images in our labeled datasets v.s mIOU and compare to baselines on ADE-Car-12 test set in Fig.~\ref{fig:num_vs_iou}. The red dash line is the fully supervised model trained on the full ADE-Car-12 training set (2600 images). Our approach, using the same architecture and hyperparameters, comparable with the fully supervised model with as few as \textbf{25} annotations, which is less than 1\% of what the fully supervised method uses. Finally, we show a comparison to fully supervised baseline on ADE-Car-5 and PASCAL-Car-5 in Table~\ref{tbl:coarse_car_cross_domain}. Here, our performance is not better than the fully supervised baseline on ADE-Car-5. We hypothesize this is due to ADE20K being out-of-domain for our model, and in-domain for the baseline, and that 2500 labeled examples used by the baseline are sufficient to train a good model for this easier 5-class task. 
Note that we outperform the baseline by 1.3\% when  both our models are evaluated in the out-of-domain setting on PASCAL-Car-5, showcasing better generalization capabilities.



\vspace{-4.5mm}
\paragraph{Ablation Studies:}
We ablate choices in our  approach on the Car category with 16 training examples. We first ablate the size of generated dataset in Table~\ref{tbl:number_data}. Increasing the number of synthesized examples from 3,000 to 10,000 improves performance, however the improvement is marginal when we further add more data. We use the uncertainty denoising strategy described in Sec.~\ref{sec:factory} and report results of filtering the most uncertain examples using different ratios. As shown in Table~\ref{tbl:uncertainty_denoise}, denoising plays an important role. Removing noisy data is the result of a trade-off between diversity and uncertainty. Removing more uncertain (noisy) data means less diversity during training.  In experiments hereon, we set the size of the generated dataset to be 10,000 and filter out the top 10\% uncertain examples. 

\vspace{-4.5mm}
\paragraph{Training Data Selection:}
\label{sec:al_result}
In our approach, as few as \textbf{20} training examples are required for achieving good accuracy, which is remarkable for this level of detail. In such a low-data regime, selecting the right images to be manually labeled is crucial. We ablate three different options. The most straightforward selection protocol is to simply choose the images  randomly from the generated dataset. A more time consuming option, but one that is typically used when collecting datasets, is to 
employ a human (CV expert) to look through the dataset and select the most representative and diverse examples. Finally, as a more advanced strategy, active learning (AL) can be used, 
where selection and model training (training Style Interpreter) alternate in a continuous loop.  
Similarly to~\cite{Kuo2018CostSensitiveAL}, we exploit ensemble-based AL strategy~\cite{8579074} followed by applying coreset~\cite{Sener2017AGA}. We reuse the ensembles and JS divergence as described in Sec.~\ref{sec:factory} to calculate image uncertainty. We filter out the top k\% most uncertain examples and run coreset with $N$ centers on the top $k + 10 \%$ to $k \%$ percent of data to select the most representative examples. We use $N=12$ and $k=10$ in this paper. Finally, we ask our CV expert to select the top 6 most realistic images to be annotated out of the subset.

We compare these strategies in Table~\ref{tbl:AL}. Our experiments always start with mean of 10k random samples as the first example and AL selects 6 training examples all together in each round.
Both manual and AL outperform random sampling. We also report standard deviation of the random selection on 7 training examples, computed over 5 rounds. Upper bound performance of RS is similar to AL or the manual strategy. Note that AL requires re-labeling each time an experiment is run, and thus is not practical for the remainder of the paper. We instead exploit the manual selection strategy.

\vspace{-4.5mm}
\paragraph{Qualitative Results:}
We showcase qualitative results on our test datasets in Fig~\ref{fig:vis_testing}. While not perfect, the results demonstrate that our approach leads to the labeling outputs of impressive quality, especially for operating in the few-shot regime. Most errors occur for thin parts (wrinkles or bird legs) or parts without visual boundaries (cat neck). 


\newcommand\hh{1.28cm}
\newcommand\ww{1.315cm}
\begin{figure*}[t!]
\vspace{-0mm}
\addtolength{\tabcolsep}{-5.5pt}
\begin{tabular}{p{1.25cm}cccccccccccc}
{\scriptsize {\bf Cars} $\qquad$  20  cls.} &
\includegraphics[align=c,height=\hh,width=\ww, trim=0 0 0 0,clip]{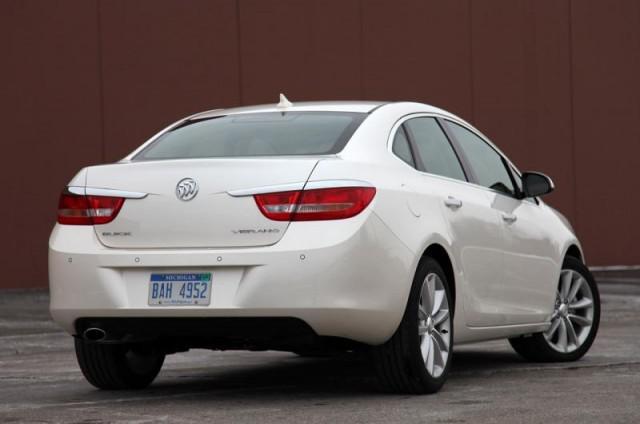} & \includegraphics[align=c,height=\hh,width=\ww, trim=0 0 0 0,clip]{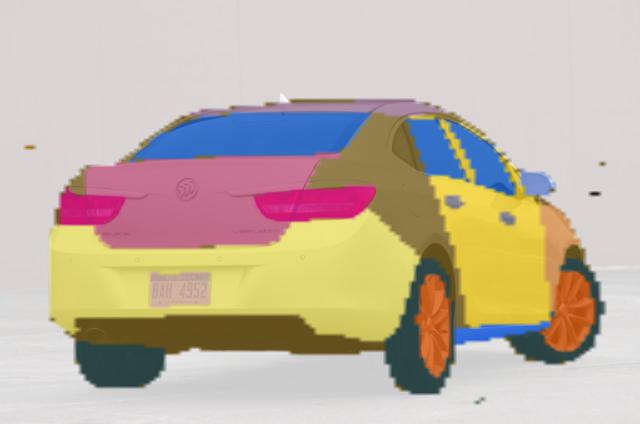} & \includegraphics[align=c,height=\hh,width=\ww,  trim=0 0 0 0,clip]{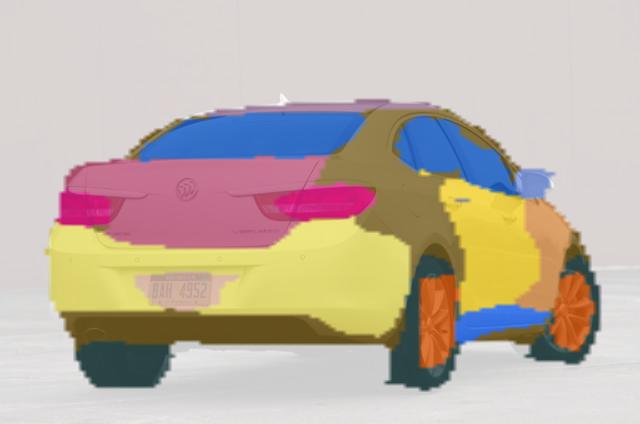} &
\includegraphics[align=c,height=\hh,width=\ww,  trim=0 0 0 0,clip]{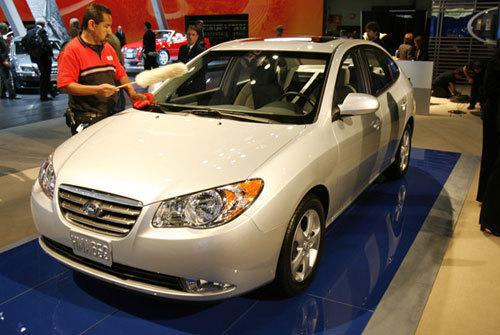} & \includegraphics[align=c,height=\hh,width=\ww,  trim=0 0 0 0,clip]{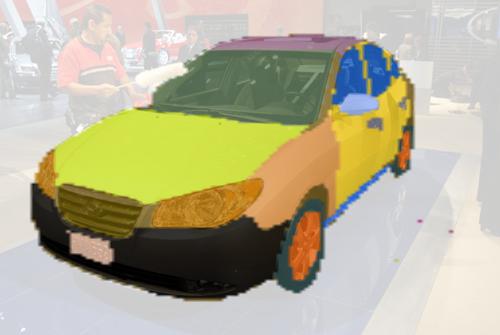} & \includegraphics[align=c,height=\hh,width=\ww,  trim=0 0 0 0,clip]{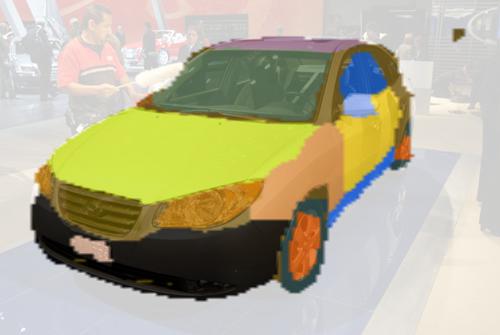} &
\includegraphics[align=c,height=\hh,width=\ww,  trim=0 0 0 0,clip]{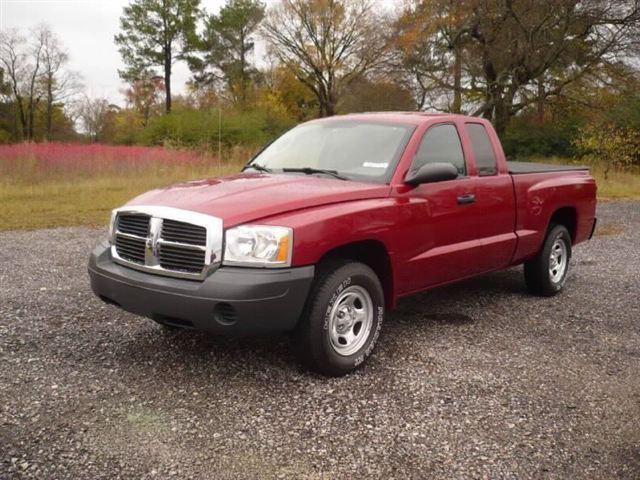} & \includegraphics[align=c,height=\hh,width=\ww,  trim=0 0 0 0,clip]{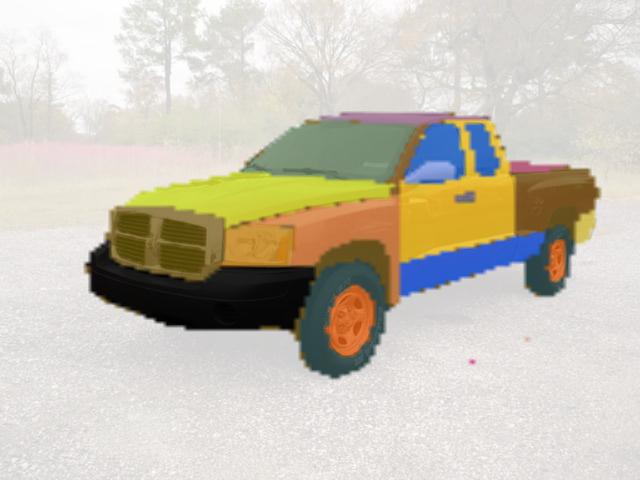} & \includegraphics[align=c,height=\hh,width=\ww,  trim=0 0 0 0,clip]{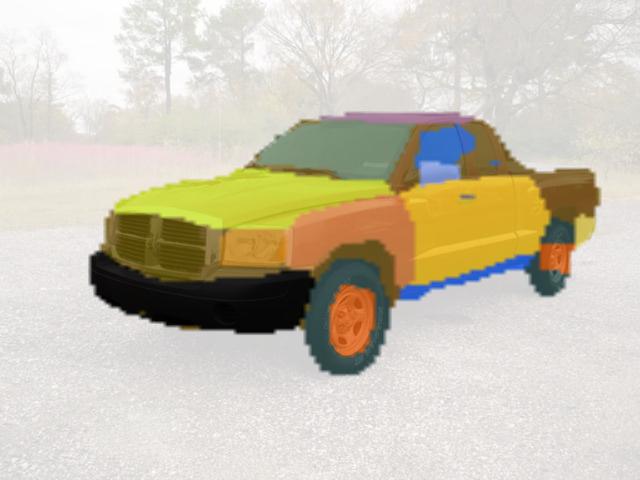} &
\includegraphics[align=c,height=\hh,width=\ww,  trim=0 0 0 0,clip]{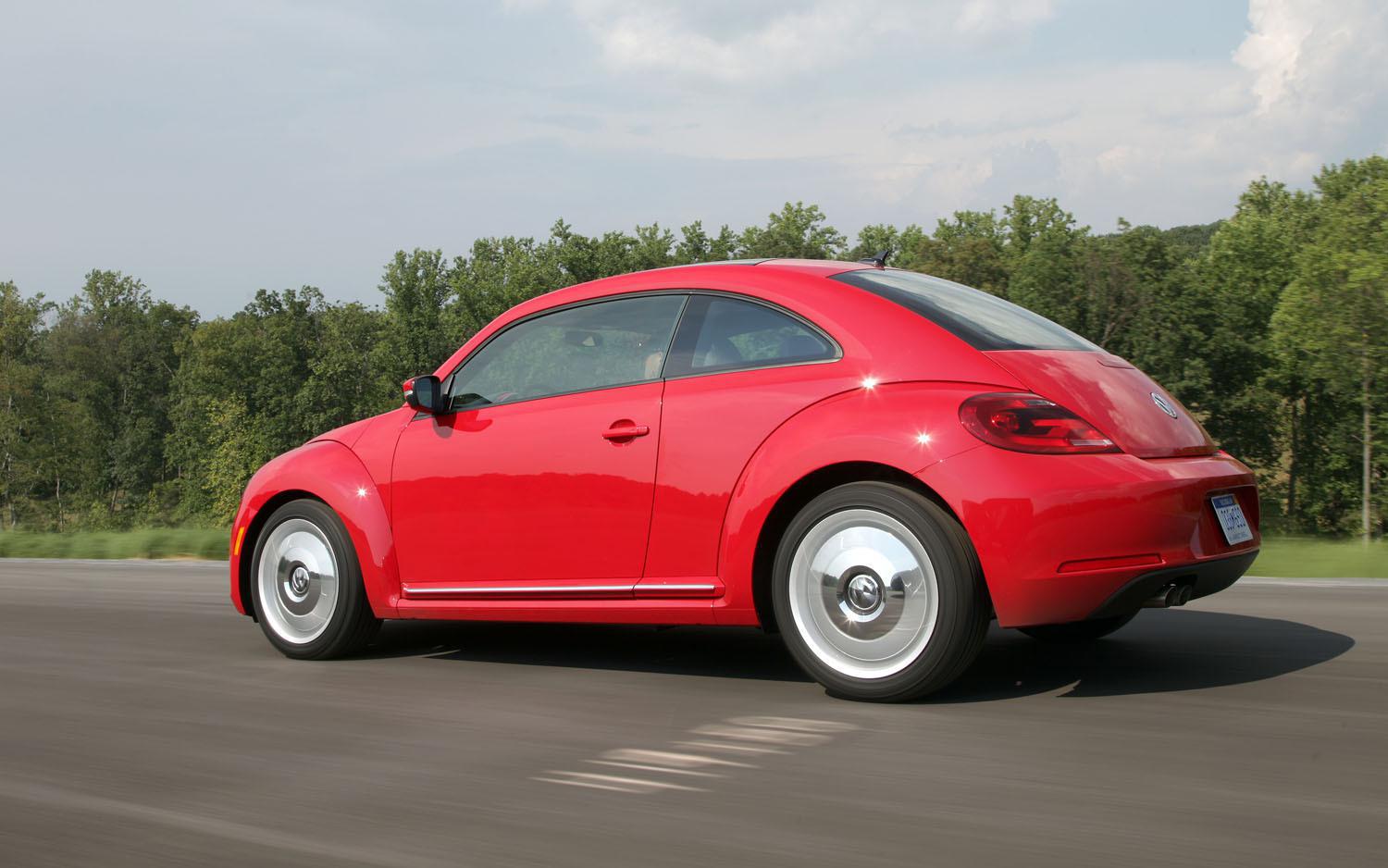} & \includegraphics[align=c,height=\hh,width=\ww,  trim=0 0 0 0,clip]{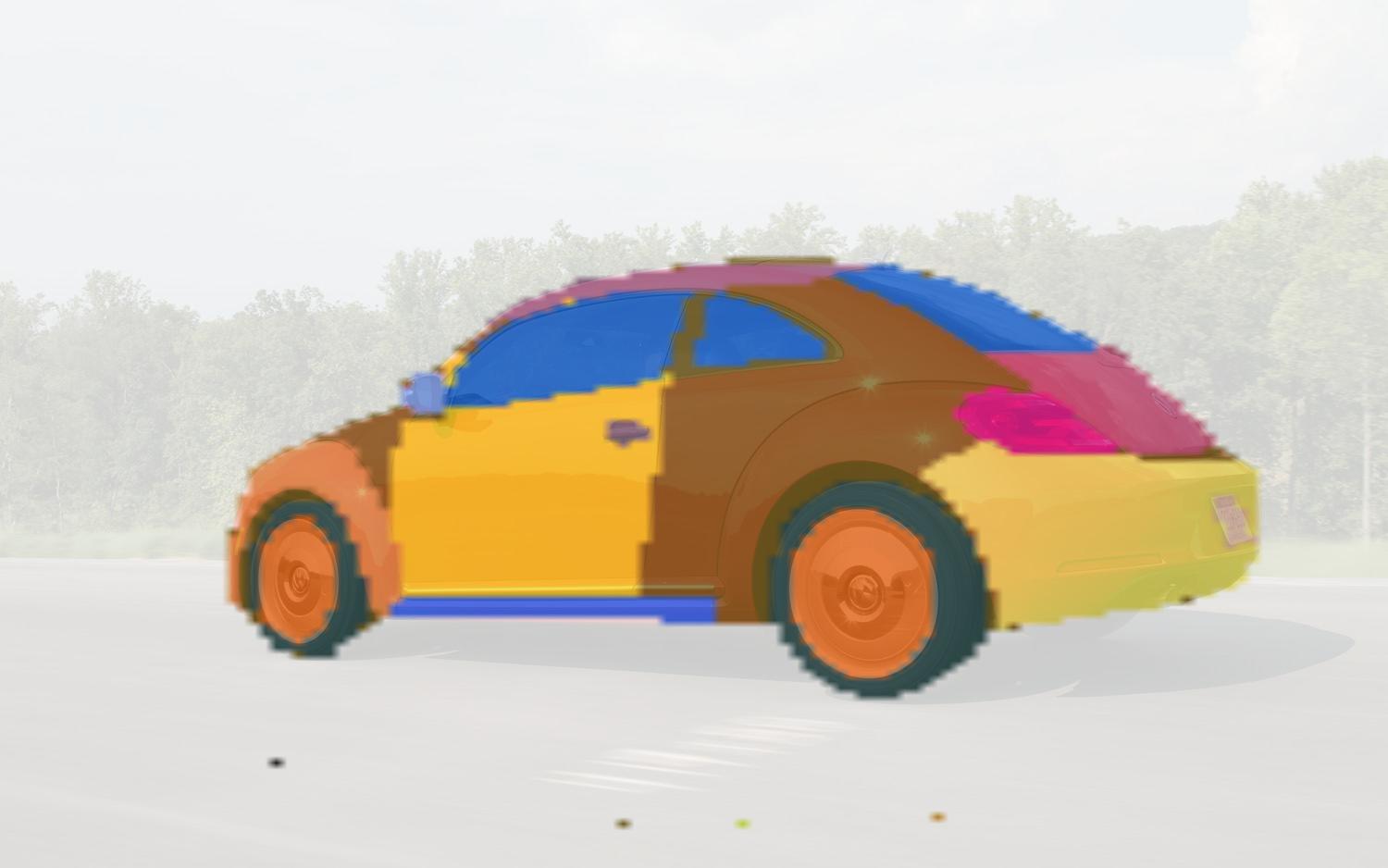} & \includegraphics[align=c,height=\hh,width=\ww,  trim=0 0 0 0,clip]{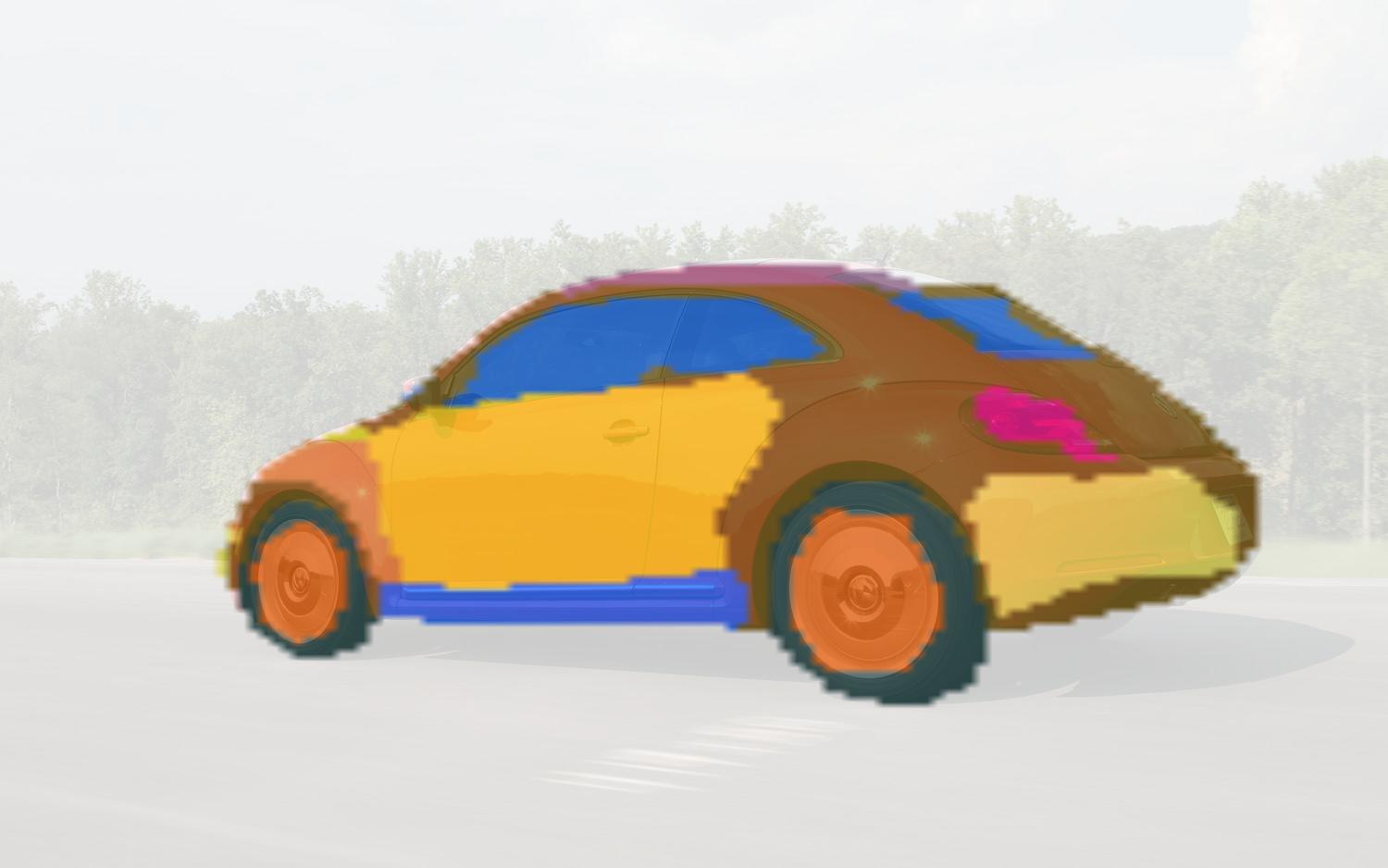} \\
{\scriptsize {\bf Cars} $\qquad$  32  cls.} &
\includegraphics[align=c,height=\hh,width=\ww, trim=0 0 0 0,clip]{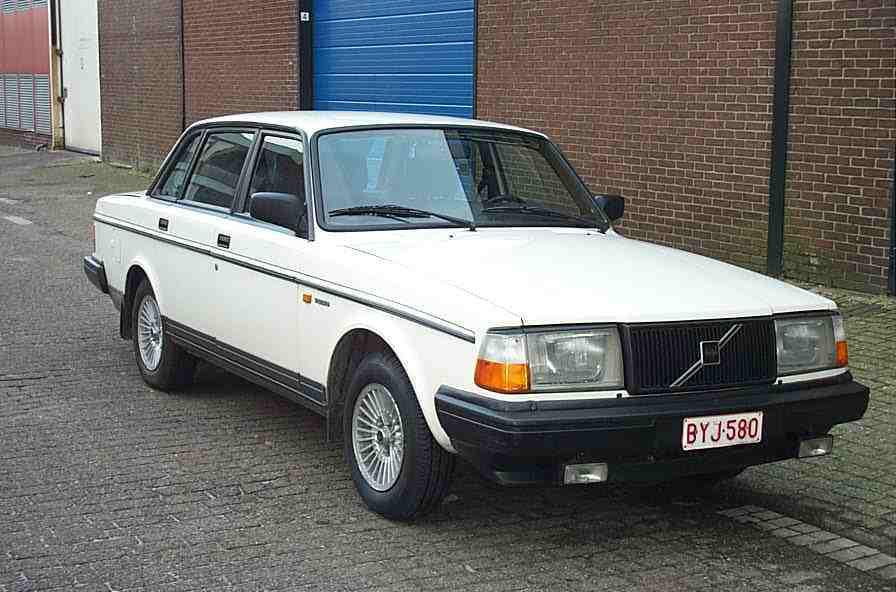} & \includegraphics[align=c,height=\hh,width=\ww, trim=0 0 0 0,clip]{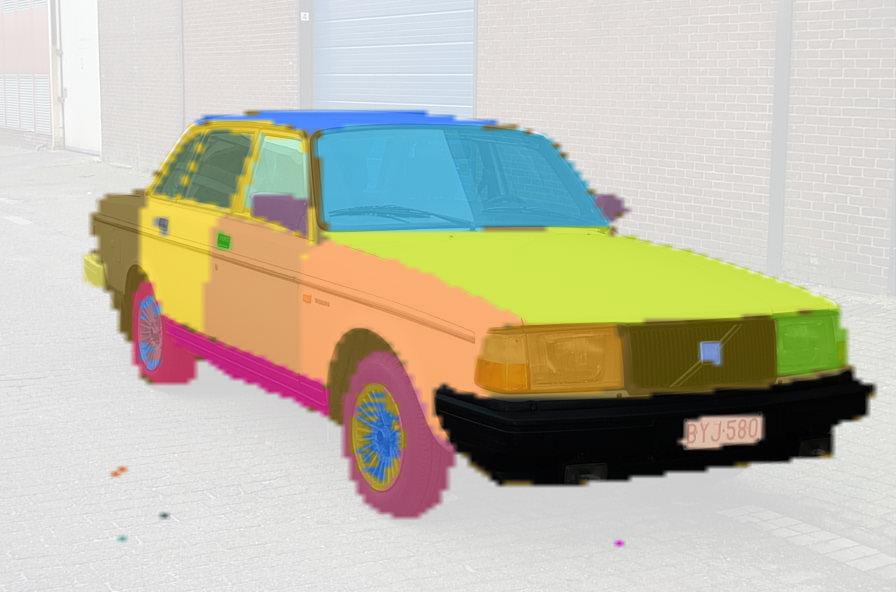} & \includegraphics[align=c,height=\hh,width=\ww,  trim=0 0 0 0,clip]{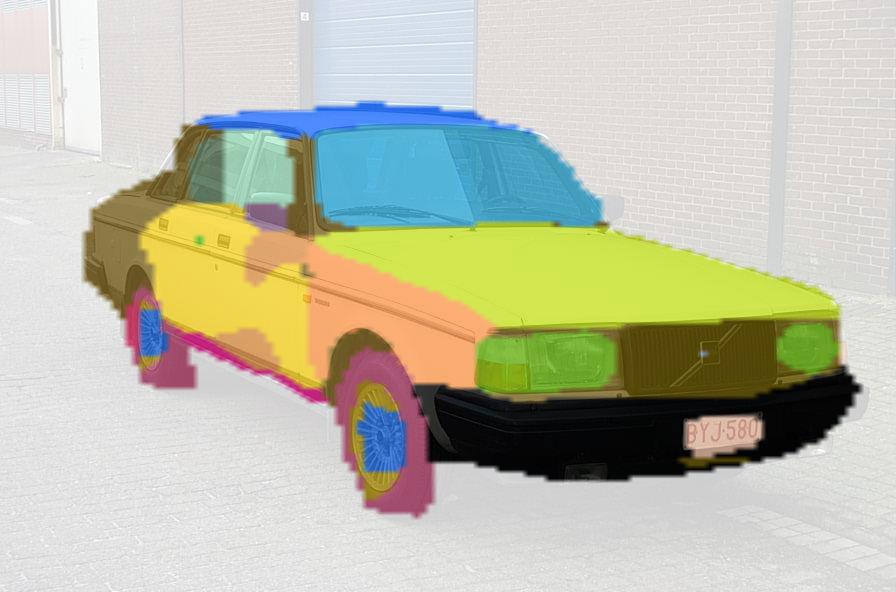} &
\includegraphics[align=c,height=\hh,width=\ww, trim=0 0 0 0,clip]{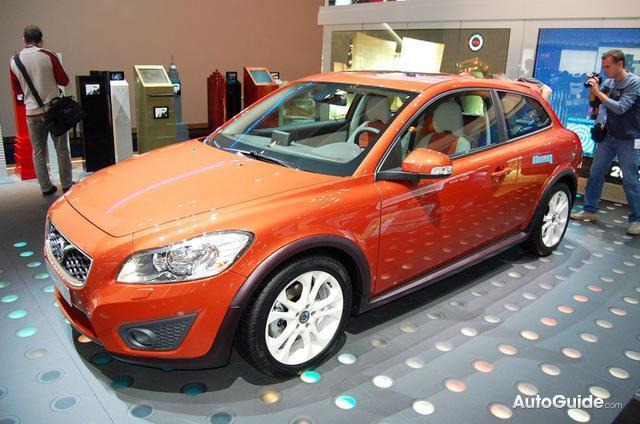} & \includegraphics[align=c,height=\hh,width=\ww, trim=0 0 0 0,clip]{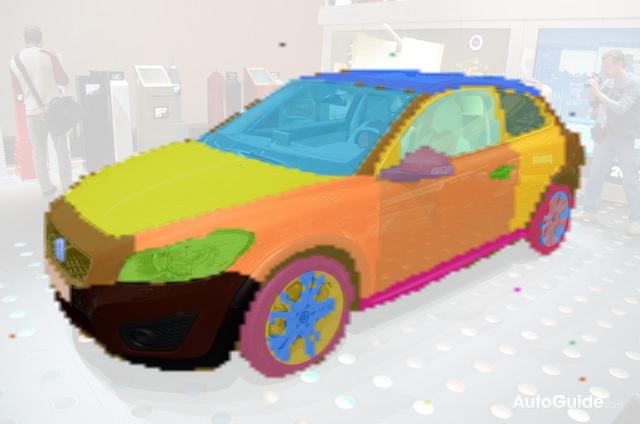} & \includegraphics[align=c,height=\hh,width=\ww,  trim=0 0 0 0,clip]{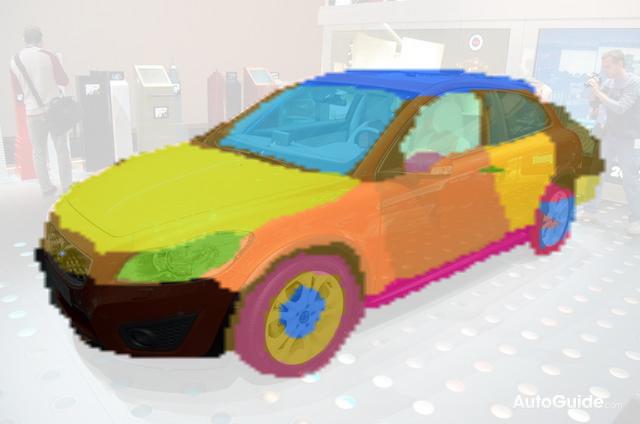} &
\includegraphics[align=c,height=\hh,width=\ww,  trim=0 0 0 0,clip]{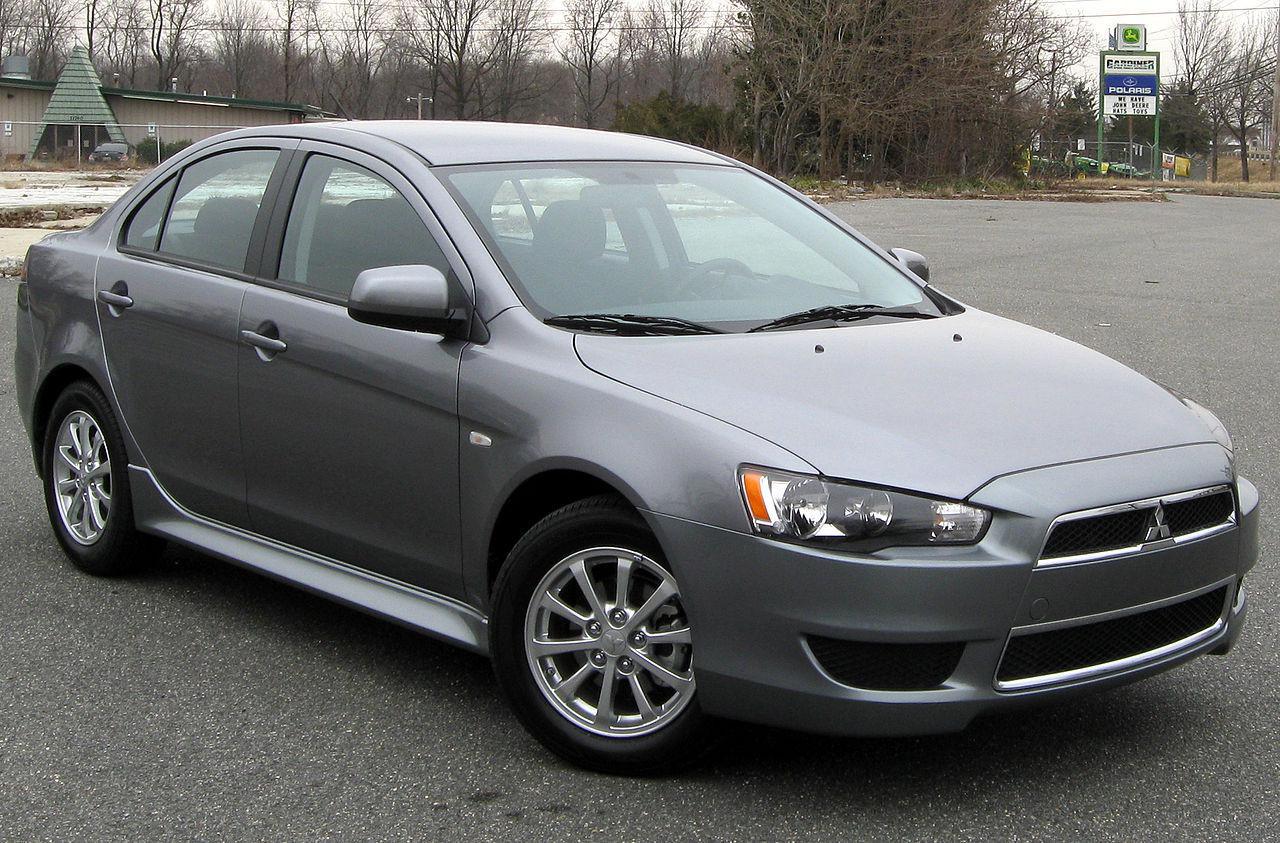} & \includegraphics[align=c,height=\hh,width=\ww,  trim=0 0 0 0,clip]{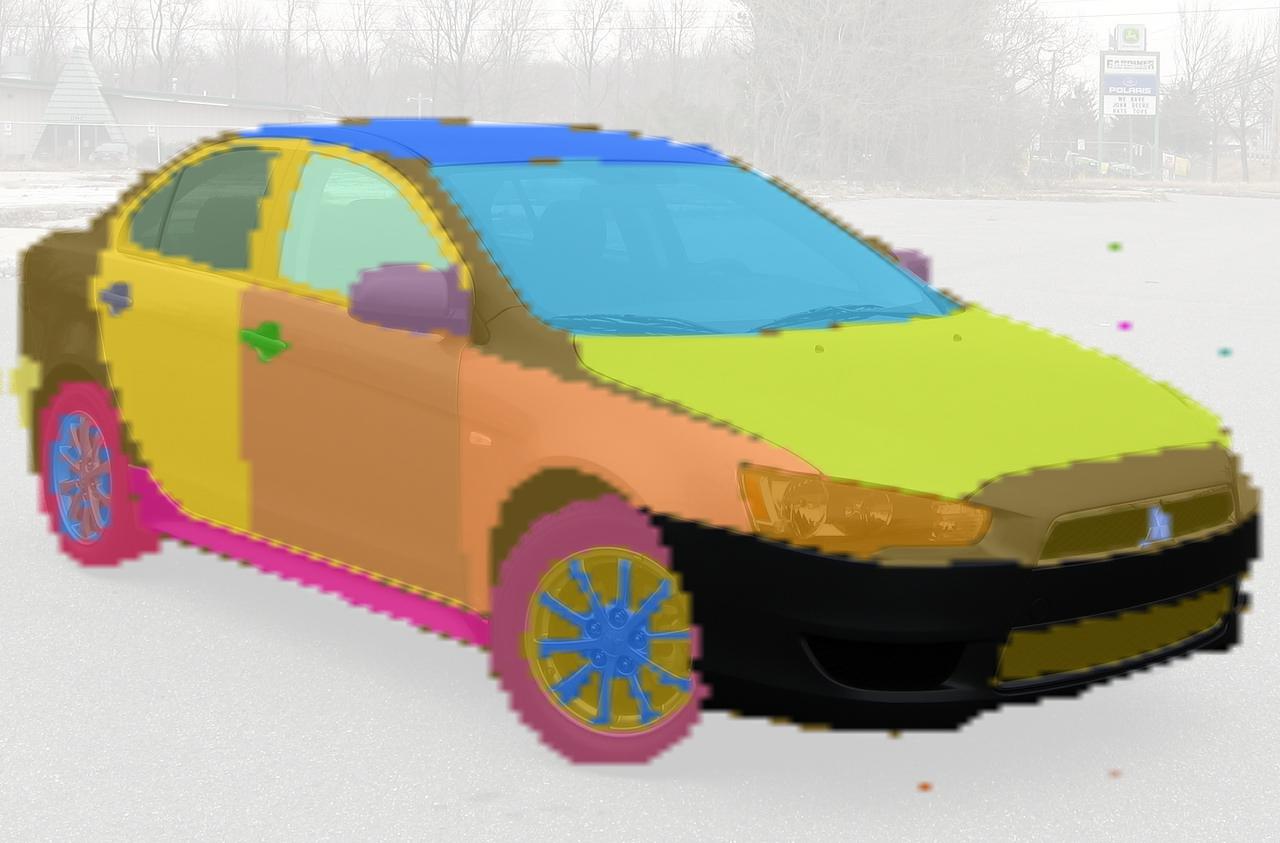} & \includegraphics[align=c,height=\hh,width=\ww,  trim=0 0 0 0,clip]{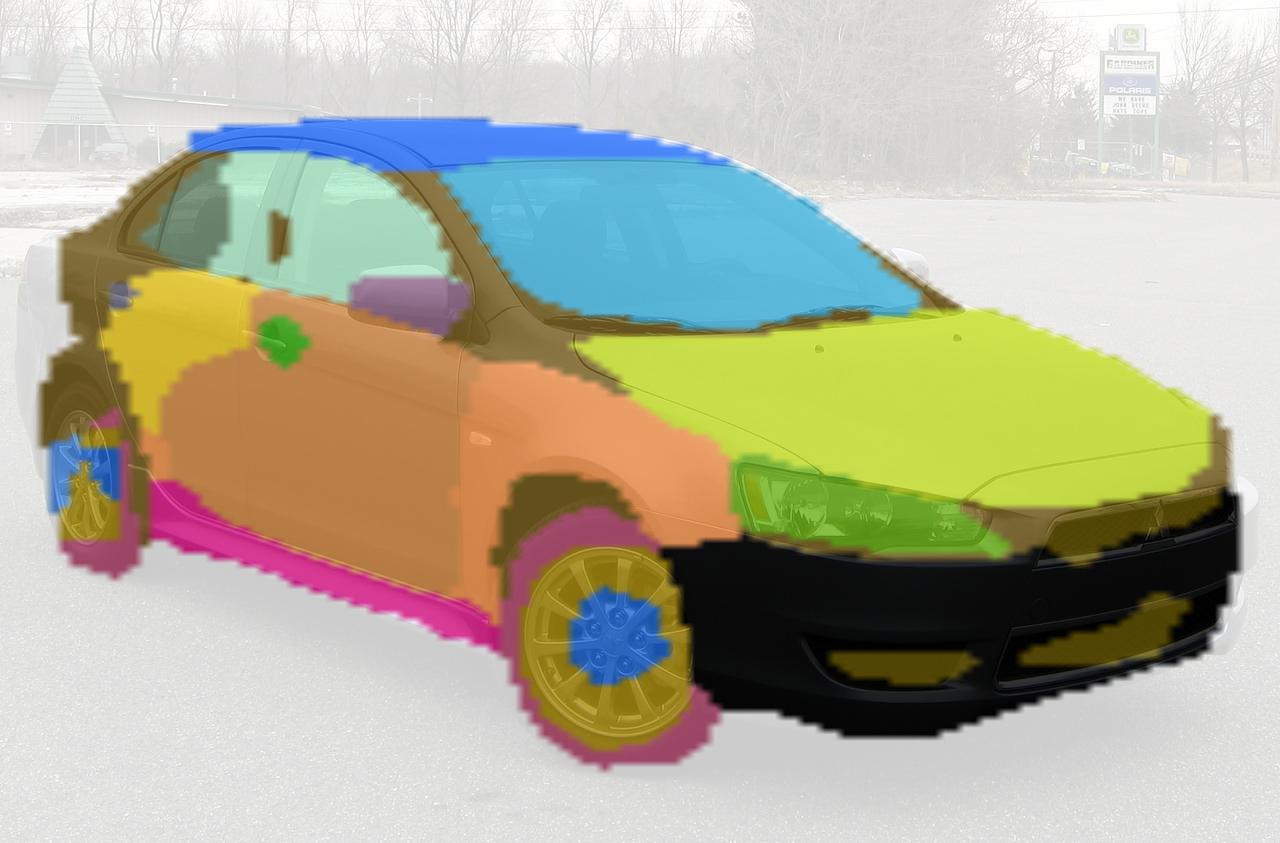} &
\includegraphics[align=c,height=\hh,width=\ww,  trim=0 0 0 0,clip]{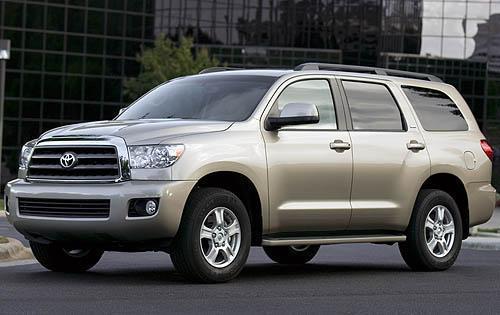} & \includegraphics[align=c,height=\hh,width=\ww,  trim=0 0 0 0,clip]{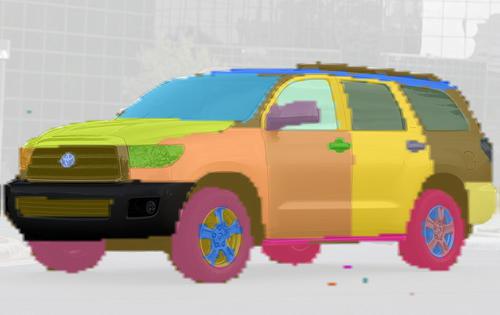} & \includegraphics[align=c,height=\hh,width=\ww,  trim=0 0 0 0,clip]{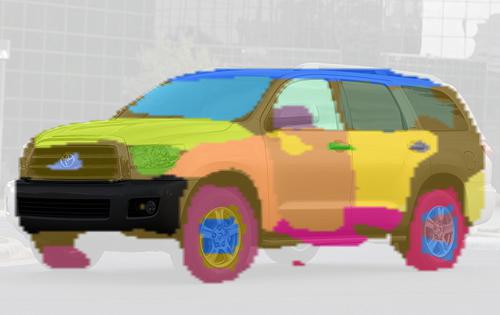} \\
{\scriptsize {\bf Faces} $\qquad$  34  cls.} &
\includegraphics[align=c,height=\hh,width=\ww, trim=0 0 0 0,clip]{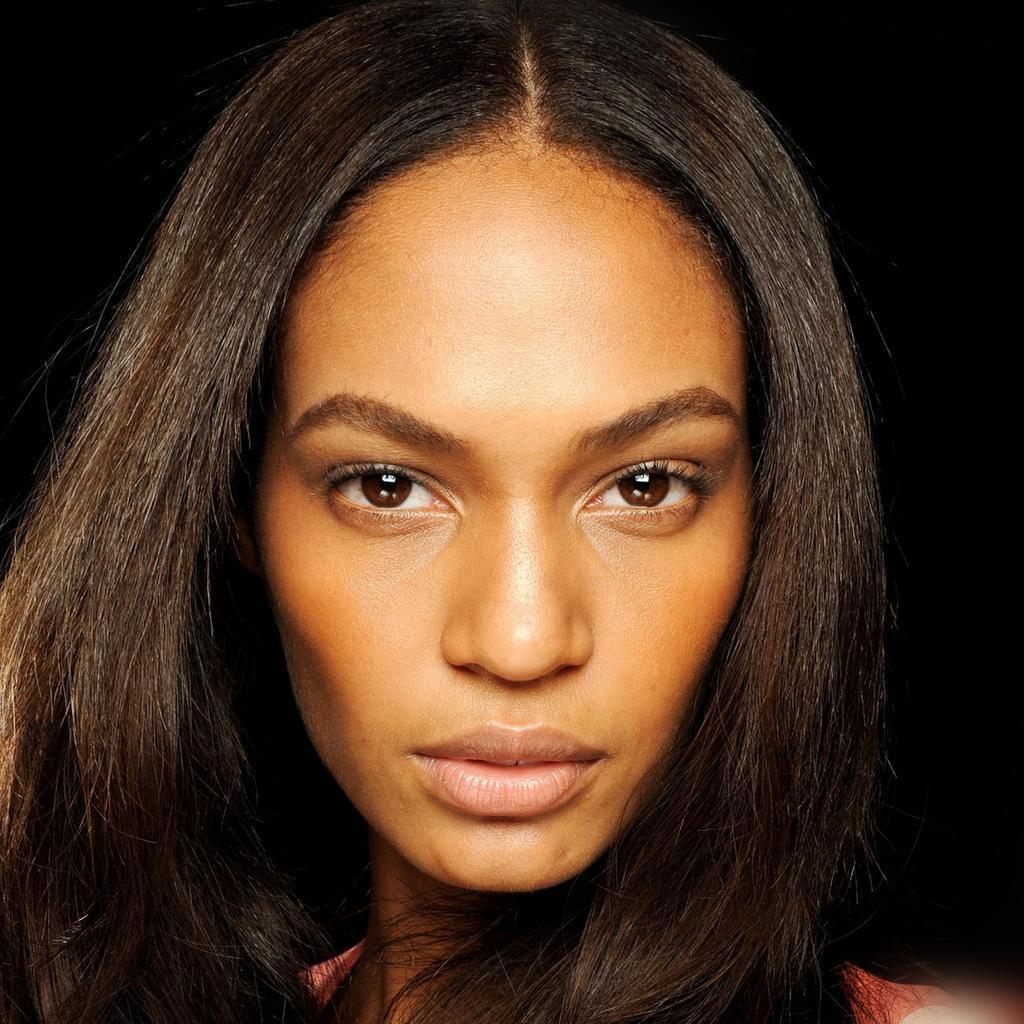} & \includegraphics[align=c,height=\hh,width=\ww, trim=0 0 0 0,clip]{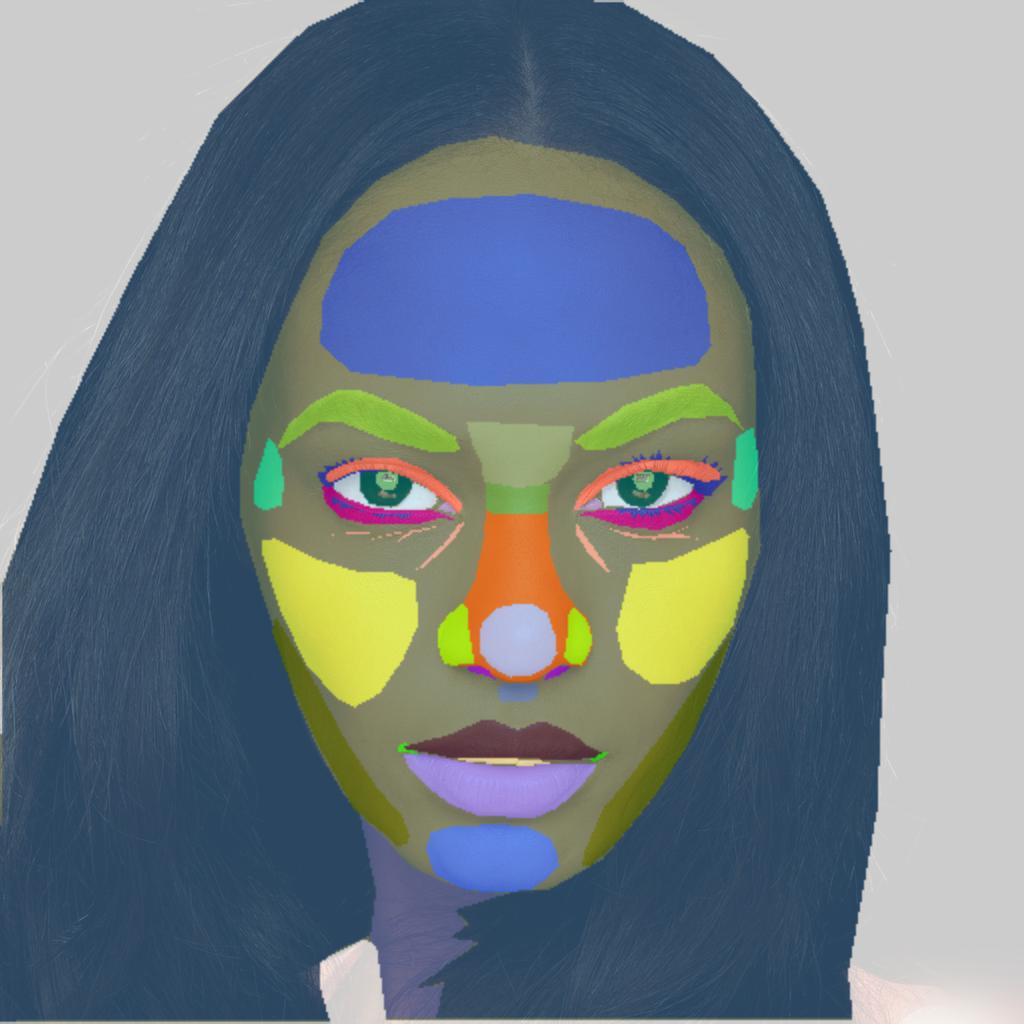} & \includegraphics[align=c,height=\hh,width=\ww,  trim=0 0 0 0,clip]{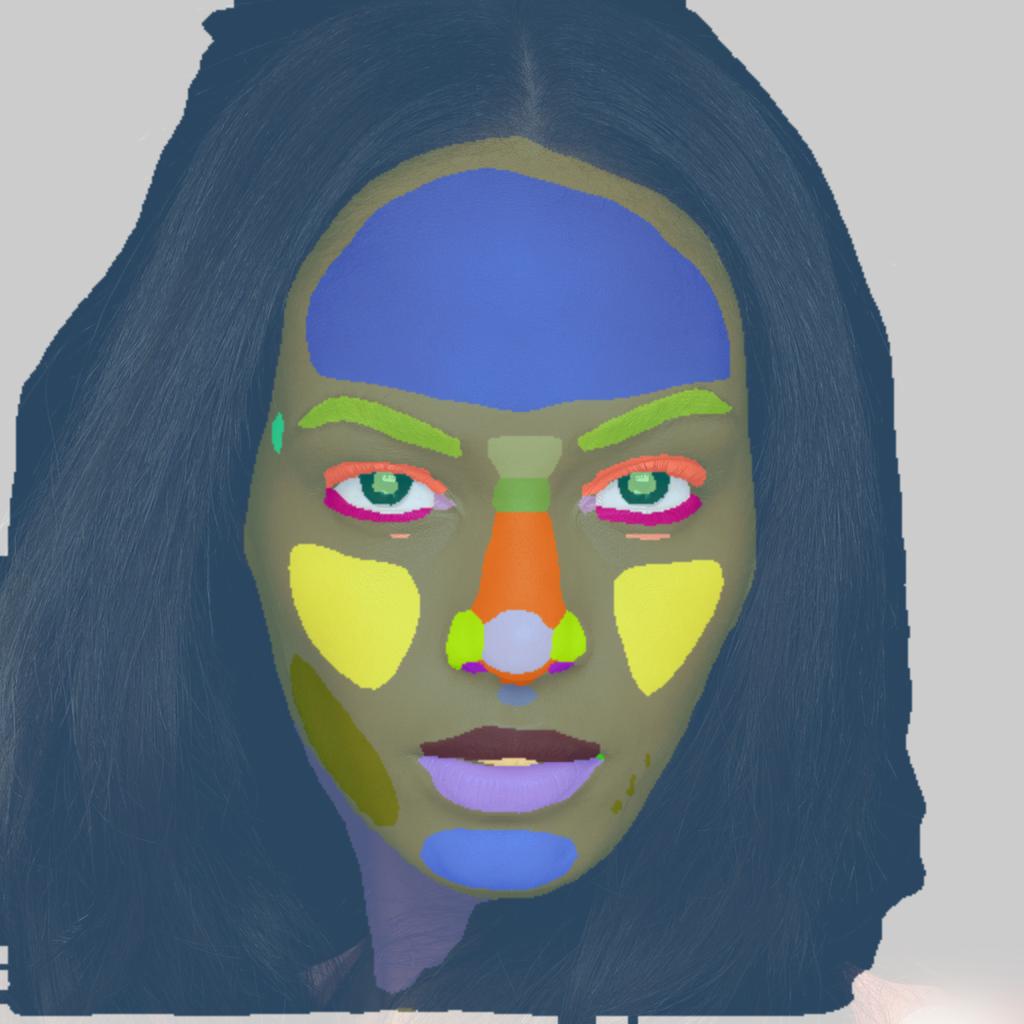} &
\includegraphics[align=c,height=\hh,width=\ww,  trim=0 0 0 0,clip]{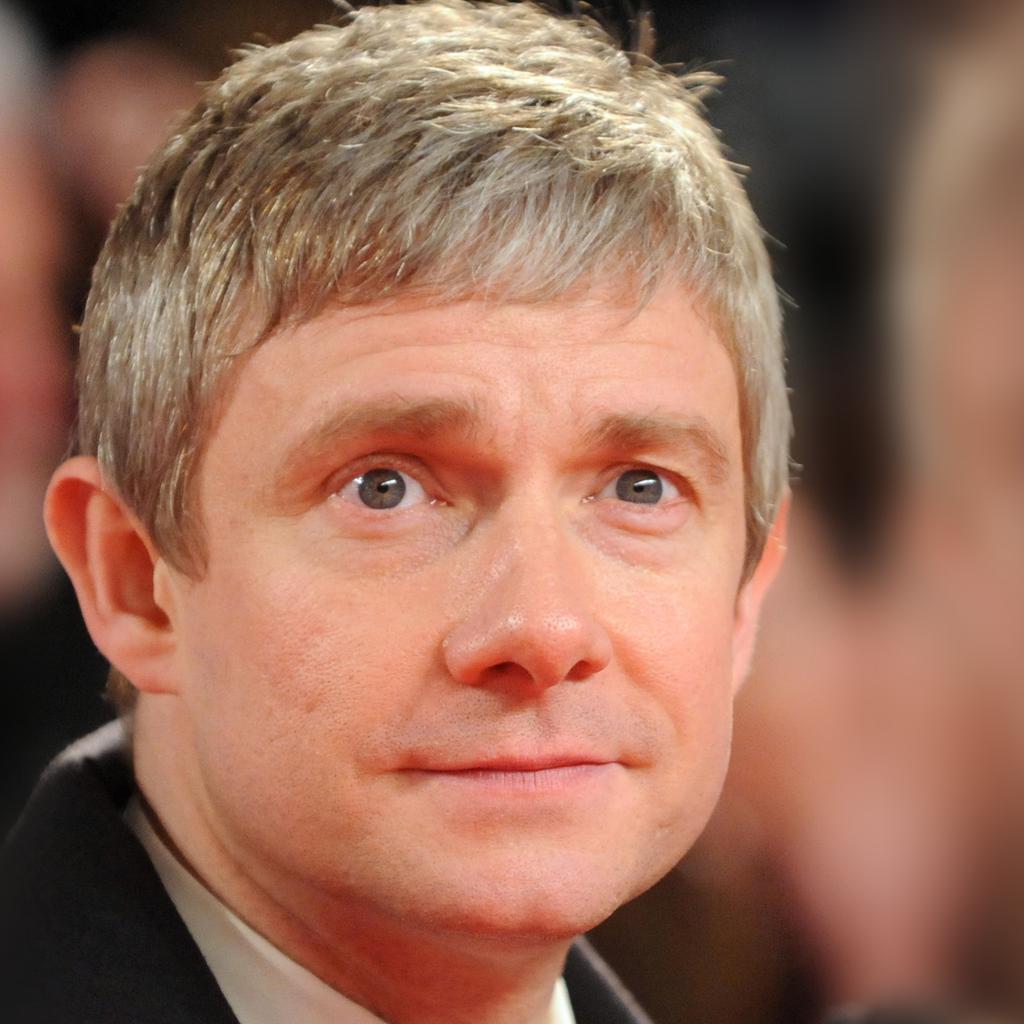} & \includegraphics[align=c,height=\hh,width=\ww,  trim=0 0 0 0,clip]{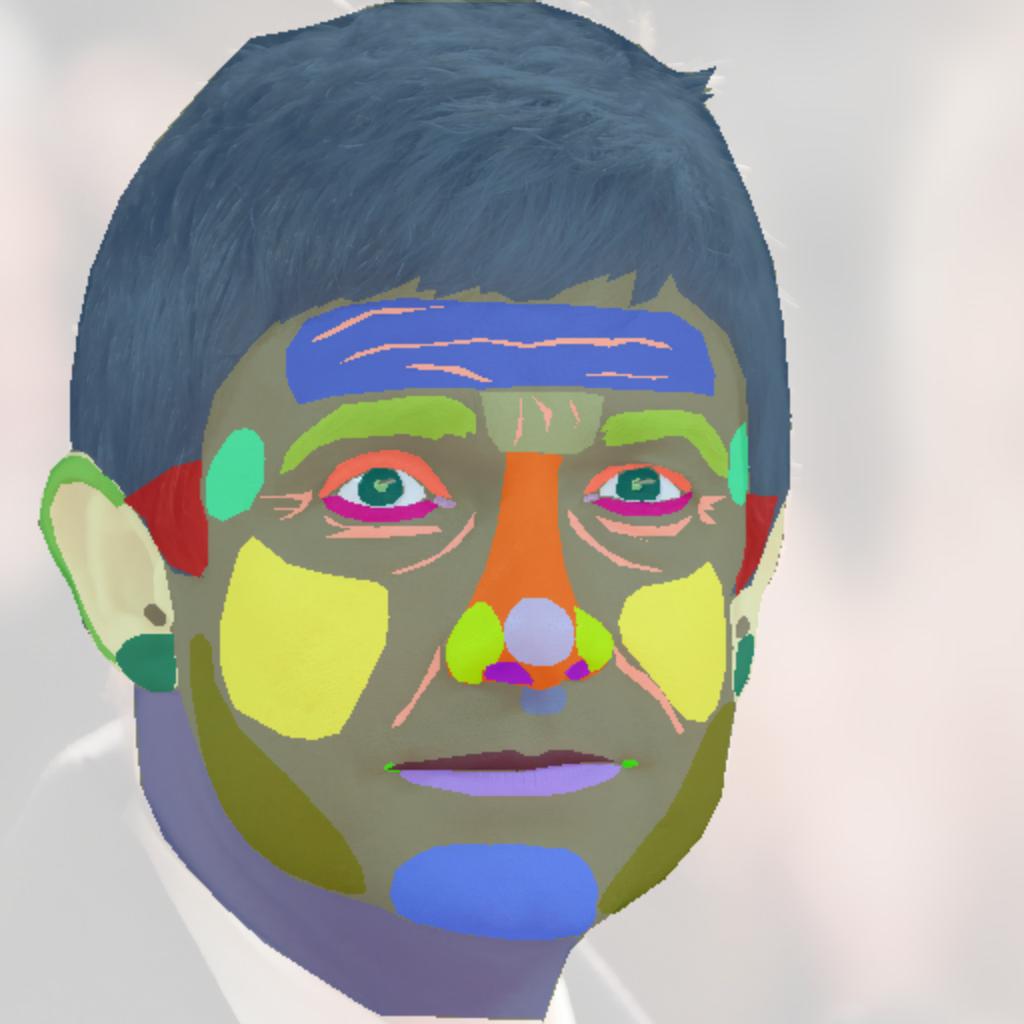} & \includegraphics[align=c,height=\hh,width=\ww,  trim=0 0 0 0,clip]{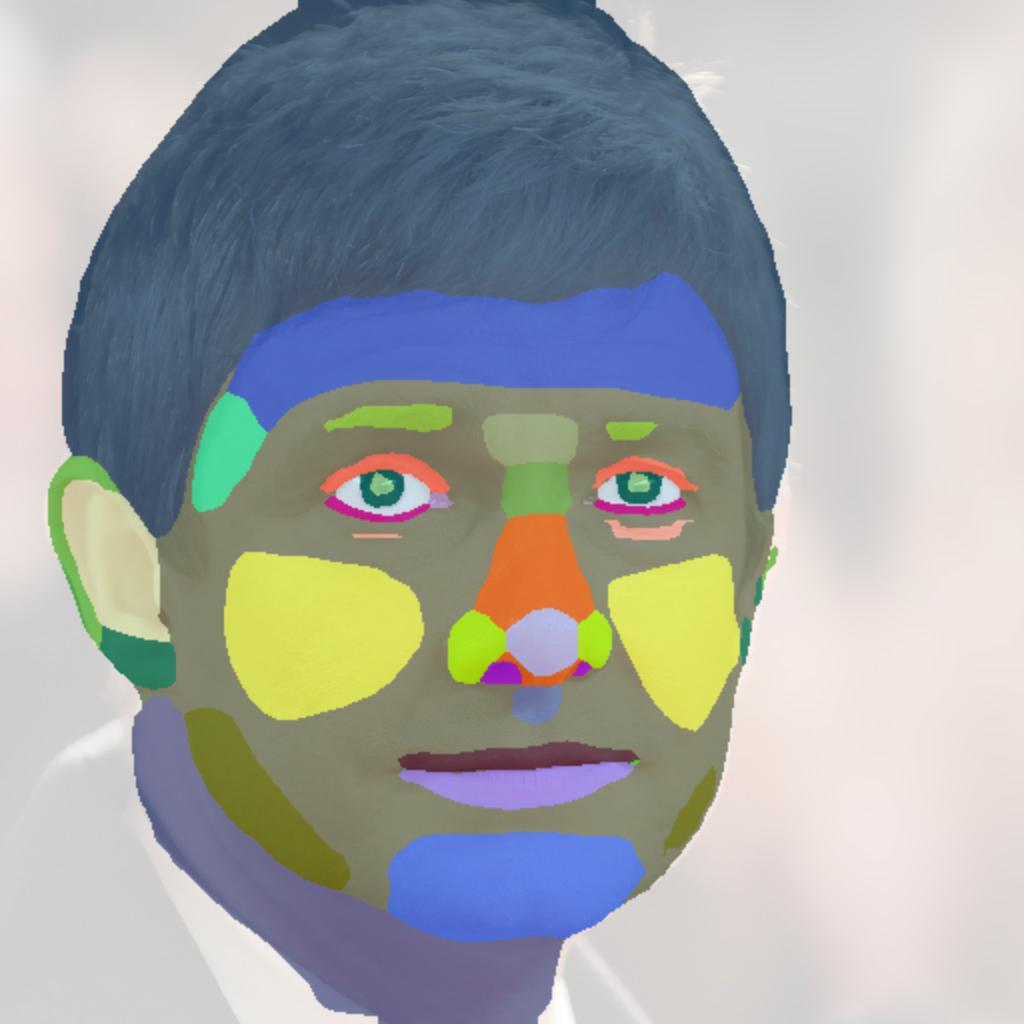} &
\includegraphics[align=c,height=\hh,width=\ww,  trim=0 0 0 0,clip]{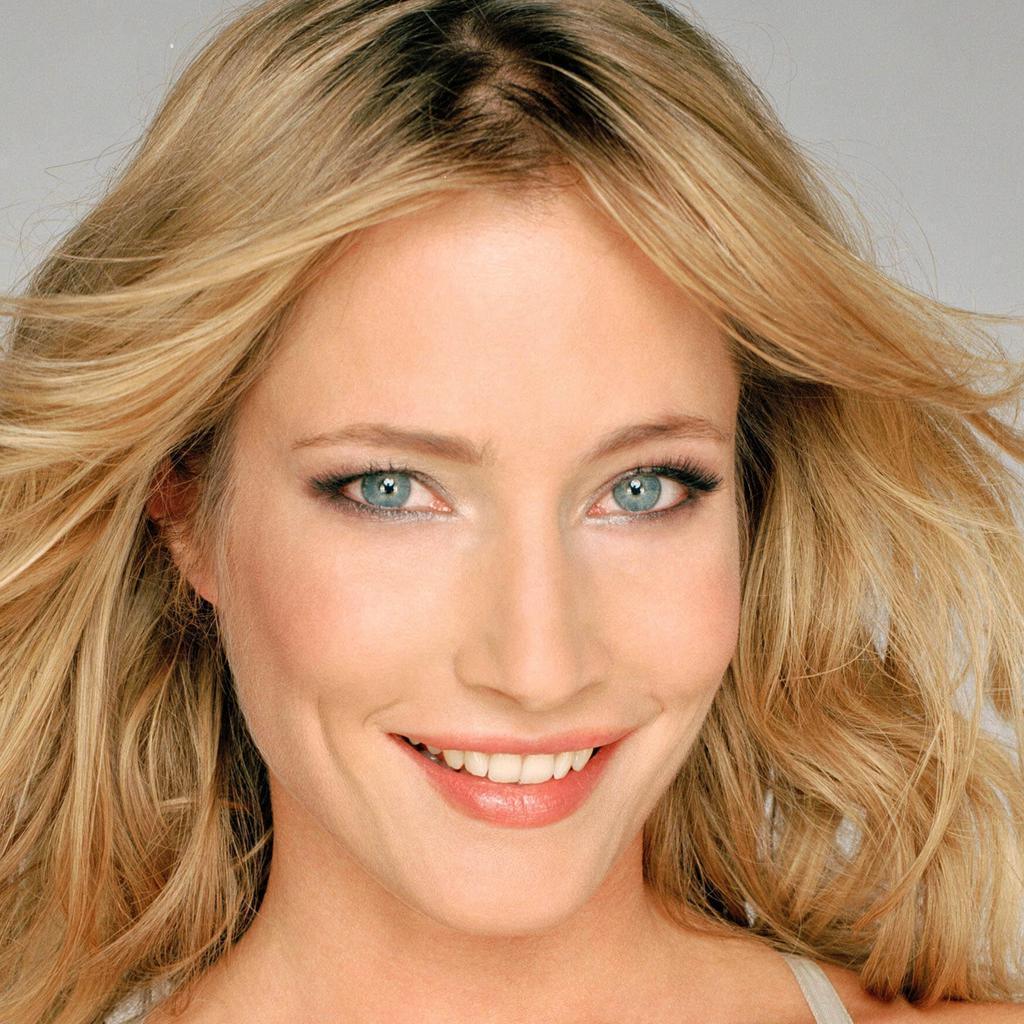} & \includegraphics[align=c,height=\hh,width=\ww,  trim=0 0 0 0,clip]{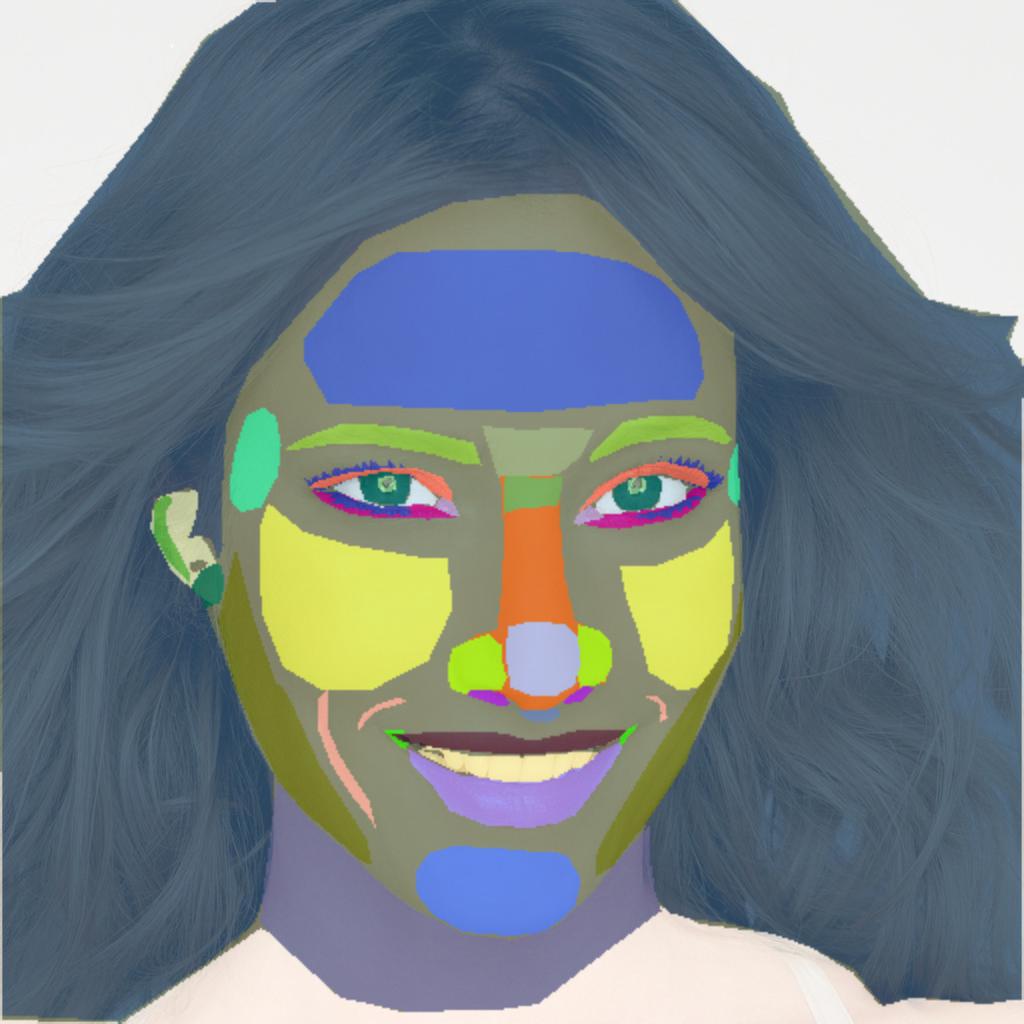} & \includegraphics[align=c,height=\hh,width=\ww,  trim=0 0 0 0,clip]{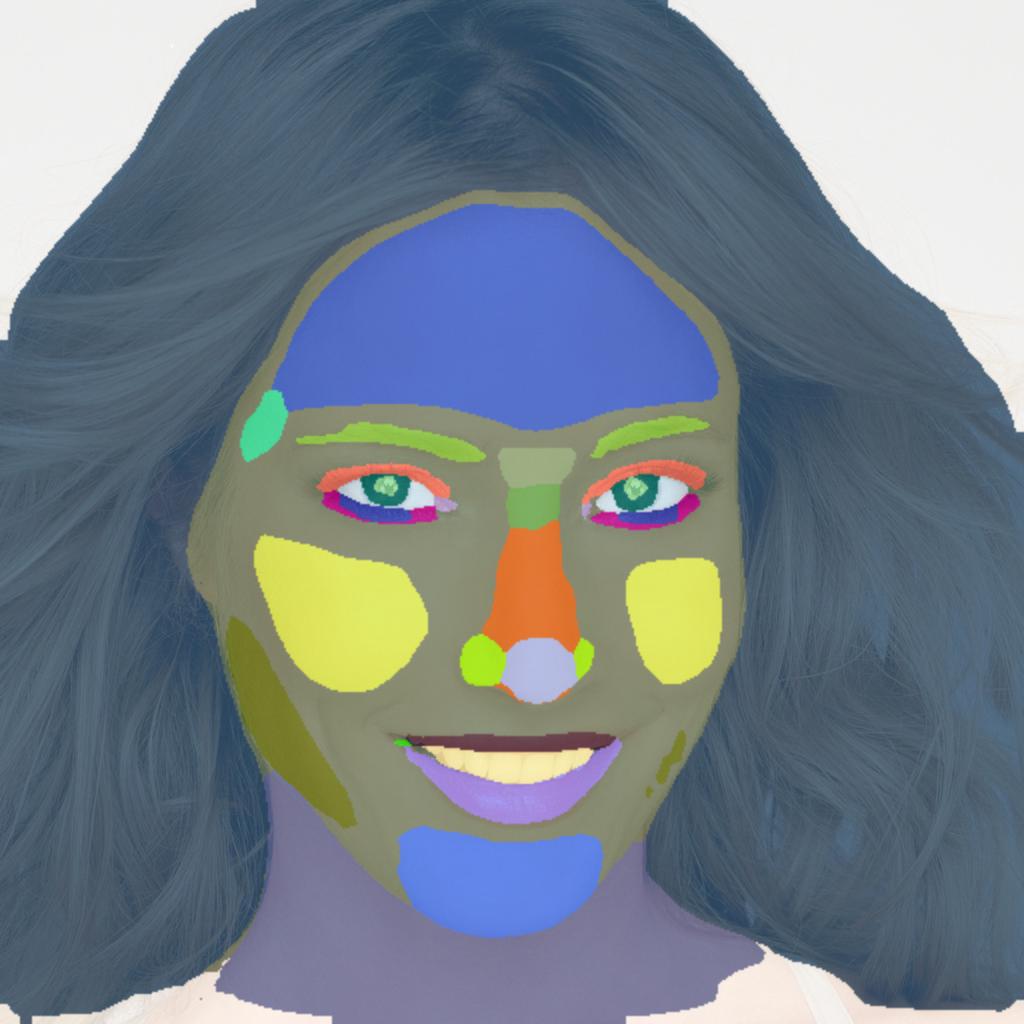} &
\includegraphics[align=c,height=\hh,width=\ww,  trim=0 0 0 0,clip]{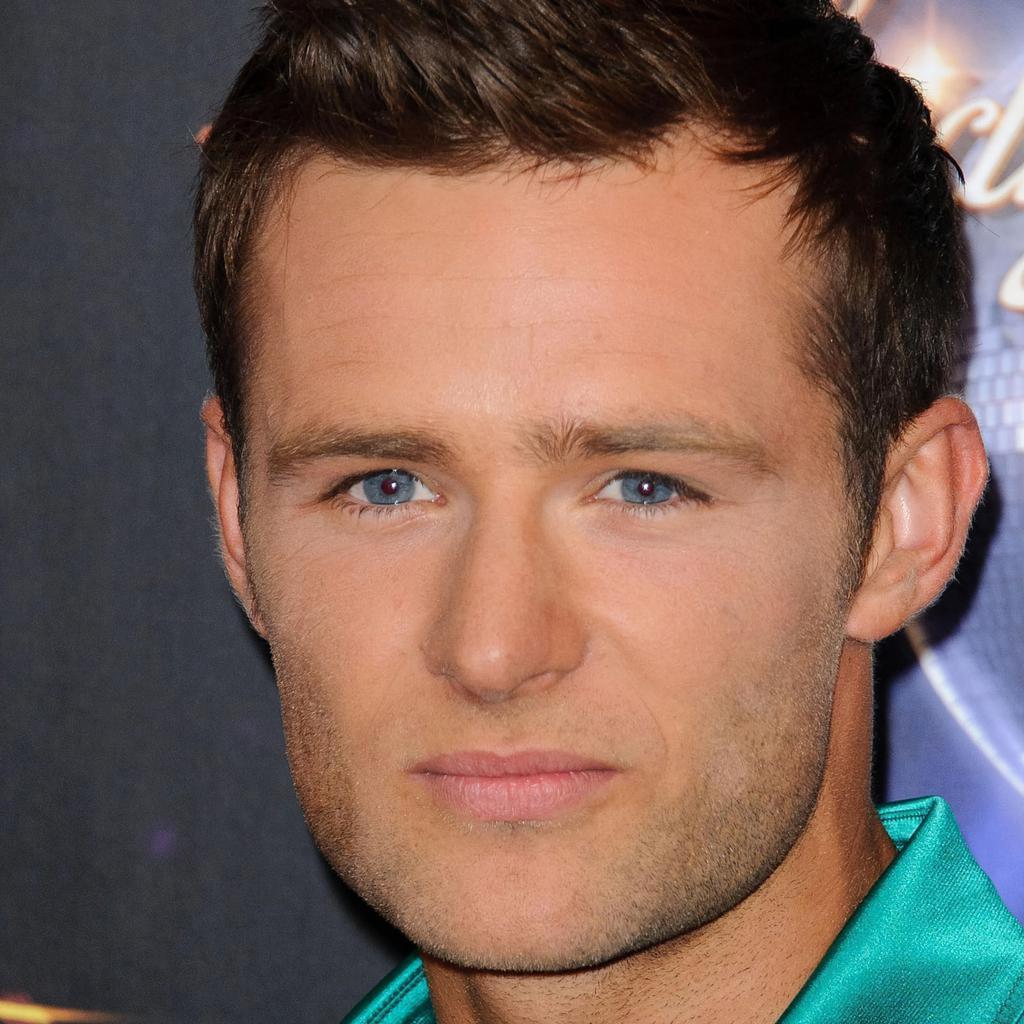} & \includegraphics[align=c,height=\hh,width=\ww,  trim=0 0 0 0,clip]{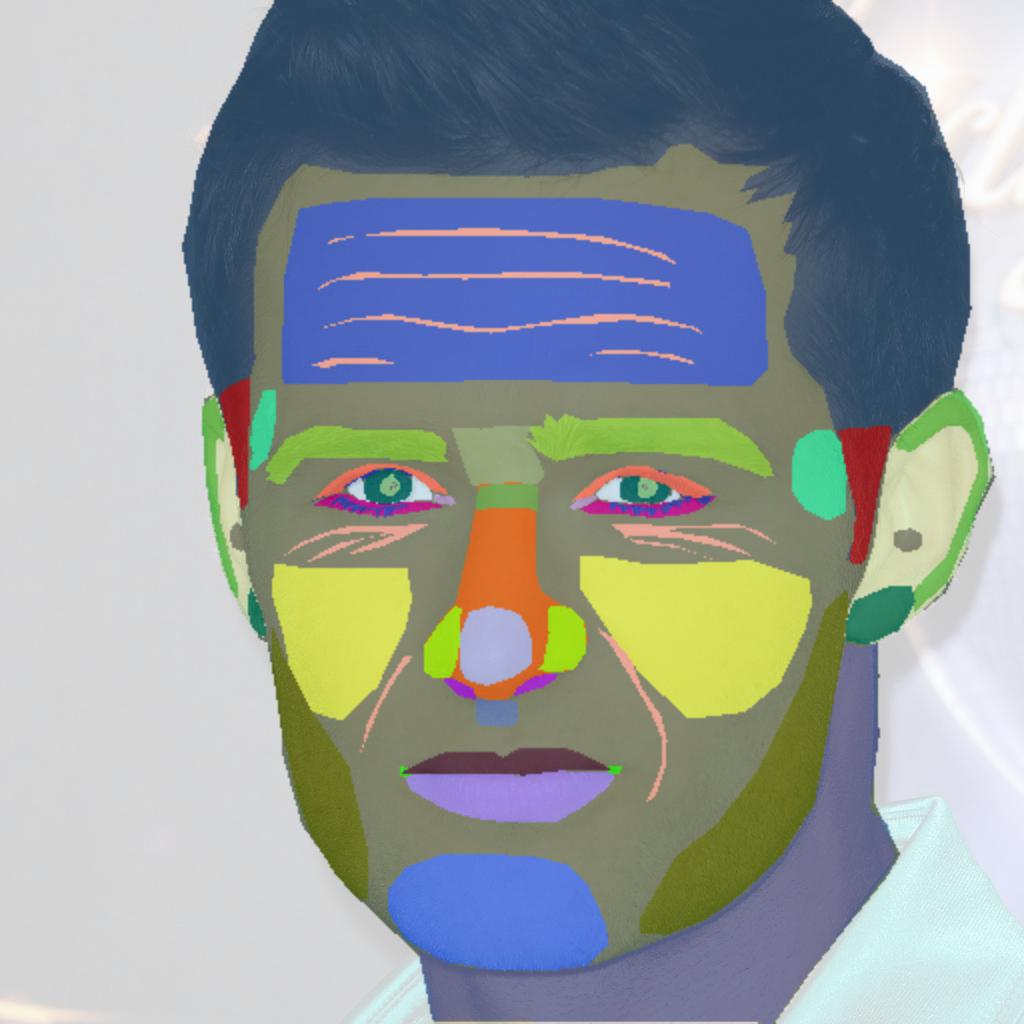} & \includegraphics[align=c,height=\hh,width=\ww,  trim=0 0 0 0,clip]{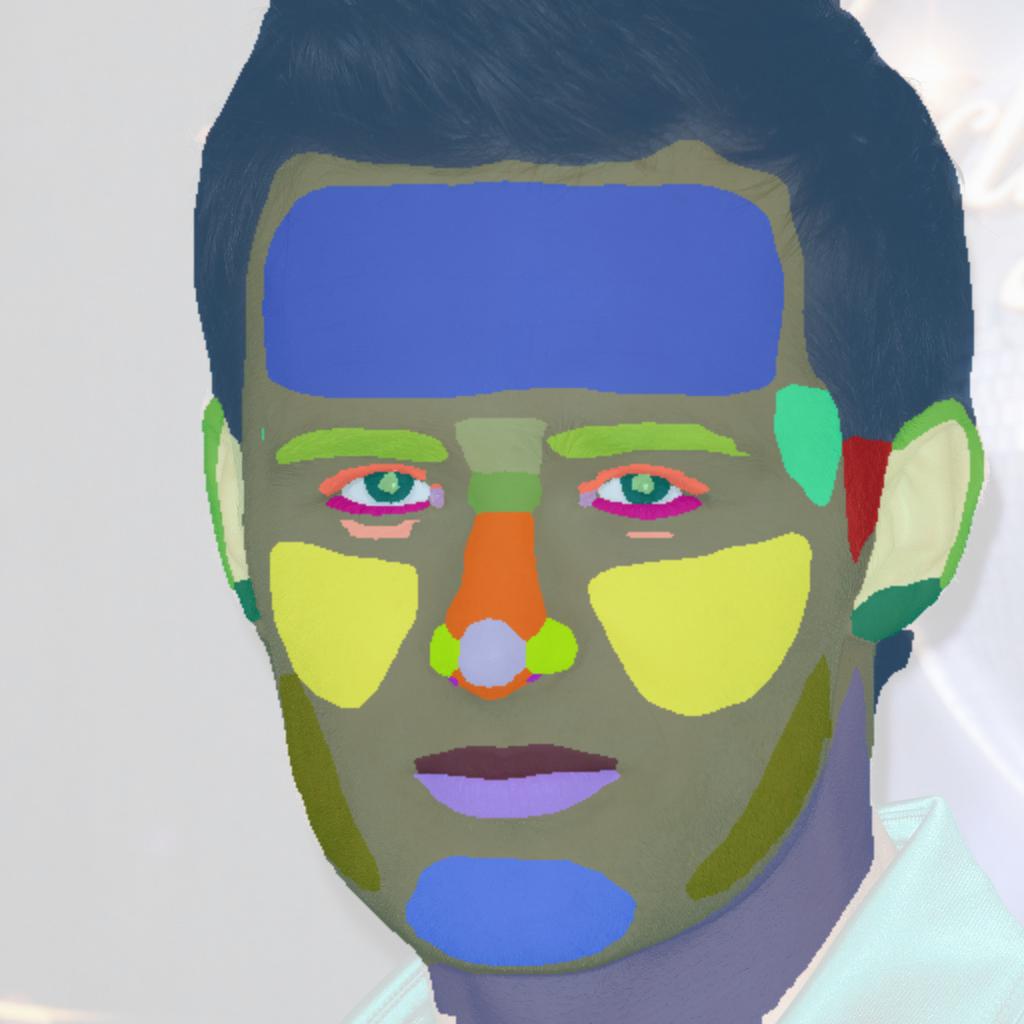}\\
{\scriptsize {\bf Faces} $\qquad$ 34  cls.} &
\includegraphics[align=c,height=\hh,width=\ww, trim=0 0 0 0,clip]{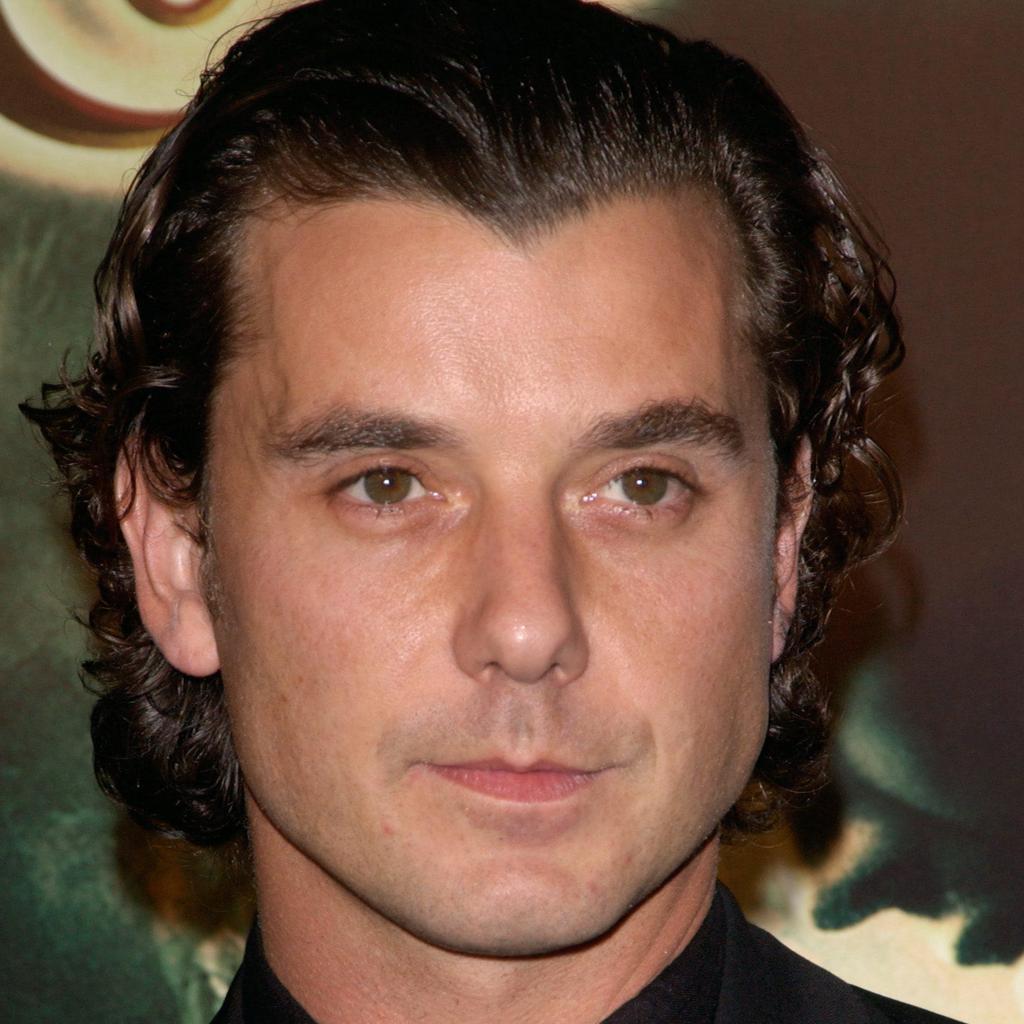} & \includegraphics[align=c,height=\hh,width=\ww, trim=0 0 0 0,clip]{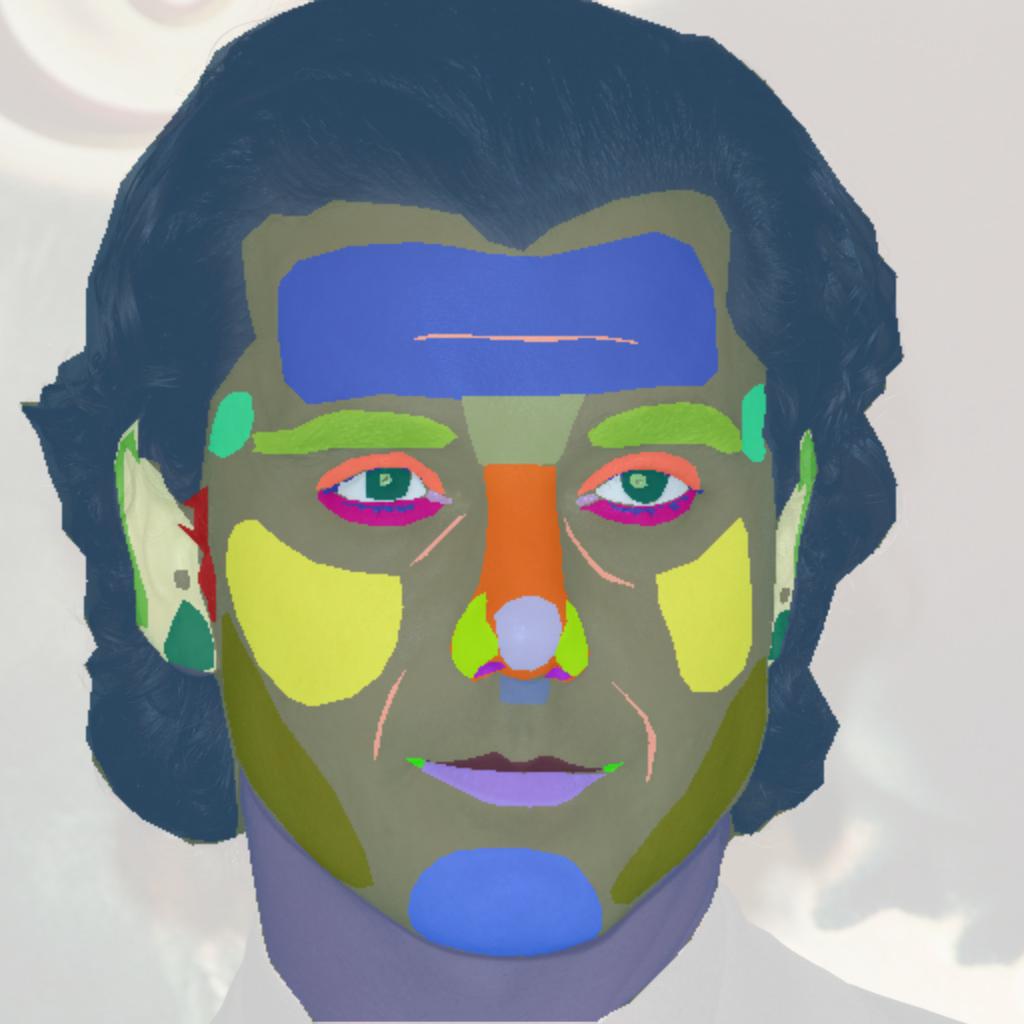} & \includegraphics[align=c,height=\hh,width=\ww,  trim=0 0 0 0,clip]{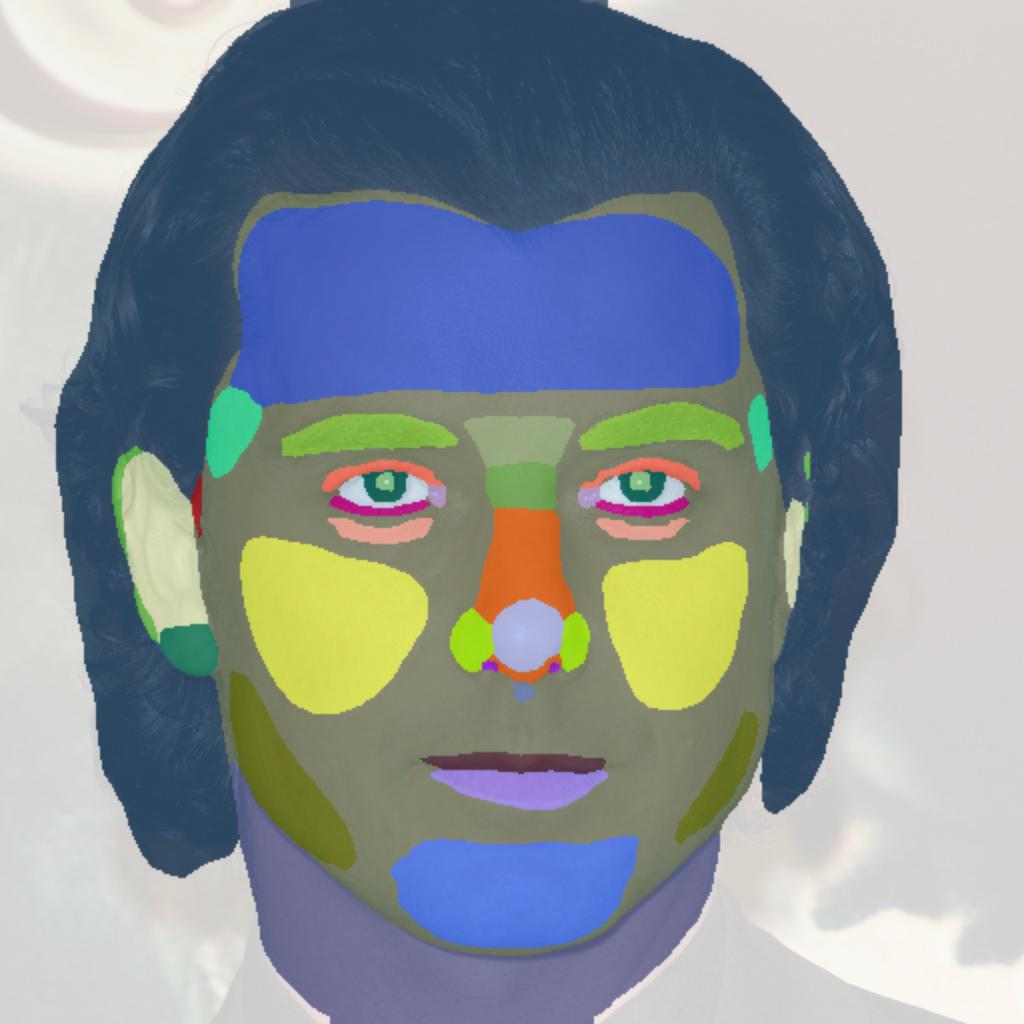} &
\includegraphics[align=c,height=\hh,width=\ww,  trim=0 0 0 0,clip]{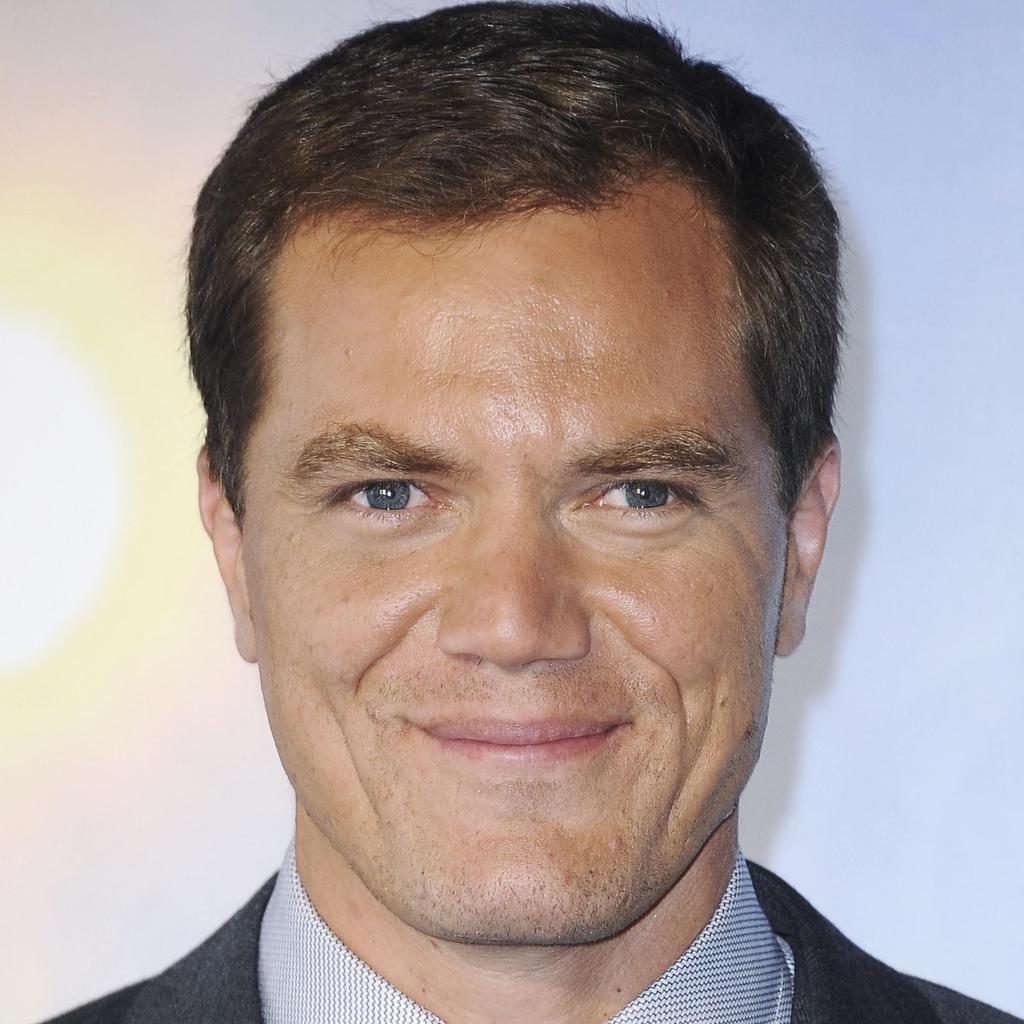} & \includegraphics[align=c,height=\hh,width=\ww,  trim=0 0 0 0,clip]{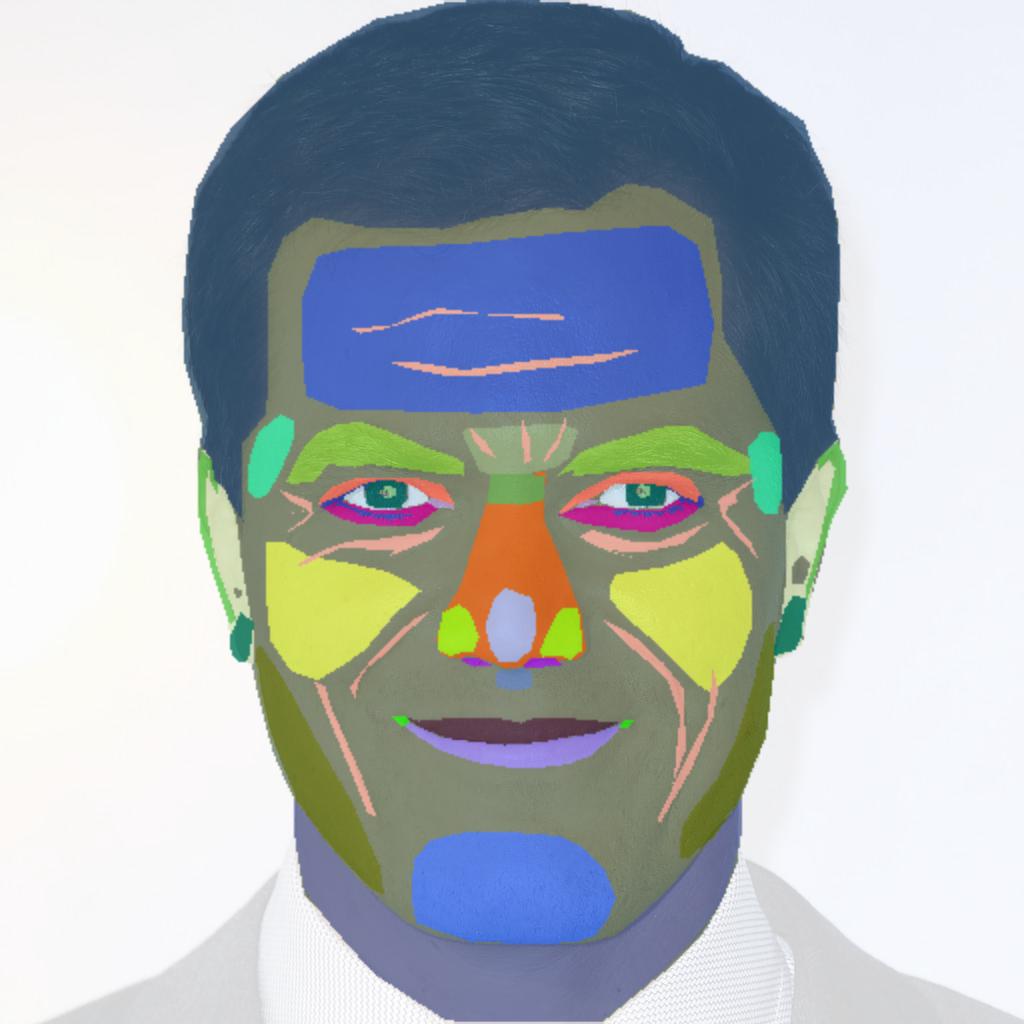} & \includegraphics[align=c,height=\hh,width=\ww,  trim=0 0 0 0,clip]{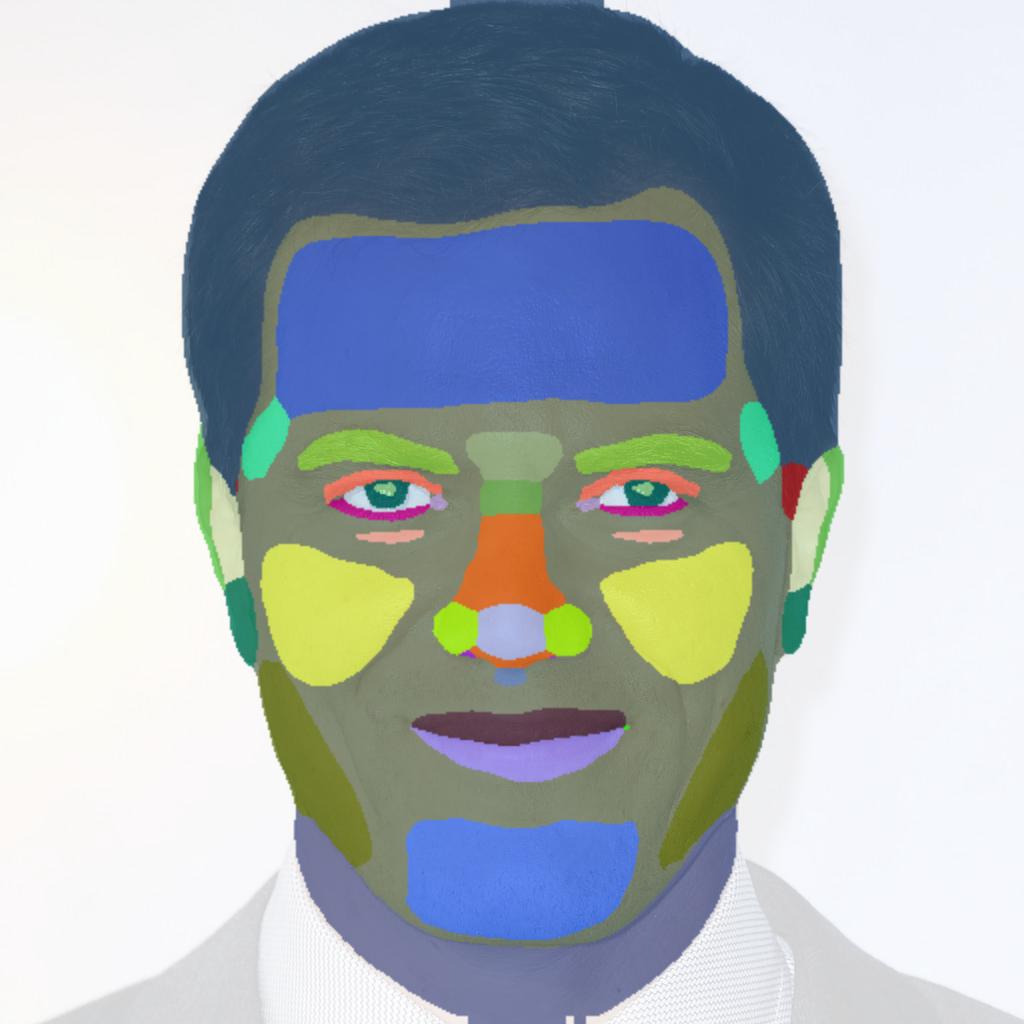} &
\includegraphics[align=c,height=\hh,width=\ww,  trim=0 0 0 0,clip]{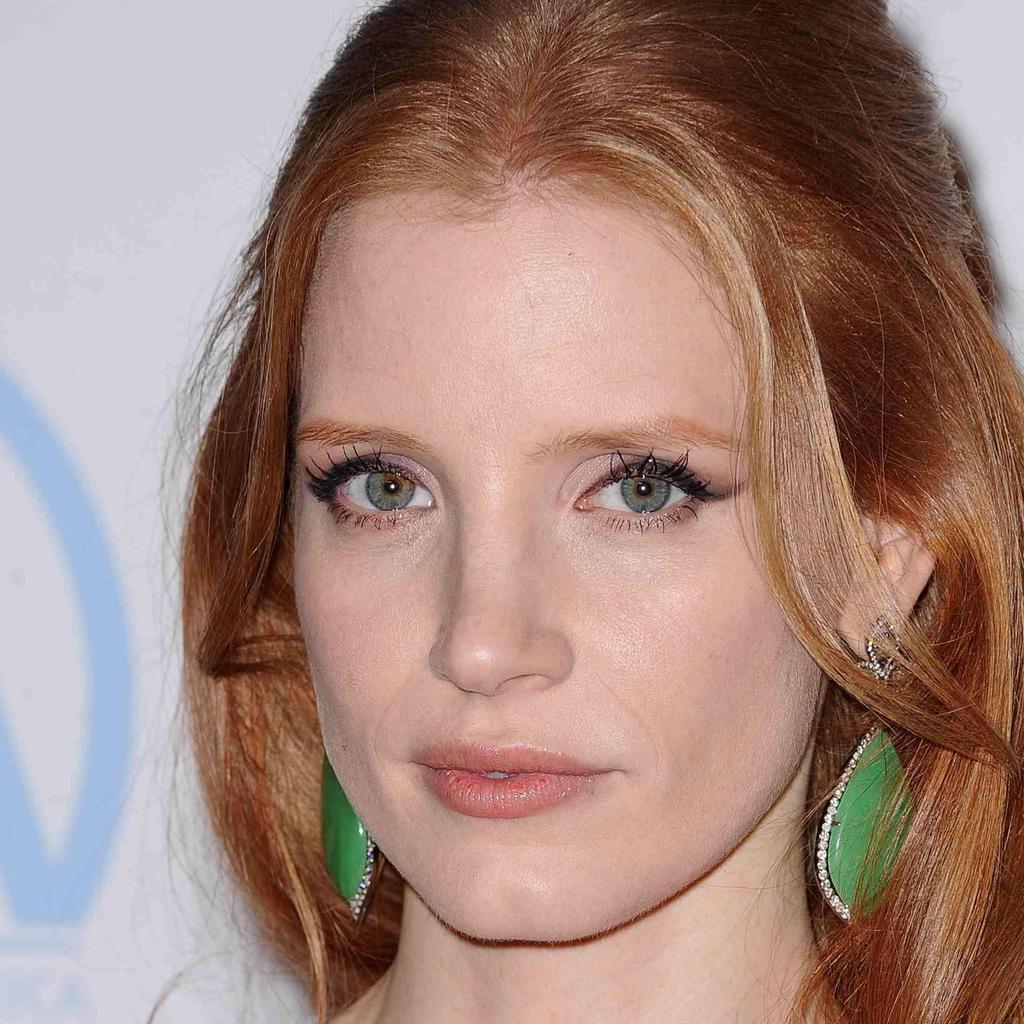} & \includegraphics[align=c,height=\hh,width=\ww,  trim=0 0 0 0,clip]{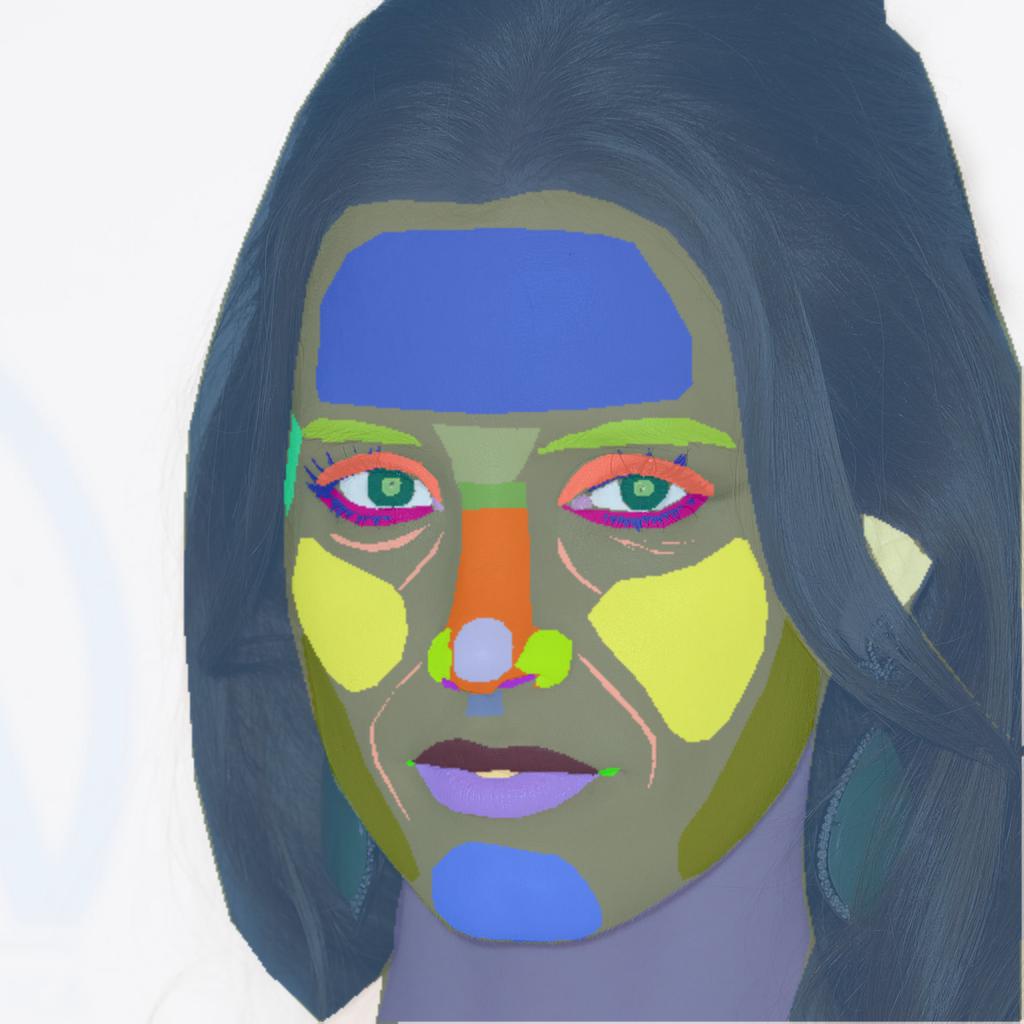} & \includegraphics[align=c,height=\hh,width=\ww,  trim=0 0 0 0,clip]{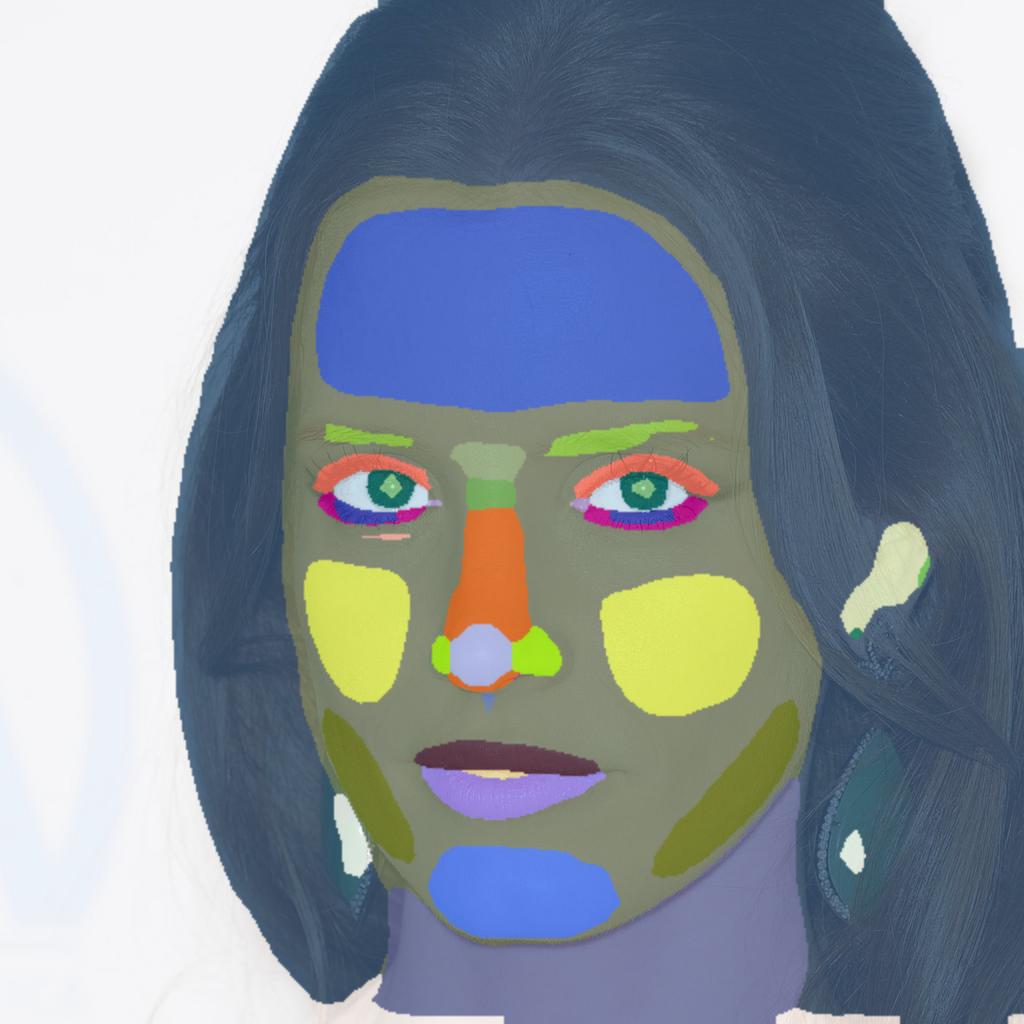} &
\includegraphics[align=c,height=\hh,width=\ww,  trim=0 0 0 0,clip]{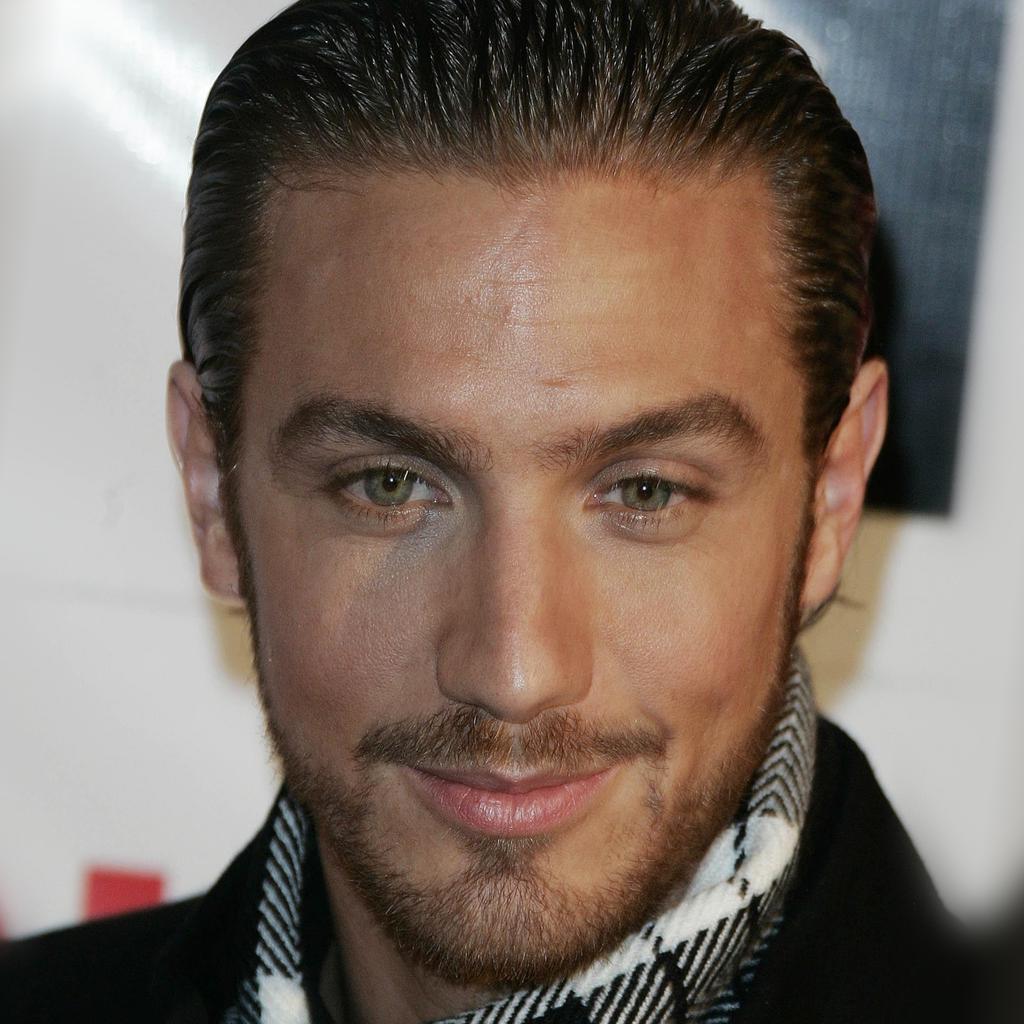} & \includegraphics[align=c,height=\hh,width=\ww,  trim=0 0 0 0,clip]{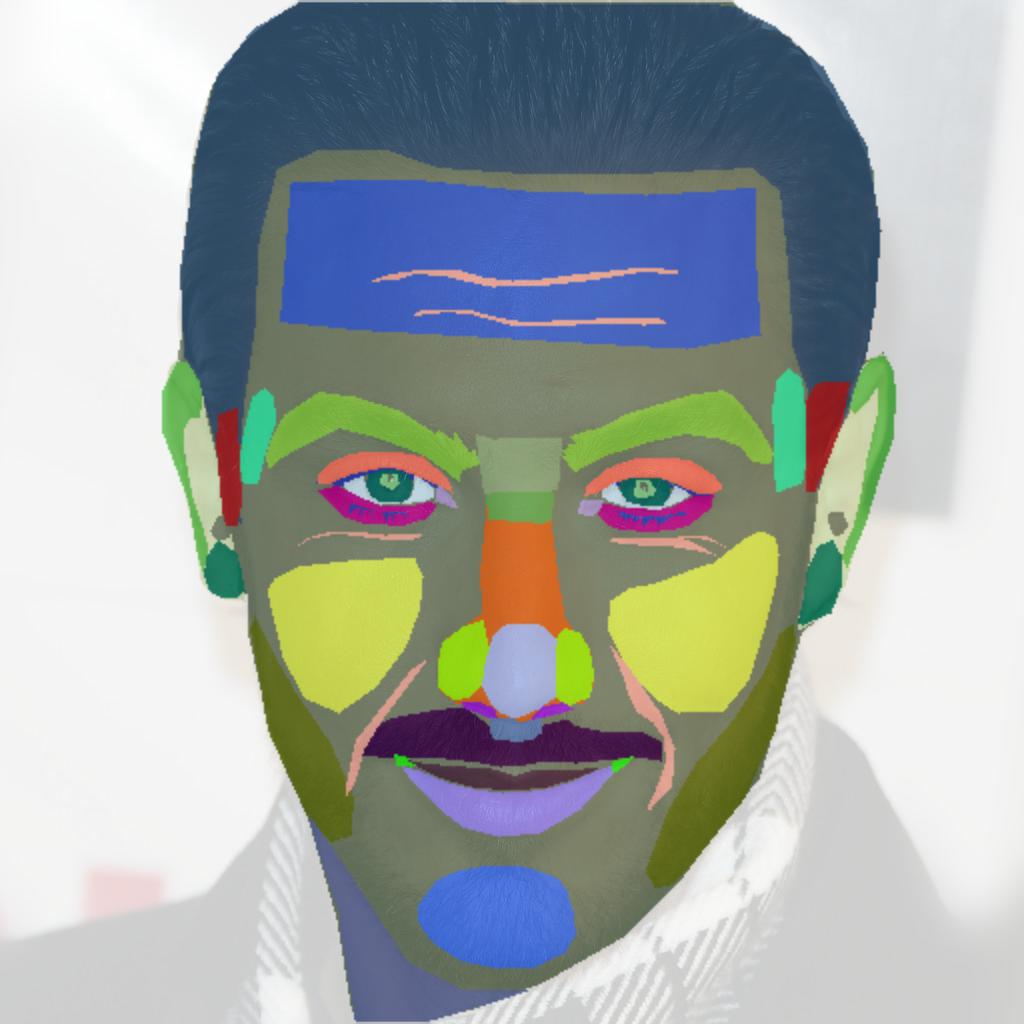} & \includegraphics[align=c,height=\hh,width=\ww,  trim=0 0 0 0,clip]{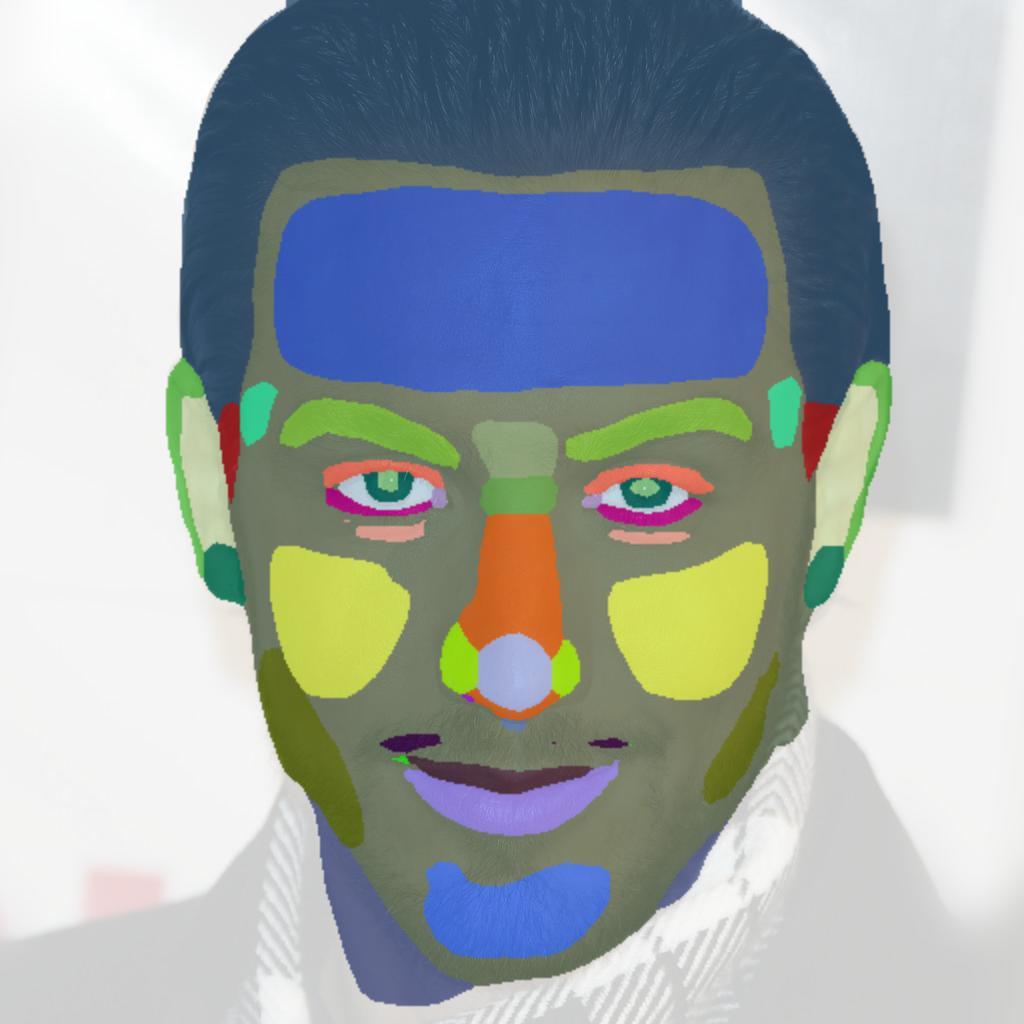}\\
{\scriptsize {\bf Birds} $\qquad$ 11  cls.} &
\includegraphics[align=c,height=\hh,width=\ww, trim=0 0 0 0,clip]{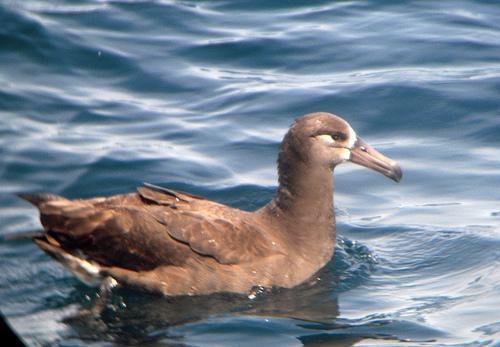} & \includegraphics[align=c,height=\hh,width=\ww, trim=0 0 0 0,clip]{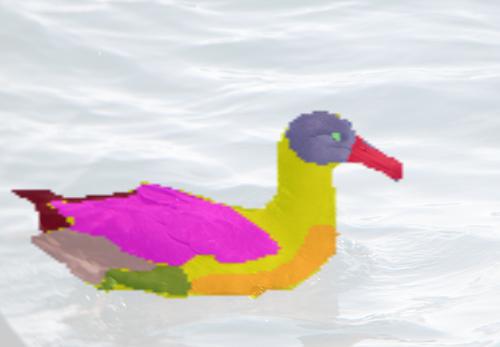} & \includegraphics[align=c,height=\hh,width=\ww,  trim=0 0 0 0,clip]{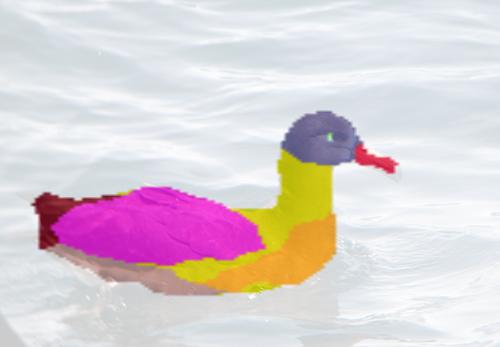} &
\includegraphics[align=c,height=\hh,width=\ww,  trim=0 0 0 50,clip]{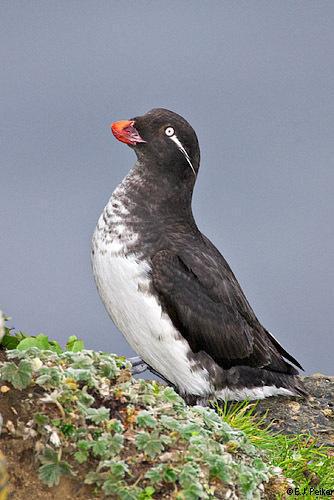} & \includegraphics[align=c,height=\hh,width=\ww,  trim=0 0 0 50,clip]{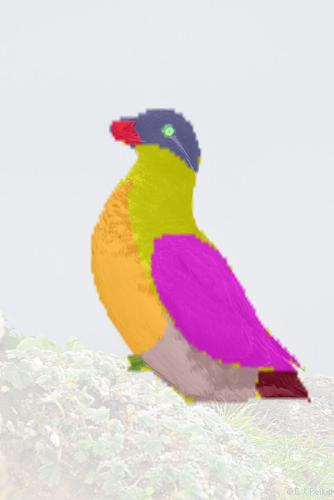} & \includegraphics[align=c,height=\hh,width=\ww,  trim=0 0 0 50,clip]{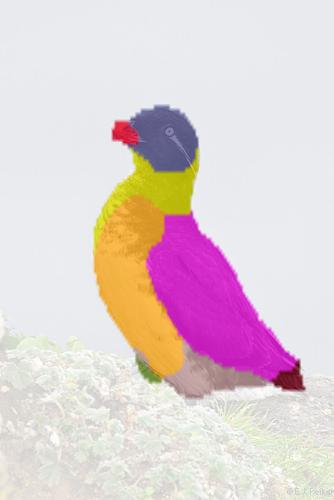} &
\includegraphics[align=c,height=\hh,width=\ww,  trim=0 0 0 0,clip]{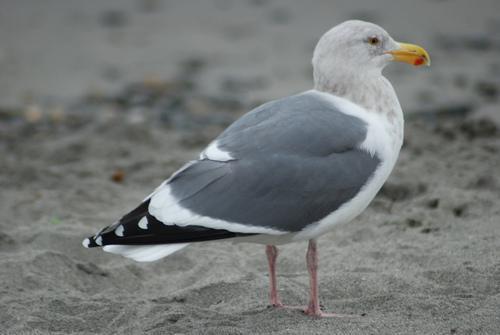} & \includegraphics[align=c,height=\hh,width=\ww,  trim=0 0 0 0,clip]{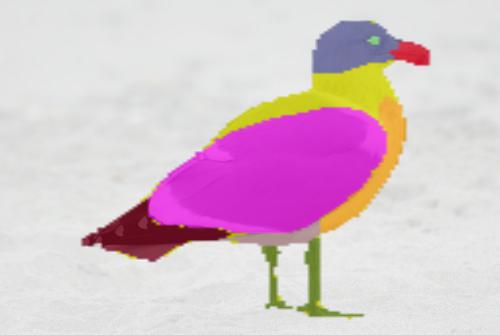} & \includegraphics[align=c,height=\hh,width=\ww,  trim=0 0 0 0,clip]{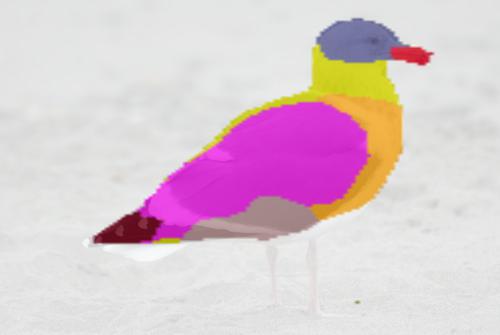} &
\includegraphics[align=c,height=\hh,width=\ww,  trim=0 0 0 0,clip]{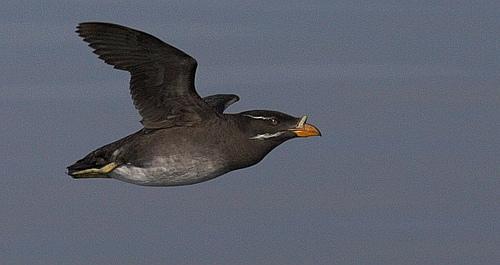} & \includegraphics[align=c,height=\hh,width=\ww,  trim=0 0 0 0,clip]{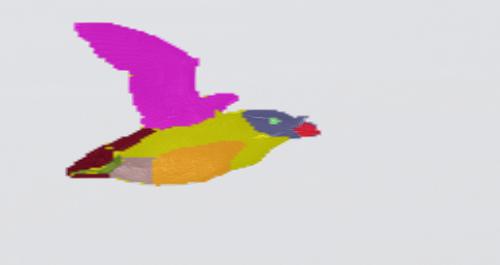} & \includegraphics[align=c,height=\hh,width=\ww,  trim=0 0 0 0,clip]{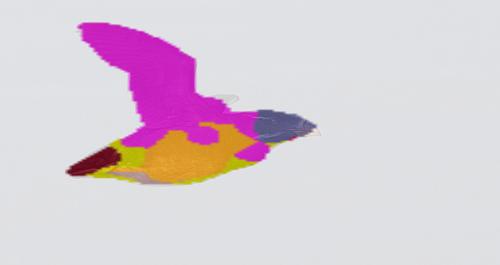}\\
{\scriptsize {\bf Cats} $\qquad$ 16  cls.} &
\includegraphics[align=c,height=\hh,width=\ww, trim=0 0 0 0,clip]{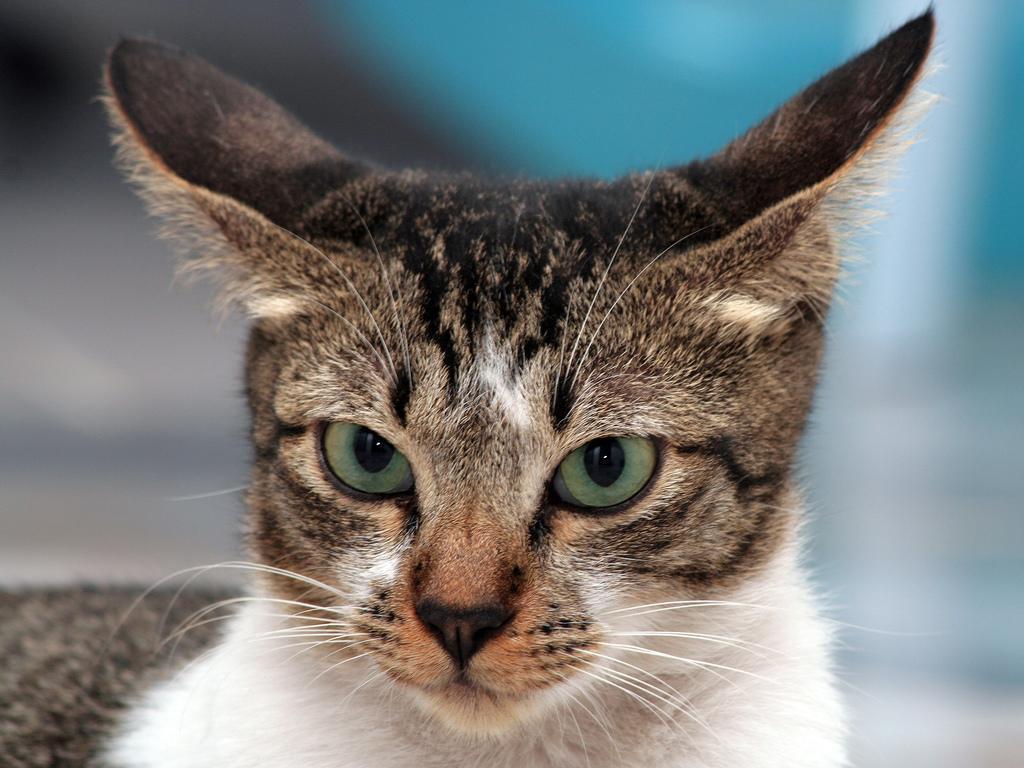} & \includegraphics[align=c,height=\hh,width=\ww, trim=0 0 0 0,clip]{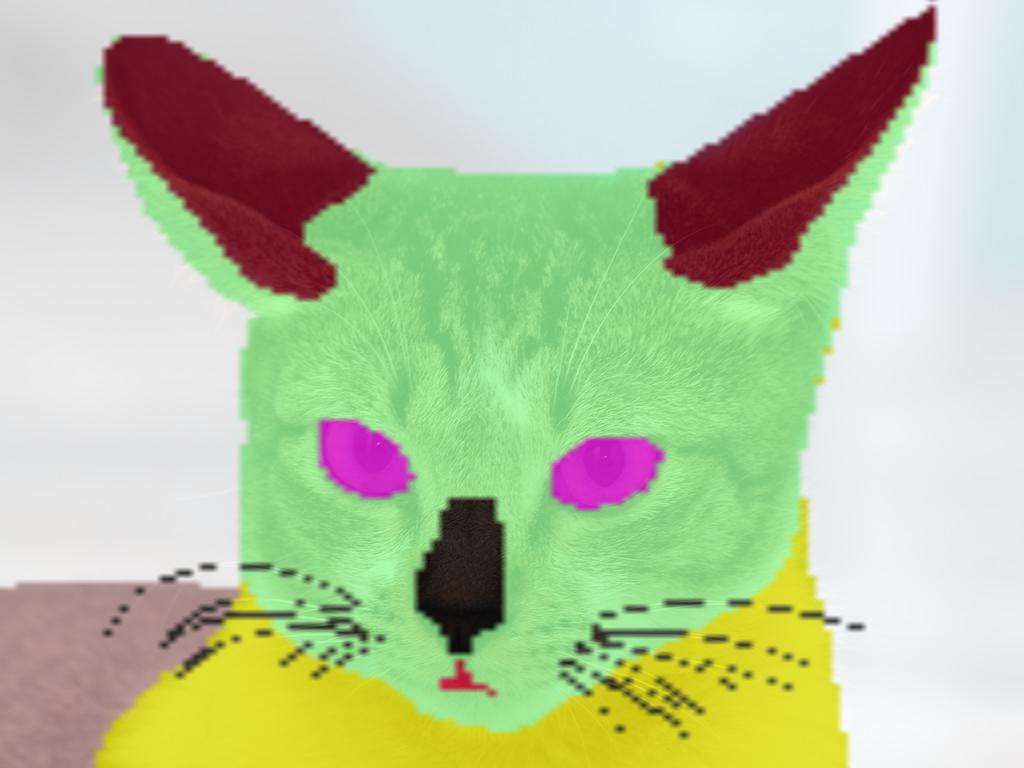} & \includegraphics[align=c,height=\hh,width=\ww,  trim=0 0 0 0,clip]{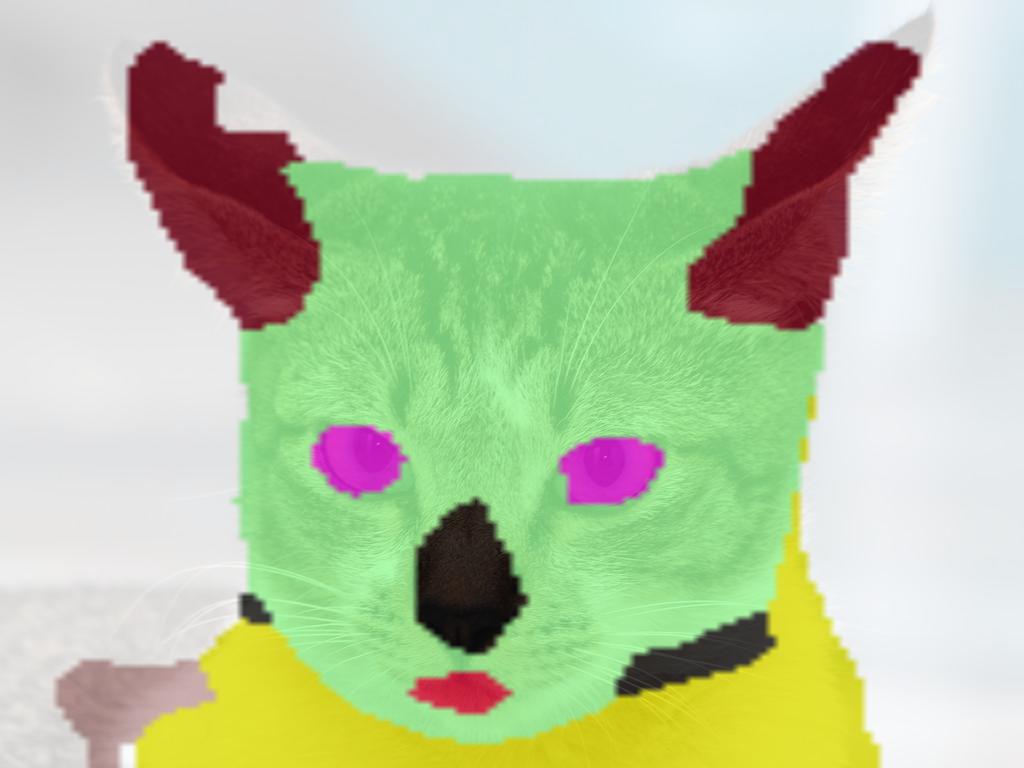} &
\includegraphics[align=c,height=\hh,width=\ww,  trim=0 0 0 0,clip]{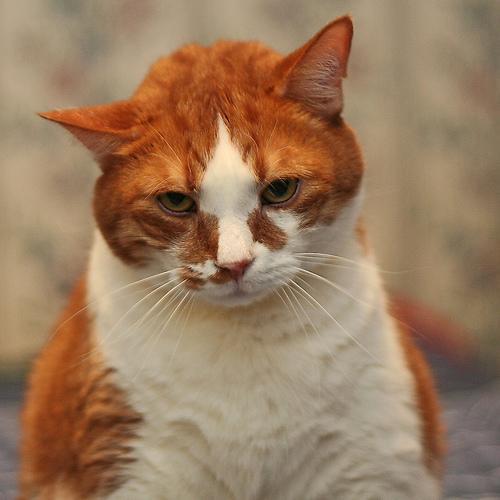} & \includegraphics[align=c,height=\hh,width=\ww,  trim=0 0 0 0,clip]{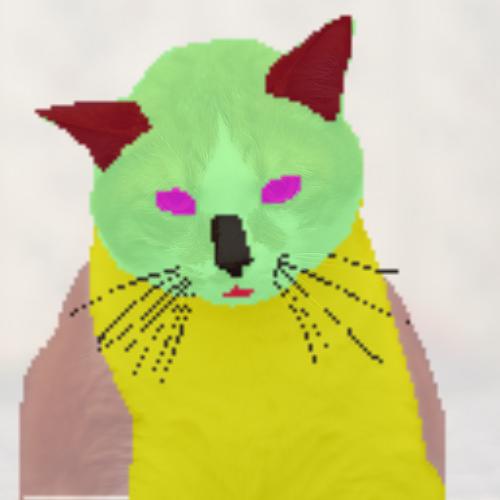} & \includegraphics[align=c,height=\hh,width=\ww,  trim=0 0 0 0,clip]{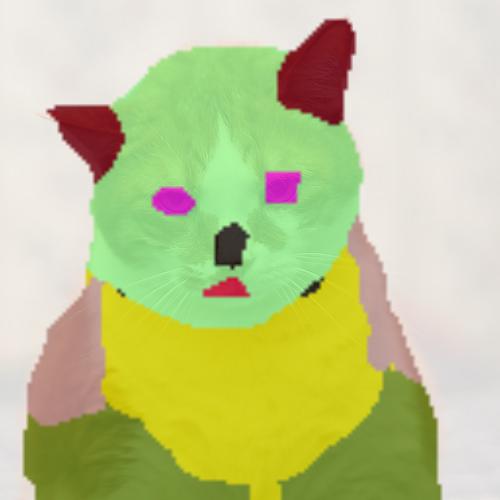} &
\includegraphics[align=c,height=\hh,width=\ww,  trim=0 0 0 0,clip]{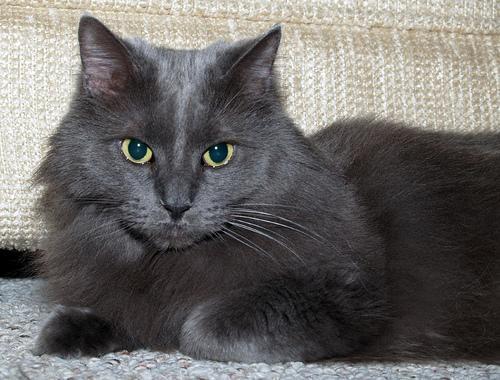} & \includegraphics[align=c,height=\hh,width=\ww,  trim=0 0 0 0,clip]{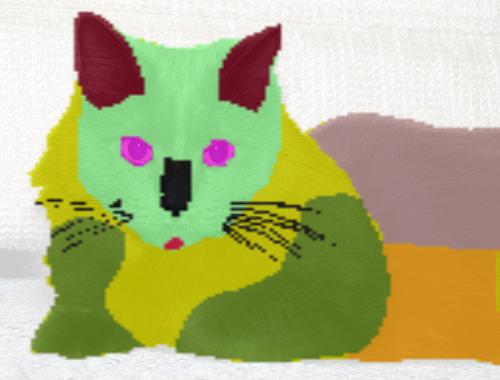} & \includegraphics[align=c,height=\hh,width=\ww,  trim=0 0 0 0,clip]{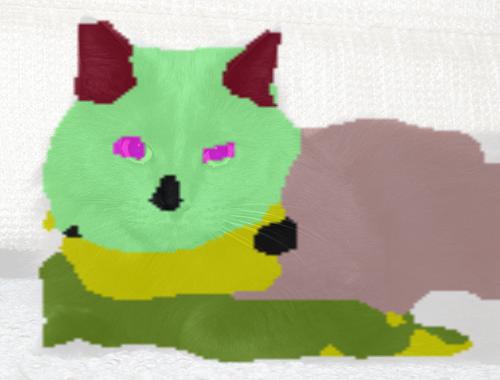} &
\includegraphics[align=c,height=\hh,width=\ww,  trim=0 0 0 0,clip]{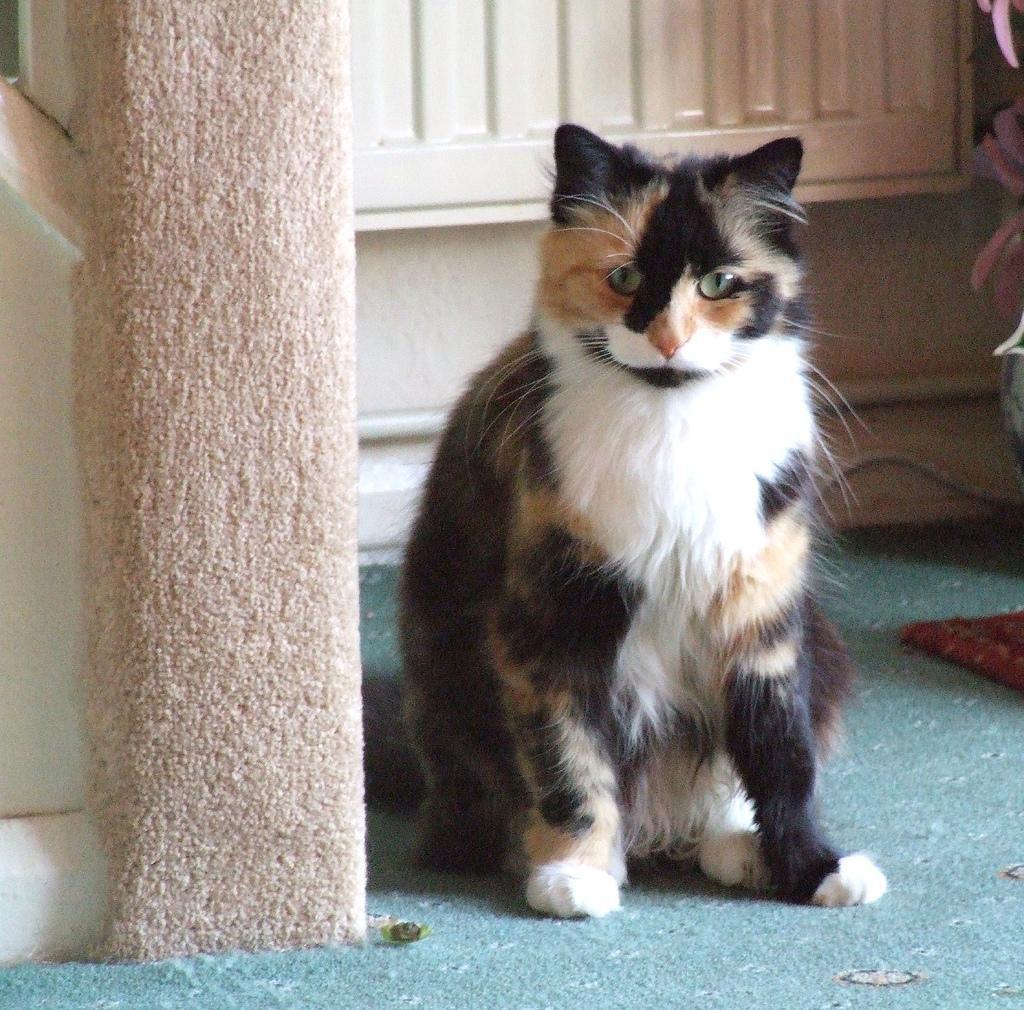} & \includegraphics[align=c,height=\hh,width=\ww,  trim=0 0 0 0,clip]{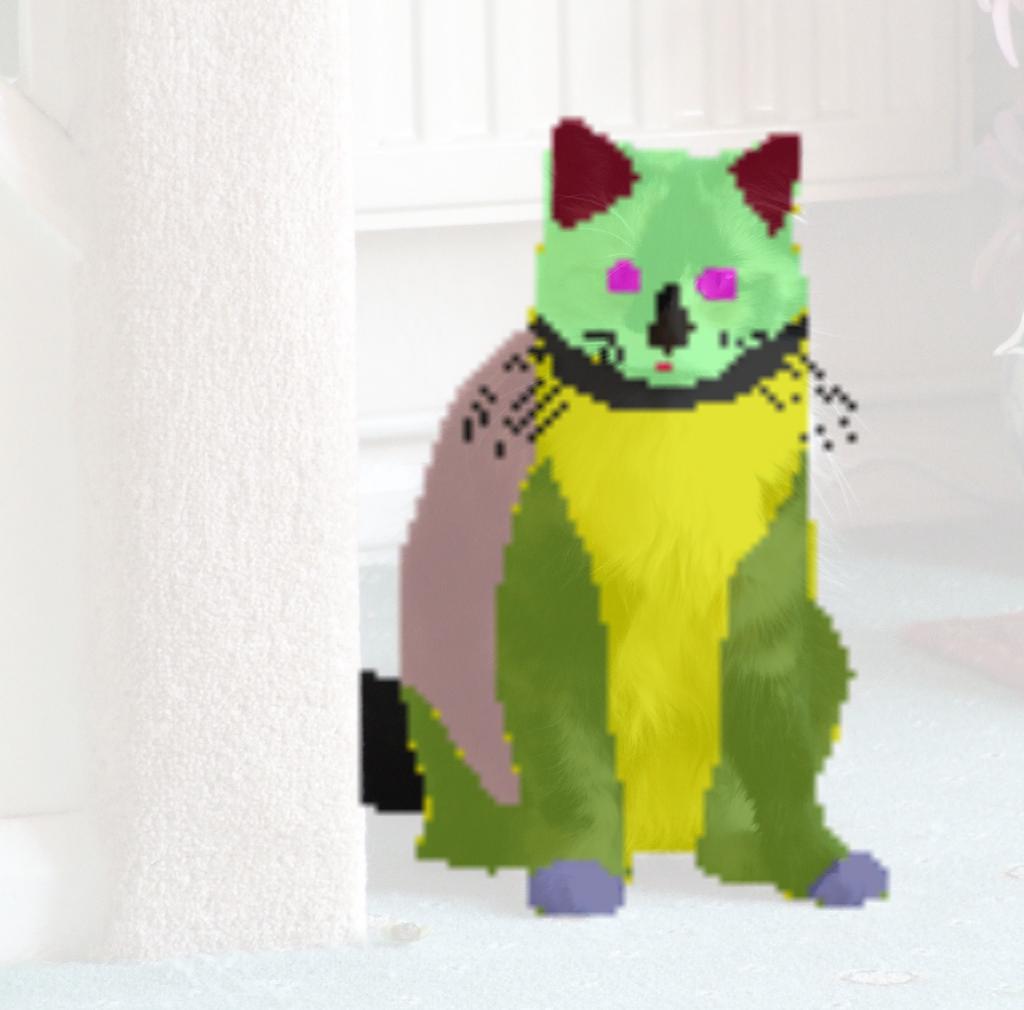} & \includegraphics[align=c,height=\hh,width=\ww,  trim=0 0 0 0,clip]{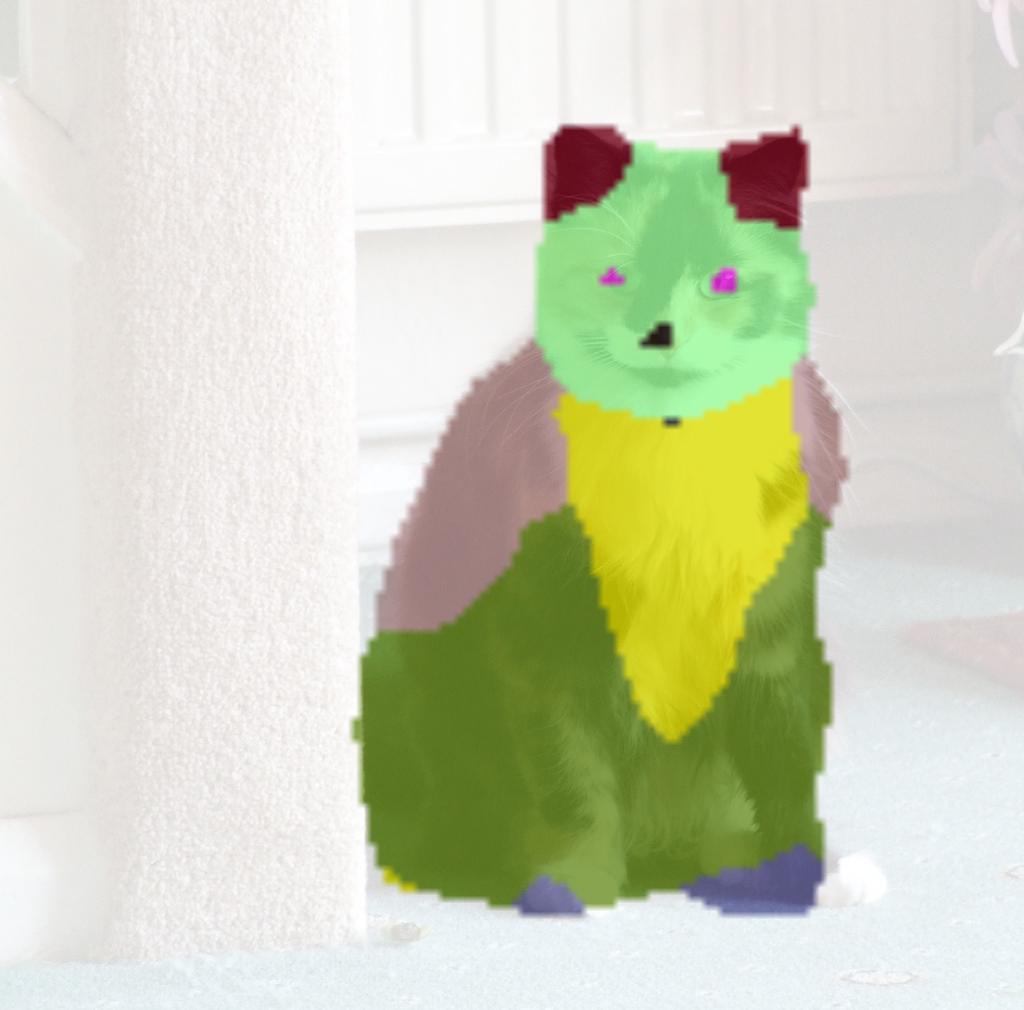} \\[-1mm]

{\scriptsize {\bf Bedrooms} $\quad$ 19  cls.} &
\includegraphics[align=c,height=\hh,width=\ww, trim=0 0 0 0,clip]{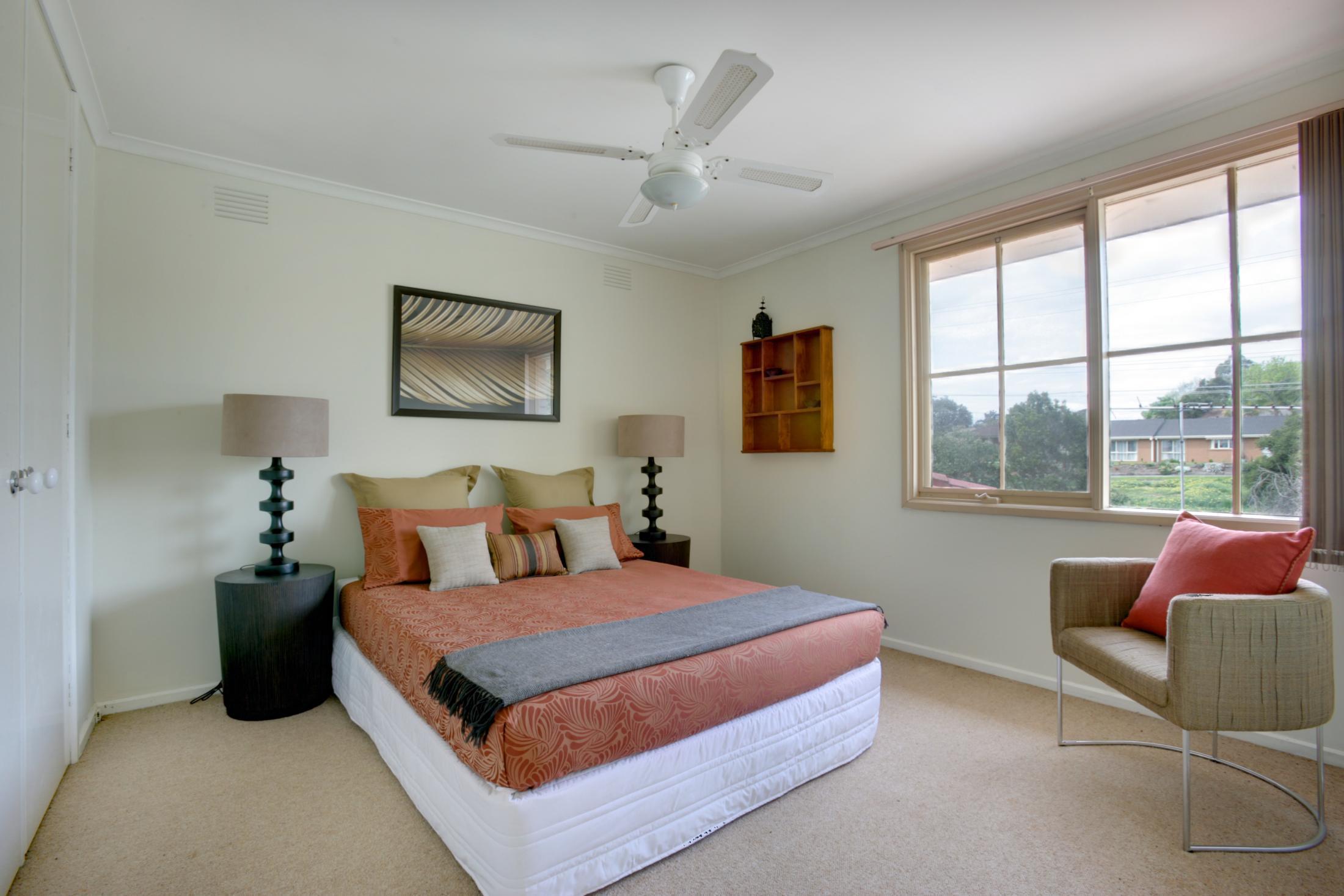} & \includegraphics[align=c,height=\hh,width=\ww, trim=0 0 0 0,clip]{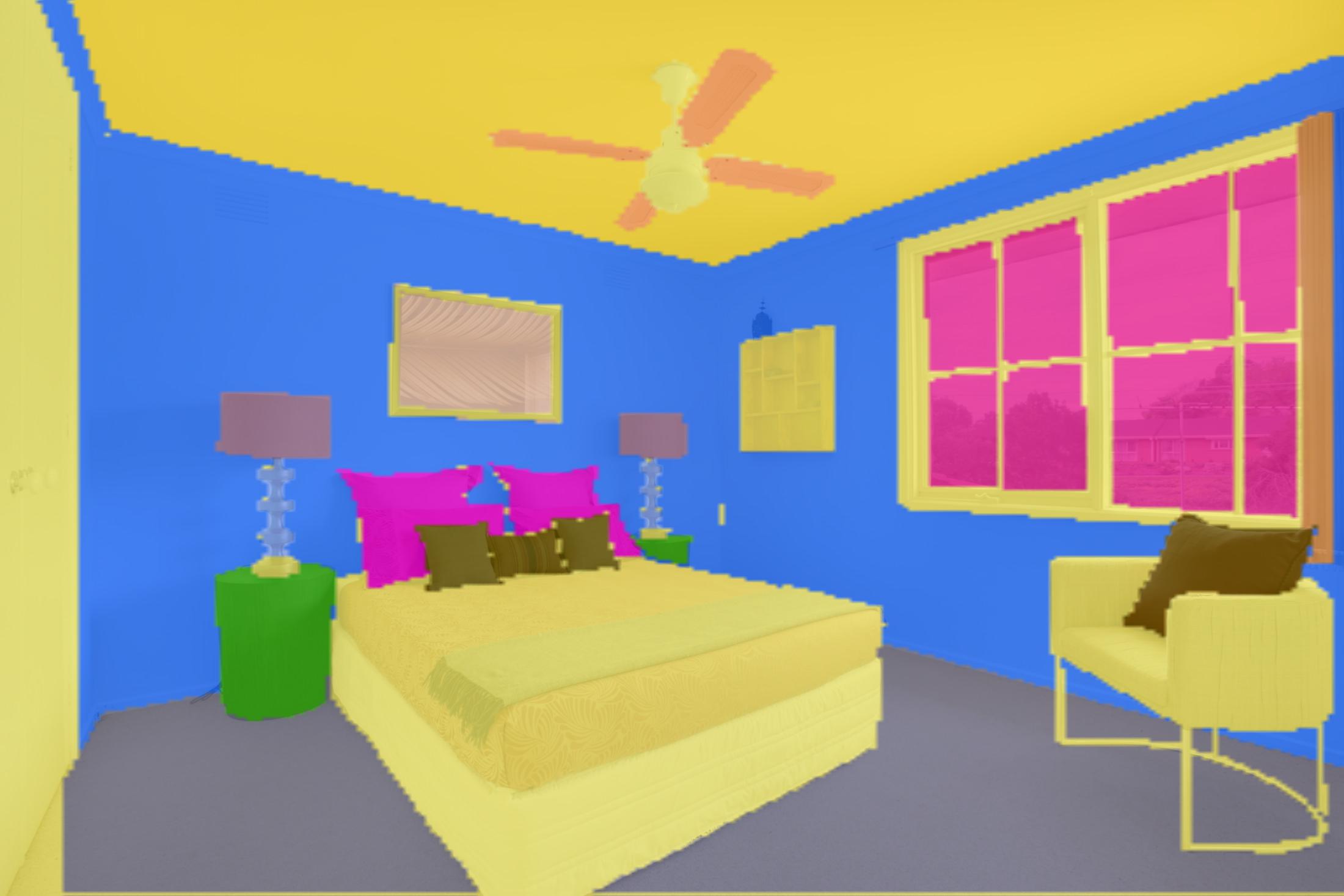} & \includegraphics[align=c,height=\hh,width=\ww,  trim=0 0 0 0,clip]{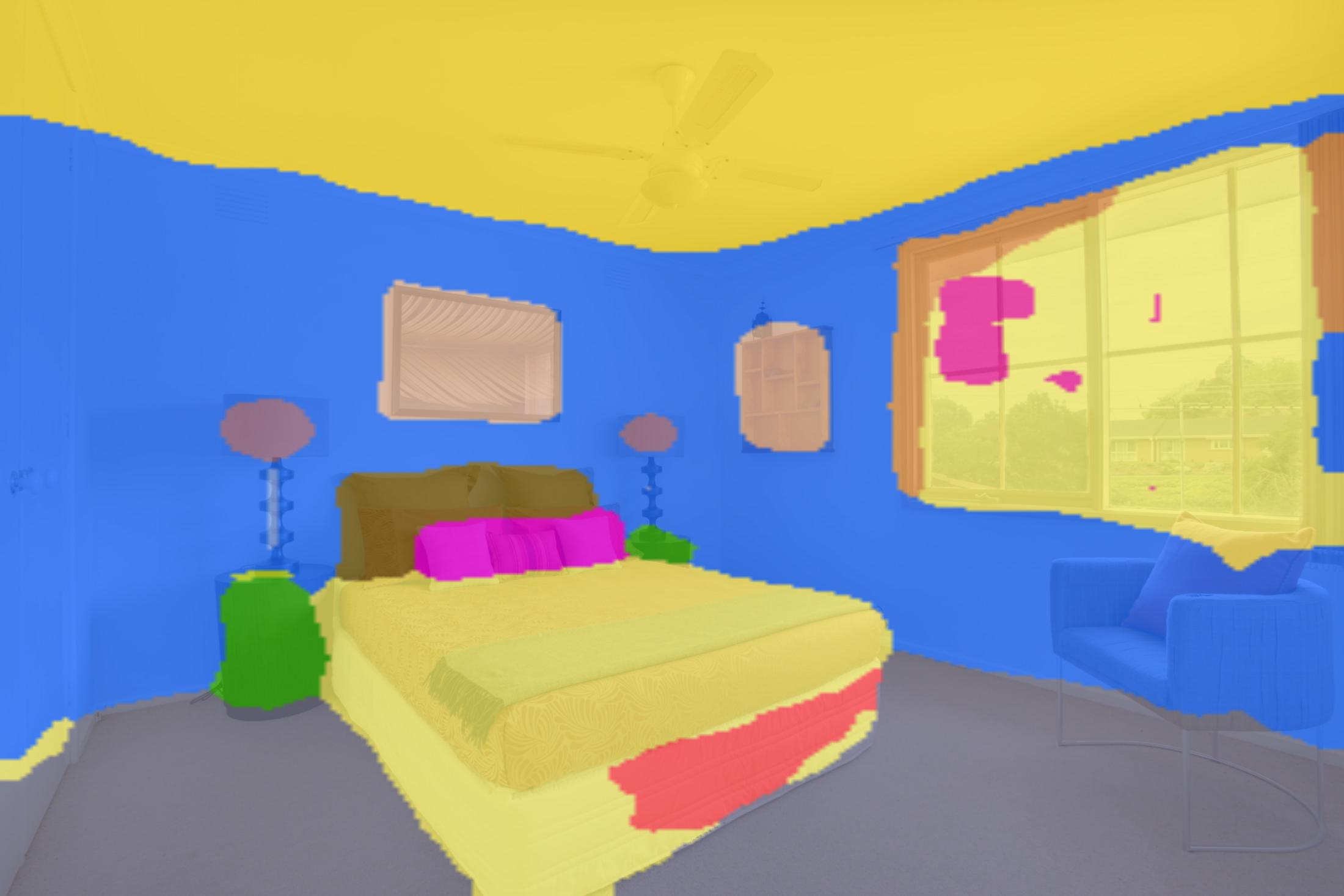} &
\includegraphics[align=c,height=\hh,width=\ww,  trim=0 0 0 0,clip]{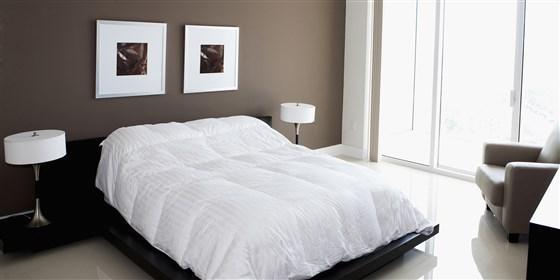} & \includegraphics[align=c,height=\hh,width=\ww,  trim=0 0 0 0,clip]{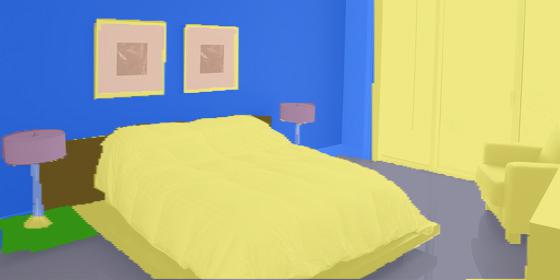} & \includegraphics[align=c,height=\hh,width=\ww,  trim=0 0 0 0,clip]{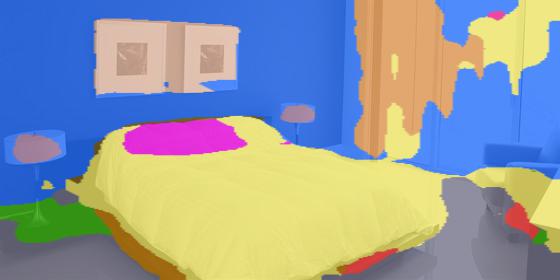} &
\includegraphics[align=c,height=\hh,width=\ww,  trim=0 0 0 0,clip]{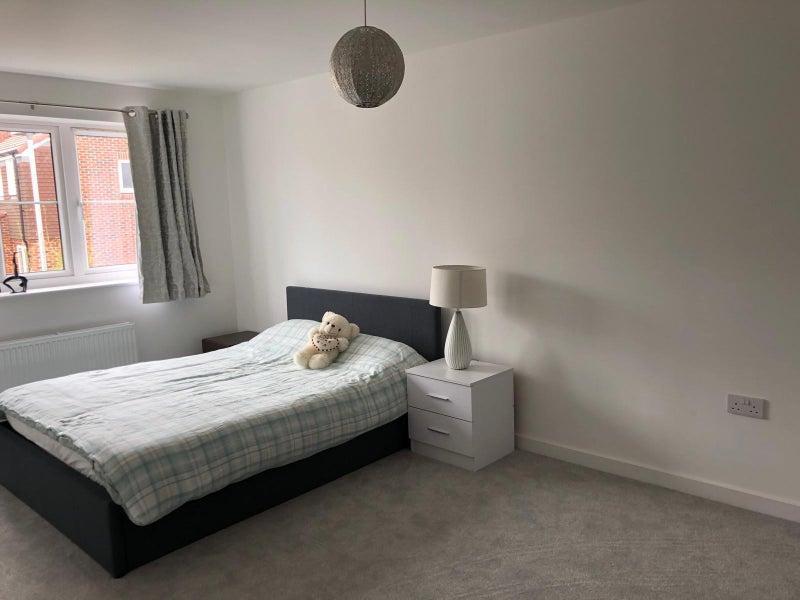} & \includegraphics[align=c,height=\hh,width=\ww,  trim=0 0 0 0,clip]{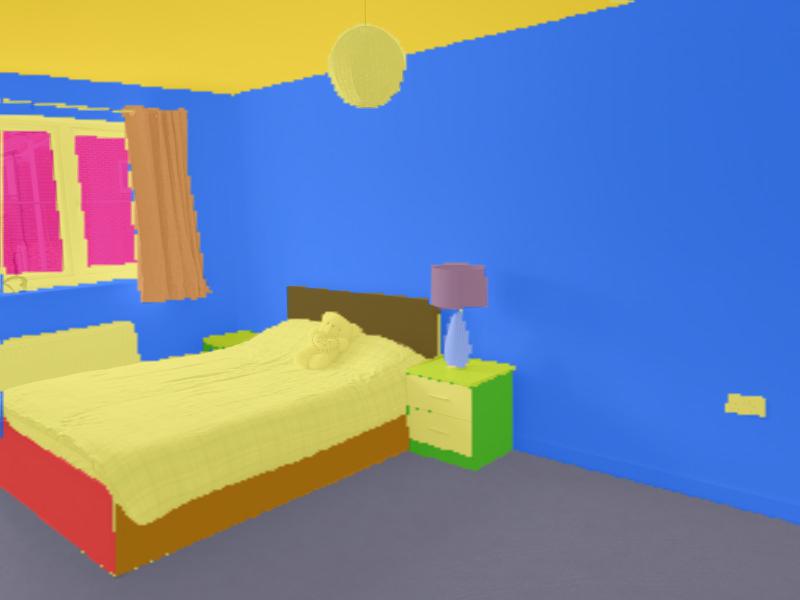} & \includegraphics[align=c,height=\hh,width=\ww,  trim=0 0 0 0,clip]{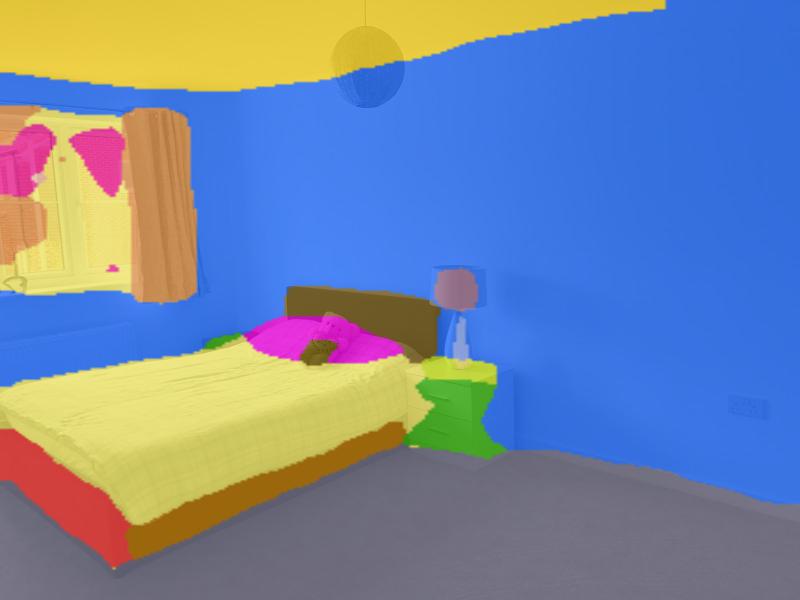} &
\includegraphics[align=c,height=\hh,width=\ww,  trim=0 0 0 0,clip]{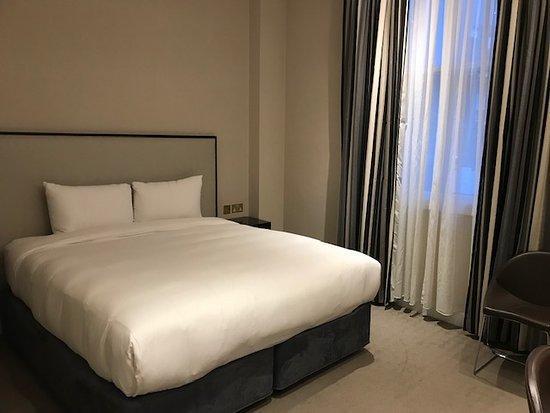} & \includegraphics[align=c,height=\hh,width=\ww,  trim=0 0 0 0,clip]{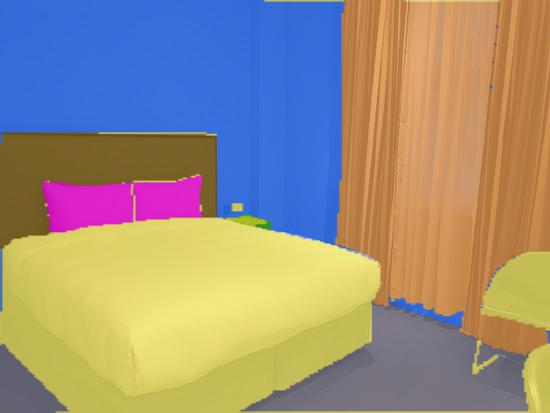} & \includegraphics[align=c,height=\hh,width=\ww,  trim=0 0 0 0,clip]{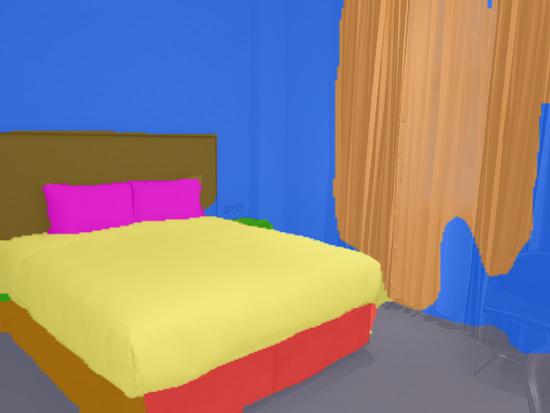} \\[-1mm]

& {\scriptsize image} & {\scriptsize groundtruth} & {\scriptsize prediction} & {\scriptsize image} & {\scriptsize groundtruth} & {\scriptsize prediction} & {\scriptsize image} & {\scriptsize groundtruth} & {\scriptsize prediction} & {\scriptsize image} & {\scriptsize groundtruth} & {\scriptsize prediction} \\
\end{tabular}
\vspace{-4mm}
\caption{\footnotesize  {\bf Qualitative Results:} We visualize predictions of DeepLab trained on {\ours}'s datasets, compared to ground-truth annotations. Typical failure cases include parts that do not have clear visual boundaries (neck of the cat), or thin structures (facial wrinkles, bird legs, cat whiskers).}
\label{fig:vis_testing}
\vspace{-2mm}
\end{figure*}

\vspace{-1.5mm}
\subsection{Keypoint Detection} 
\label{sec:kp_result}
\vspace{-1mm}
We showcase the generality of {\ours} by testing on another task, \ie keypoint detection. 

\vspace{-3.5mm}
\paragraph{Experimental Settings:}
\vspace{-1mm}
We follow the common practice of keypoint detection, i.e. predicting heatmaps instead of keypoint locations. We apply the same strategy and settings as in the part segmentation experiments, except that the model outputs a heatmap per class instead of a probability distribution,  and L2 loss instead of cross-entropy loss is used.  Similarly, we compare our approach to the Transfer-Learning baseline on Car and Bird. We evaluate the bird model on the CUB bird dataset,  while the car model on 20 manually-labeled real images since no previous car dataset have keypoints annotation as fine as ours. 

\vspace{-3mm}
\paragraph{Results:}
\vspace{-1mm}
Performance evaluation is reported on the test set in Table~\ref{tbl:Keypoints_result}, with qualitative results in Fig.~\ref{fig:vis_kp}. 
Results demonstrate that our approach significantly outperforms the fine-tuning baseline using the same annotation budget.

\newcommand\hhh{1.18cm}
\newcommand\www{1.42cm}
\begin{figure*}[t!]
\vspace{-0mm}
\addtolength{\tabcolsep}{-5.5pt}
\begin{tabular}{cccccccccccc}
\includegraphics[align=c,height=\hhh,width=\www, trim=10 20 0 20,clip]{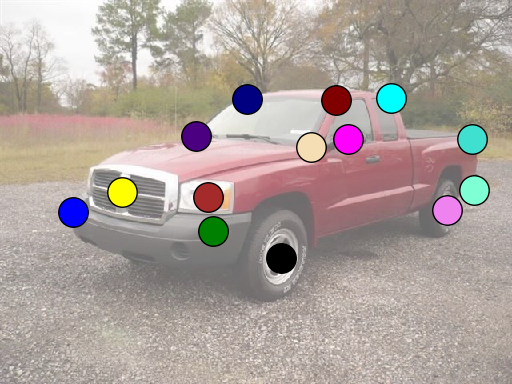} & \includegraphics[align=c,height=\hhh,width=\www, trim=10 20 0 20,clip]{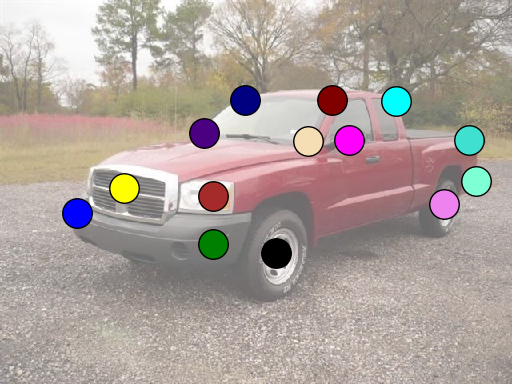} &
\includegraphics[align=c,height=\hhh,width=\www, trim=20 0 0 0,clip]{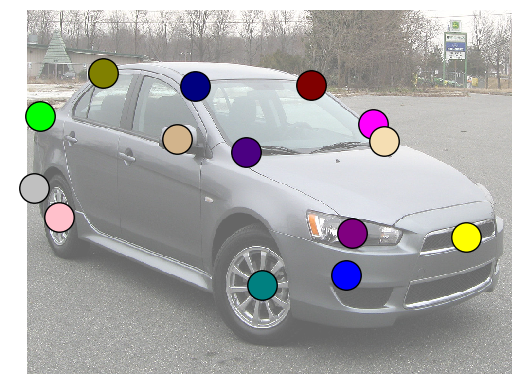} & \includegraphics[align=c,height=\hhh,width=\www, trim=25 0 0 0,clip]{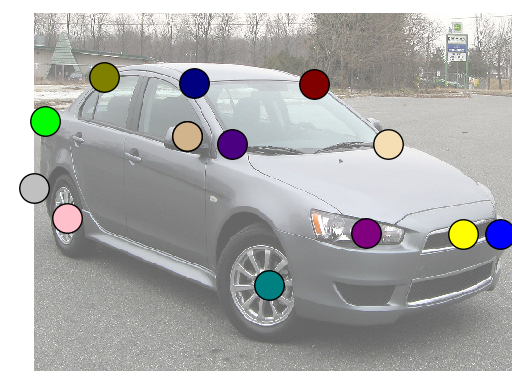} &
\includegraphics[align=c,height=\hhh,width=\www, trim=0 20 0 10,clip]{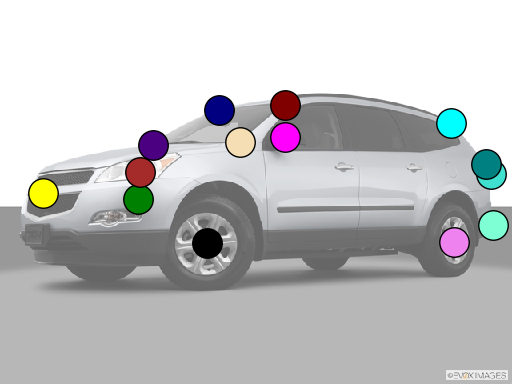} & \includegraphics[align=c,height=\hhh,width=\www, trim=0 20 0 10,clip]{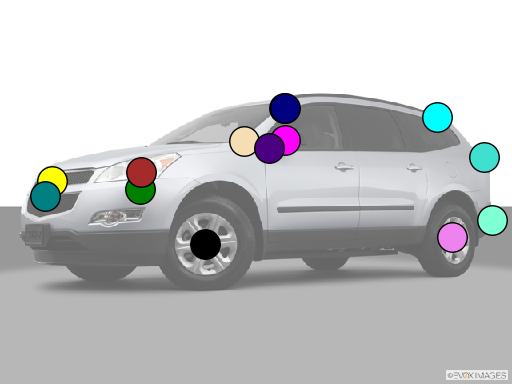} &
\includegraphics[align=c,height=\hhh,width=\www, trim=0 0 0 20,clip]{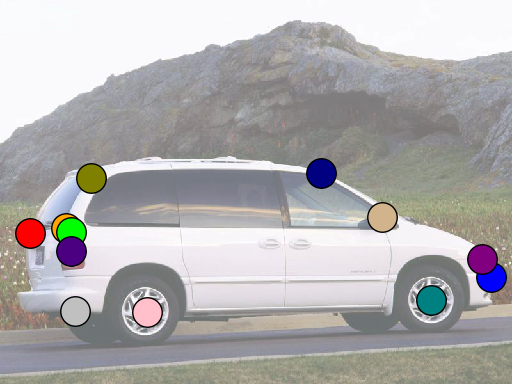} & \includegraphics[align=c,height=\hhh,width=\www, trim=0 0 0 20,clip]{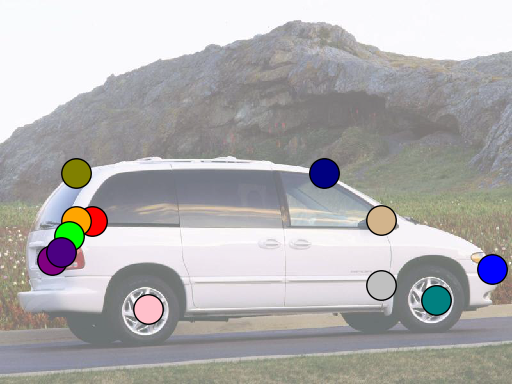} &
\includegraphics[align=c,height=\hhh,width=\www, trim=0 0 0 20,clip]{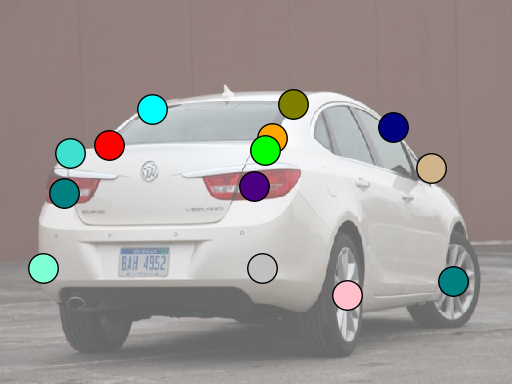} & \includegraphics[align=c,height=\hhh,width=\www, trim=0 0 0 20,clip]{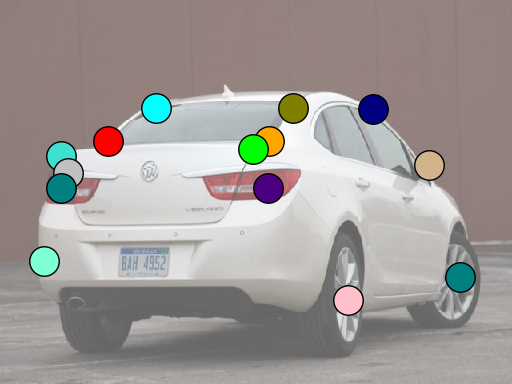} &
\includegraphics[align=c,height=\hhh,width=\www, trim=0 10 0 20,clip]{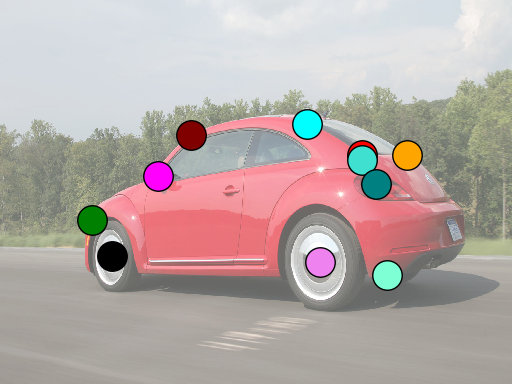} & \includegraphics[align=c,height=\hhh,width=\www, trim=0 10 0 20,clip]{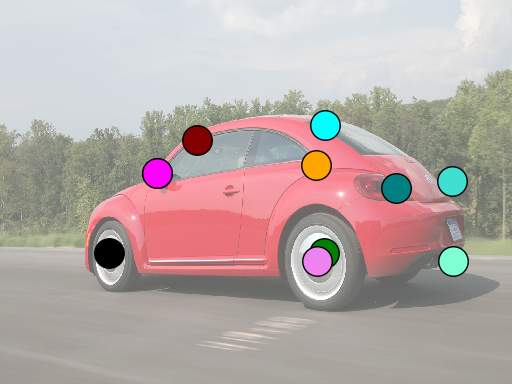} \\
\includegraphics[align=c,height=\hhh,width=\www, trim=0 0 0 0,clip]{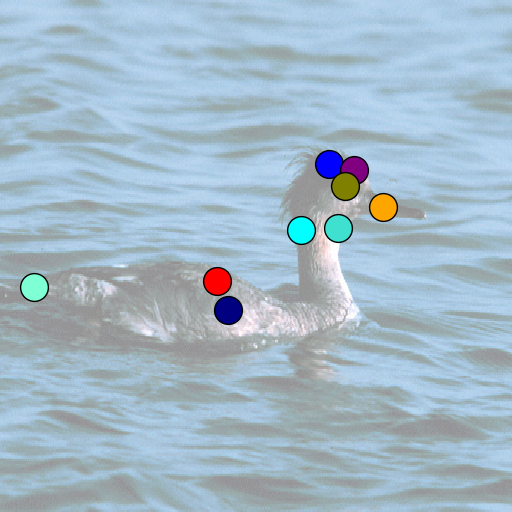} & \includegraphics[align=c,height=\hhh,width=\www, trim=0 0 0 0,clip]{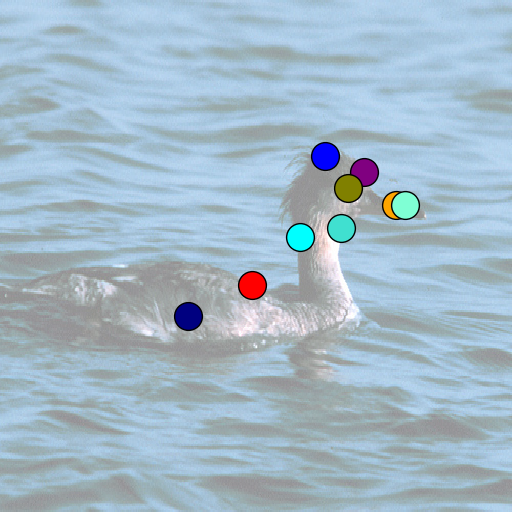} &
\includegraphics[align=c,height=\hhh,width=\www, trim=30 40 30 40,clip]{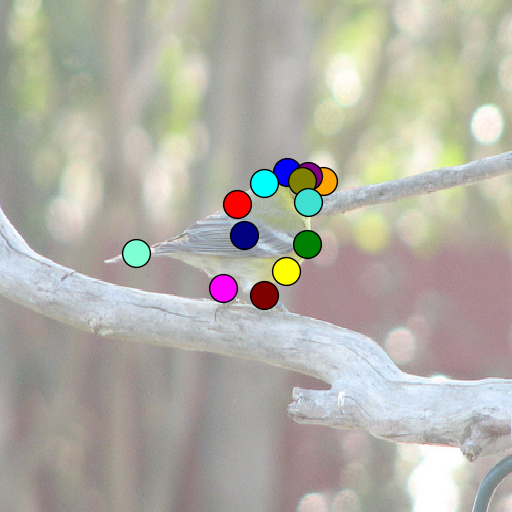} & \includegraphics[align=c,height=\hhh,width=\www, trim=30 40 30 40,clip]{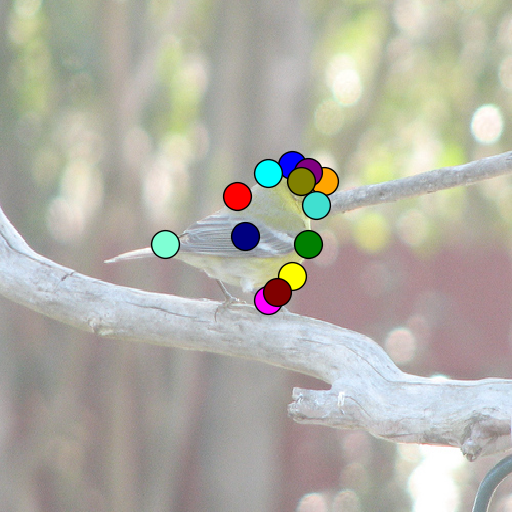} &
\includegraphics[align=c,height=\hhh,width=\www, trim=0 0 0 0,clip]{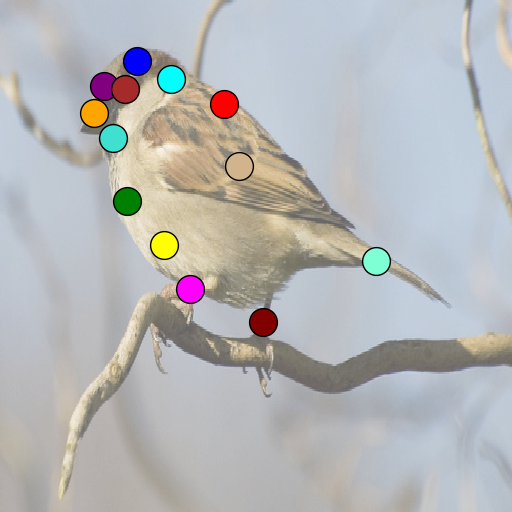} & \includegraphics[align=c,height=\hhh,width=\www, trim=0 0 0 0,clip]{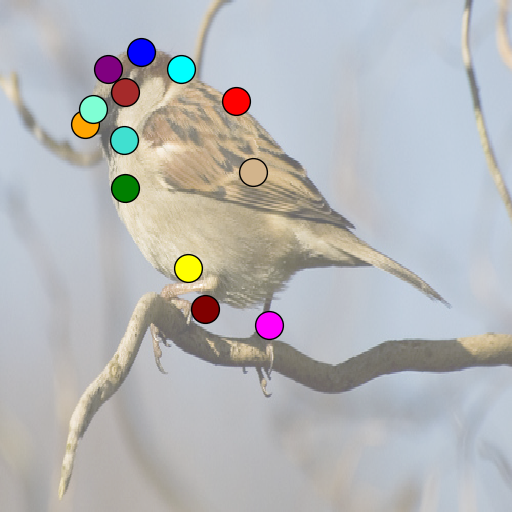} &
\includegraphics[align=c,height=\hhh,width=\www, trim=0 0 0 0,clip]{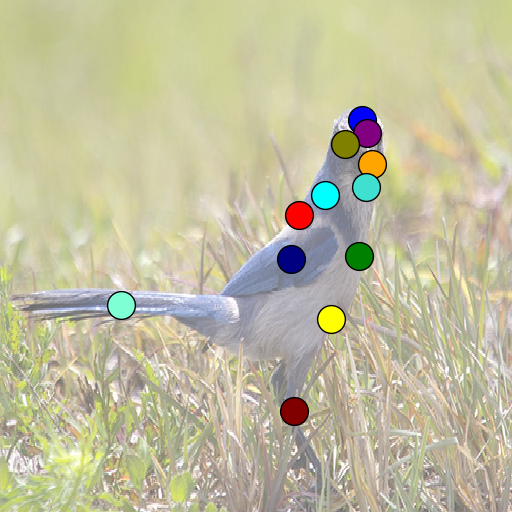} & \includegraphics[align=c,height=\hhh,width=\www, trim=0 0 0 0,clip]{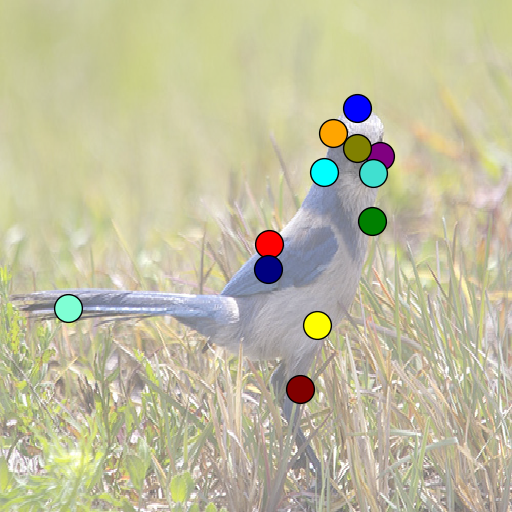} &
\includegraphics[align=c,height=\hhh,width=\www, trim=0 0 0 0,clip]{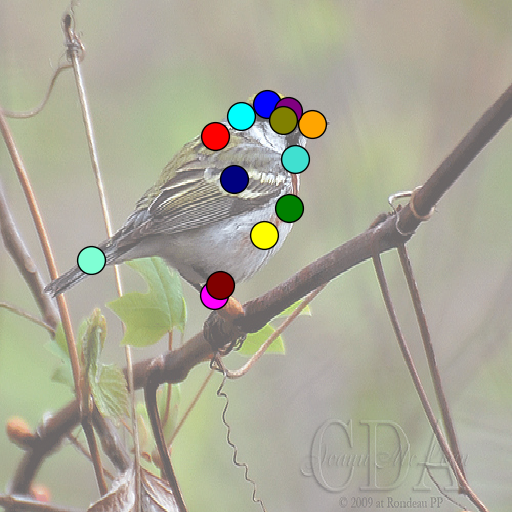} & \includegraphics[align=c,height=\hhh,width=\www, trim=0 0 0 0,clip]{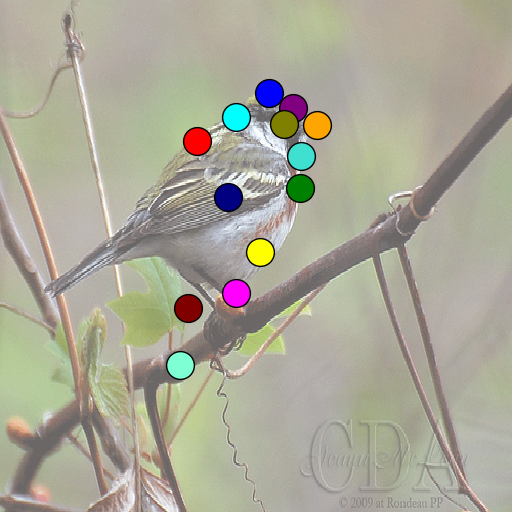} &
\includegraphics[align=c,height=\hhh,width=\www, trim=0 0 0 0,clip]{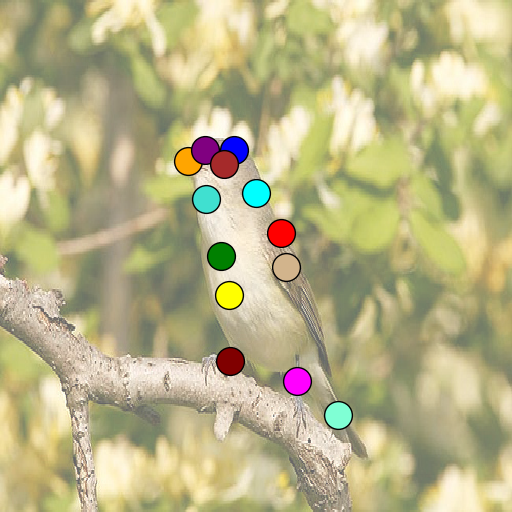} & \includegraphics[align=c,height=\hhh,width=\www, trim=0 0 0 0,clip]{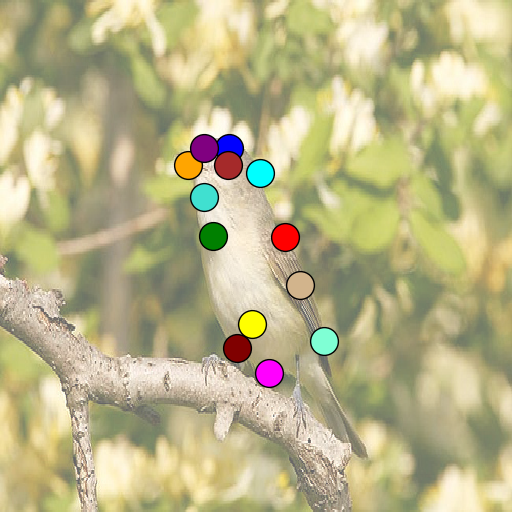} \\[-0.7mm]
{\scriptsize groundtruth}  & {\scriptsize prediction} & {\scriptsize groundtruth}  & {\scriptsize prediction} & {\scriptsize groundtruth}  & {\scriptsize prediction} & {\scriptsize groundtruth}  & {\scriptsize prediction}& {\scriptsize groundtruth}  & {\scriptsize prediction}& {\scriptsize groundtruth}  & {\scriptsize prediction}\\ 
\end{tabular}
\vspace{-4mm}
\caption{\footnotesize {\bf Qualitative results for Keypoint Detection}. \textbf{First row:} Model trained on the generated dataset using 30 human-provided annotations. Results shown are on CUB-Bird test set. \textbf{Second row:} Here 16 human-provided annotations are used.  Results shown are on Car-20 test set. }
\label{fig:vis_kp}
\vspace{-1.5mm}
\end{figure*}

\newcommand\hn{1.53cm}
\begin{figure*}[t!]
\vspace{-0.5mm}
\centering
\begin{minipage}[t!] {\textwidth}
\centering
\includegraphics[height=\hn]{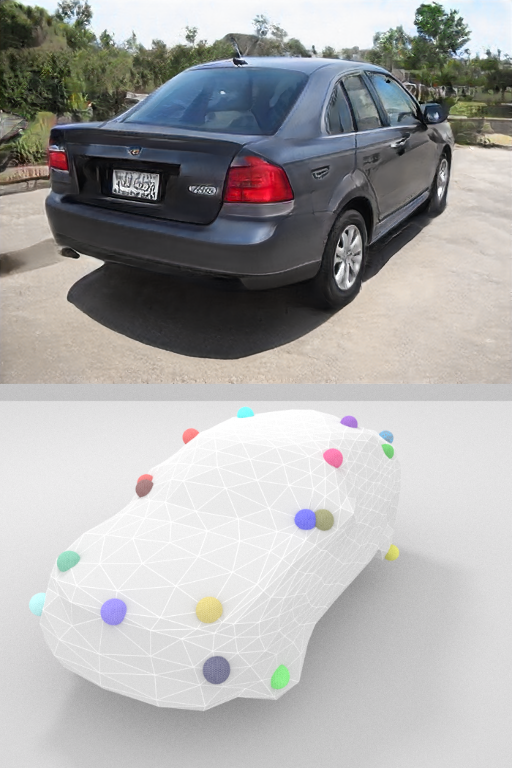}\includegraphics[height=\hn, trim=0 0 650 0,clip]{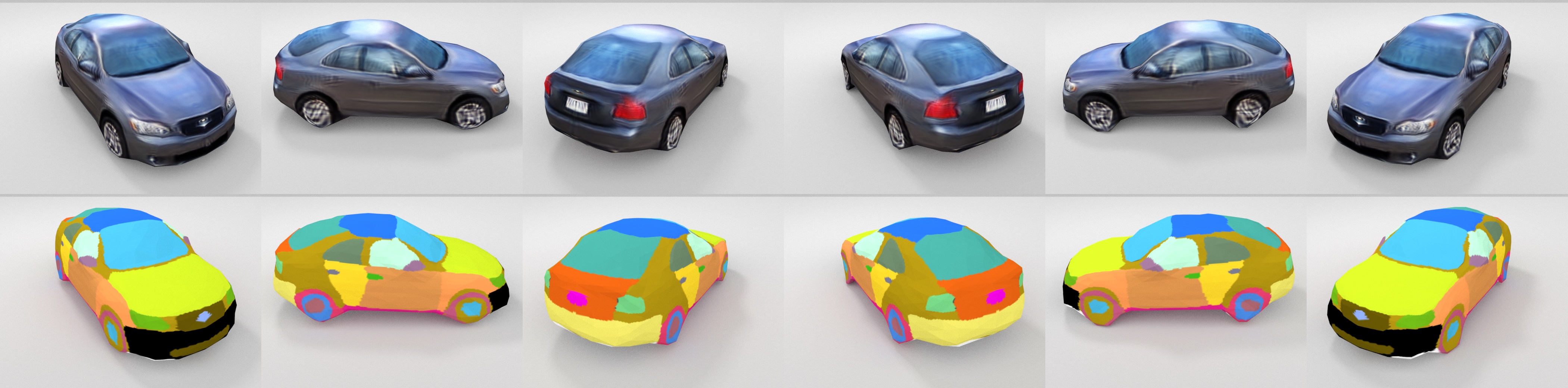}\includegraphics[height=\hn]{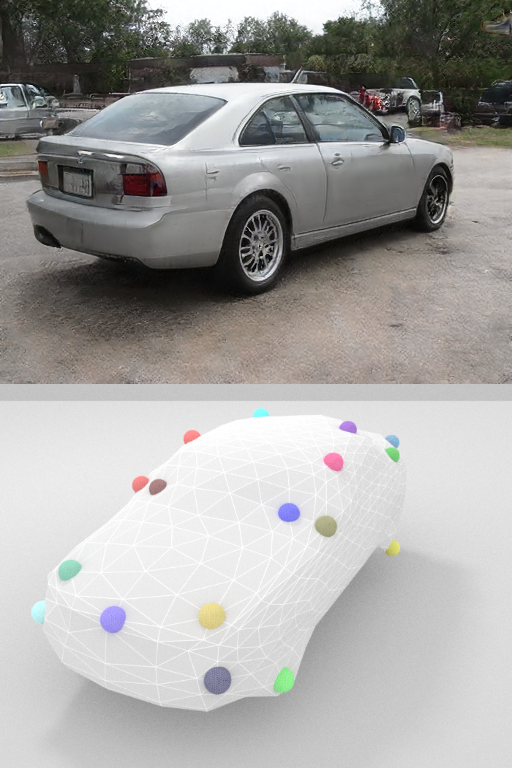}\includegraphics[height=\hn, trim=0 0 650 0,clip]{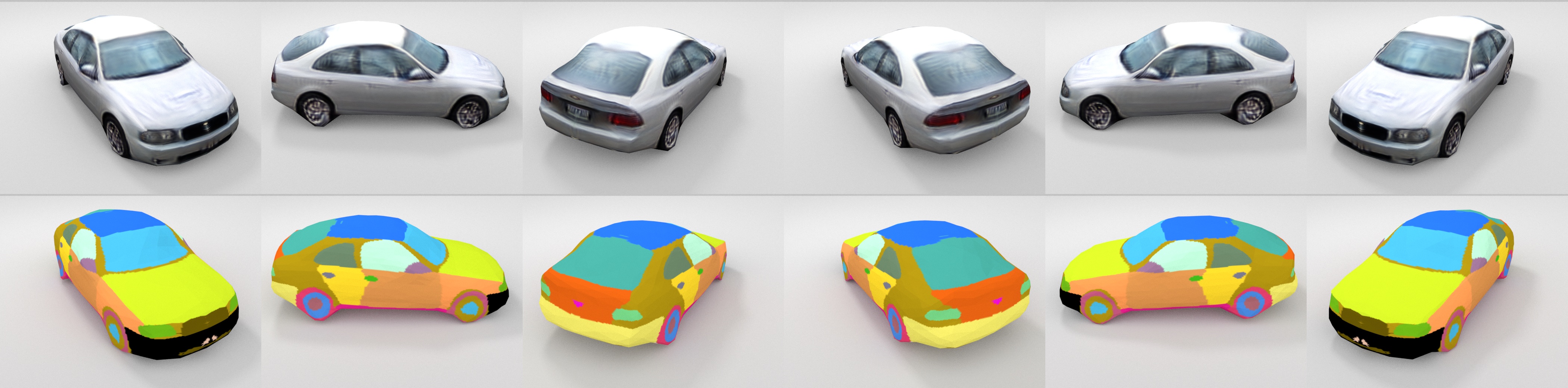}\includegraphics[height=\hn]{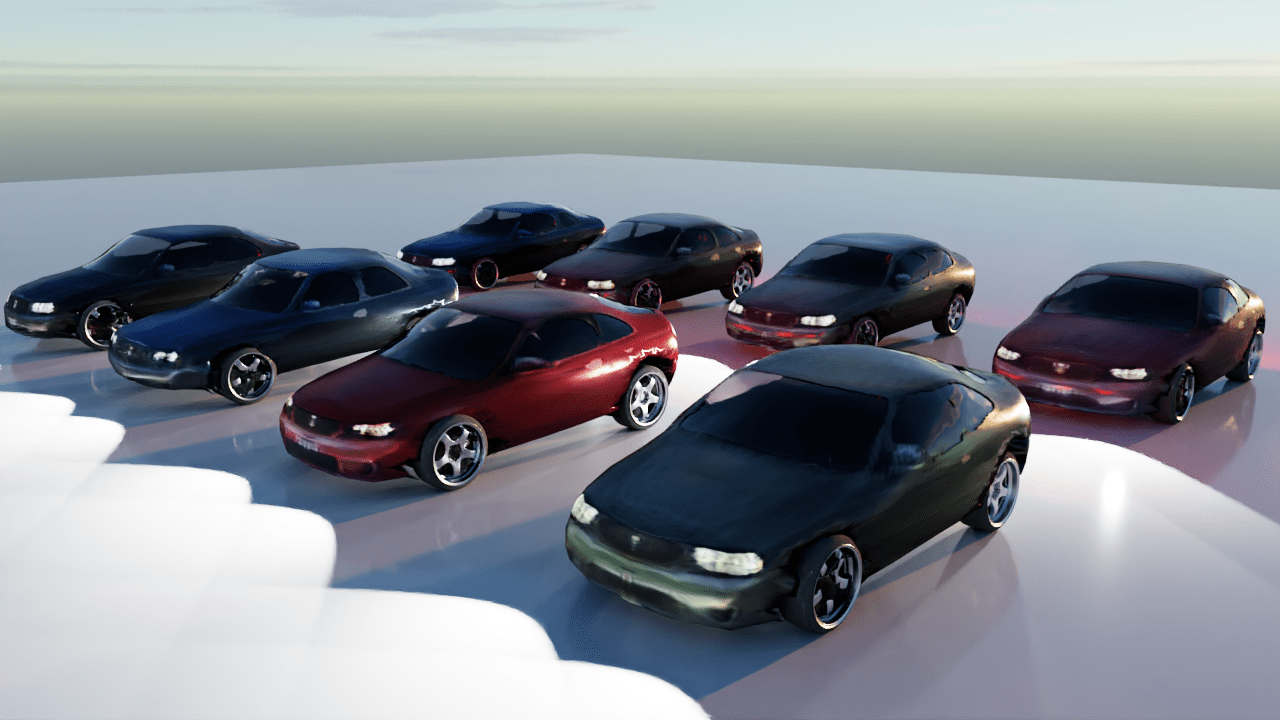}\includegraphics[height=\hn, trim=0 0 200 0,clip]{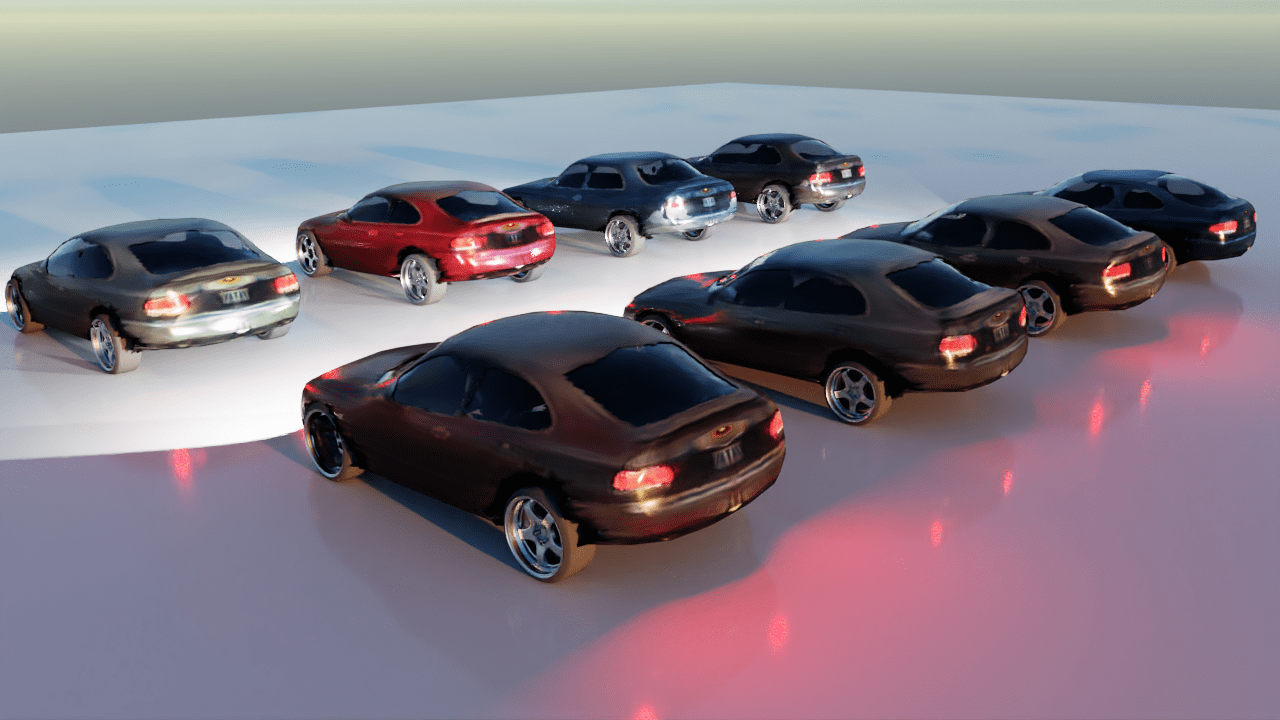}
\end{minipage}

\vspace{-2.5mm}
\caption{\footnotesize {\bf 3D Application:} We showcase our detailed part segmentation and keypoint detection in reconstructing animatable 3D objects from monocular images. We follow~\cite{zhang2020image} for training the inverse graphics network, but augment it with 3D part segmentation and 3D keypoint branches, which we supervise with 2D losses. Top left corner shows the input image, keypoint prediction is in the bottom, followed by a rendering of the predicted textured \& segmented 3D model into several views. We show an animated scene with lit front and back lights in the last column, which is made possible due to our 3D part segmentation. Cars have physics, rigged wheels, and can be driven virtually. See Supplementary for a video.}
\label{fig:3d}
\vspace{-2.0mm}
\end{figure*}

\vspace{-2mm}
\subsection{3D Application: Animatable 3D Assets}
\label{sec:3d}
\vspace{-2mm}
We now showcase how detailed part and keypoint prediction tasks can be leveraged in one downstream application. In particular, we aim to perform 3D reconstruction from single images (inverse rendering) to get rich 3D assets that can be animated realistically and potentially used in 3D games. This result is the first result of its kind. 

We focus on cars.  We aim to utilize the predicted keypoints as a way to estimate better 3D shape from monocular images. We further aim to map part segmentation to the estimated 3D model, which can then be used for post-processing: 1) placing correct materials for each part such as transparent windshields, 2) creating emissive lighting, and 3) replacing wheels with rigged wheels from an asset store, to enable the estimated 3D cars to drive realistically. 

We follow a similar pipeline as in~\cite{zhang2020image} to predict 3D shapes, texture, but also predict 3D parts and 3D keypoints. In particular, we first use StyleGAN to generate multiview images for different content codes. We then use our Style Interpreter to generate part  and keypoint labels. We train an inverse graphics network that accepts an image as input and predicts 3D shape, texture, 3D part labeling and 3D keypoints, by utilizing differentiable rendering~\cite{dib-r}. 
For 3D parts, we predict a part map, and paste it onto 3D shape (deformed sphere) in the same manner as for texture. For 3D keypoints, we learn a probability distribution over all vertices in the deformed shape.  We utilize all losses from~\cite{dib-r,zhang2020image}, and add an L2 loss on the projected keypoints, and Cross Entropy loss on the projected part segmentation. Details are in the Appendix. 

\vspace{-4mm}
\paragraph{Results:}
\vspace{-1.5mm}
We provide qualitative results in Fig.~\ref{fig:3d}, with additional results in Appendix, by highlighting the predicted part segmentation and animatable 3D assets. 

%% file: doc/conc.tex
\vspace{-1mm}
\section{Conclusions}
\vspace{-1mm}

We proposed a simple but powerful approach for semi-supervised learning with few labels. We exploited the learned latent space of the state-of-the-art generative model StyleGAN, and showed that an effective classifier can be trained on top from only a few human-annotated images. 
We manually label tiny datasets corresponding to 7 different tasks, each to a high detail. Training on these, our {\ours} synthesizes large labeled datasets on which computer vision architectures can be trained. Our approach is shown to outperform all semi-supervised baselines significantly, in some cases surpassing fully supervised approaches trained with two orders of magnitude more data. 
We believe this is only the first step towards more effective training of deep networks. In the future, we plan to extend DatasetGAN to handle a large and diverse set of classes.

%% file: main.bbl
\begin{thebibliography}{10}

\bibitem{bachman2019learning}
Philip Bachman, R~Devon Hjelm, and William Buchwalter.
\newblock Learning representations by maximizing mutual information across
  views.
\newblock In {\em Advances in Neural Information Processing Systems}, pages
  15535--15545, 2019.

\bibitem{8579074}
W.~H. {Beluch}, T.~{Genewein}, A.~{Nurnberger}, and J.~M. {Kohler}.
\newblock The power of ensembles for active learning in image classification.
\newblock In {\em 2018 IEEE/CVF Conference on Computer Vision and Pattern
  Recognition}, pages 9368--9377, 2018.

\bibitem{berthelot2019mixmatch}
David Berthelot, Nicholas Carlini, Ian Goodfellow, Nicolas Papernot, Avital
  Oliver, and Colin Raffel.
\newblock Mixmatch: A holistic approach to semi-supervised learning.
\newblock In {\em NeurIPS}, 2019.

\bibitem{brock2018large}
Andrew Brock, Jeff Donahue, and Karen Simonyan.
\newblock Large scale gan training for high fidelity natural image synthesis.
\newblock In {\em International Conference on Learning Representations}, 2018.

\bibitem{bucher2017generating}
Maxime Bucher, St{\'e}phane Herbin, and Fr{\'e}d{\'e}ric Jurie.
\newblock Generating visual representations for zero-shot classification.
\newblock In {\em Proceedings of the IEEE International Conference on Computer
  Vision Workshops}, pages 2666--2673, 2017.

\bibitem{chen2017deeplab}
Liang-Chieh Chen, George Papandreou, Iasonas Kokkinos, Kevin Murphy, and Alan~L
  Yuille.
\newblock Deeplab: Semantic image segmentation with deep convolutional nets,
  atrous convolution, and fully connected crfs.
\newblock {\em IEEE trans. on pattern analysis and machine intelligence},
  40(4):834--848, 2017.

\bibitem{chen2020simple}
Ting Chen, Simon Kornblith, Mohammad Norouzi, and Geoffrey Hinton.
\newblock A simple framework for contrastive learning of visual
  representations.
\newblock {\em arXiv preprint arXiv:2002.05709}, 2020.

\bibitem{chen2020big}
Ting Chen, Simon Kornblith, Kevin Swersky, Mohammad Norouzi, and Geoffrey~E
  Hinton.
\newblock Big self-supervised models are strong semi-supervised learners.
\newblock {\em Advances in Neural Information Processing Systems}, 33, 2020.

\bibitem{dib-r}
Wenzheng Chen, Jun Gao, Huan Ling, Edward Smith, Jaakko Lehtinen, Alec
  Jacobson, and Sanja Fidler.
\newblock Learning to predict 3d objects with an interpolation-based
  differentiable renderer.
\newblock In {\em Advances In Neural Information Processing Systems}, 2019.

\bibitem{choi2019self}
Jaehoon Choi, Taekyung Kim, and Changick Kim.
\newblock Self-ensembling with gan-based data augmentation for domain
  adaptation in semantic segmentation.
\newblock In {\em ICCV}, pages 6830--6840, 2019.

\bibitem{cordts2016cityscapes}
Marius Cordts, Mohamed Omran, Sebastian Ramos, Timo Rehfeld, Markus Enzweiler,
  Rodrigo Benenson, Uwe Franke, Stefan Roth, and Bernt Schiele.
\newblock The cityscapes dataset for semantic urban scene understanding.
\newblock In {\em CVPR}, pages 3213--3223, 2016.

\bibitem{metasim20}
Jeevan Devaranjan, Amlan Kar, and Sanja Fidler.
\newblock Meta-sim2: Unsupervised learning of scene structure for synthetic
  data generation.
\newblock In {\em ECCV}, 2020.

\bibitem{Everingham10}
M.~Everingham, L.~Van~Gool, C.~K.~I. Williams, J.~Winn, and A.~Zisserman.
\newblock The pascal visual object classes (voc) challenge.
\newblock {\em International Journal of Computer Vision}, 88(2):303--338, June
  2010.

\bibitem{felix2018multi}
Rafael Felix, Vijay~BG Kumar, Ian Reid, and Gustavo Carneiro.
\newblock Multi-modal cycle-consistent generalized zero-shot learning.
\newblock In {\em Proceedings of the European Conference on Computer Vision
  (ECCV)}, pages 21--37, 2018.

\bibitem{galeev2020learning}
Danil Galeev, Konstantin Sofiiuk, Danila Rukhovich, Mikhail Romanov, Olga
  Barinova, and Anton Konushin.
\newblock Learning high-resolution domain-specific representations with a gan
  generator.
\newblock {\em arXiv preprint arXiv:2006.10451}, 2020.

\bibitem{goodfellow2014generative}
Ian Goodfellow, Jean Pouget-Abadie, Mehdi Mirza, Bing Xu, David Warde-Farley,
  Sherjil Ozair, Aaron Courville, and Yoshua Bengio.
\newblock Generative adversarial nets.
\newblock In {\em NeurIPS}, pages 2672--2680, 2014.

\bibitem{grill2020bootstrap}
Jean-Bastien Grill, Florian Strub, Florent Altch{\'e}, Corentin Tallec, Pierre
  Richemond, Elena Buchatskaya, Carl Doersch, Bernardo Avila~Pires, Zhaohan
  Guo, Mohammad Gheshlaghi~Azar, et~al.
\newblock Bootstrap your own latent-a new approach to self-supervised learning.
\newblock {\em Advances in Neural Information Processing Systems}, 33, 2020.

\bibitem{hadsell2006dimensionality}
Raia Hadsell, Sumit Chopra, and Yann LeCun.
\newblock Dimensionality reduction by learning an invariant mapping.
\newblock In {\em 2006 IEEE Computer Society Conference on Computer Vision and
  Pattern Recognition (CVPR'06)}, volume~2, pages 1735--1742. IEEE, 2006.

\bibitem{he2020momentum}
Kaiming He, Haoqi Fan, Yuxin Wu, Saining Xie, and Ross Girshick.
\newblock Momentum contrast for unsupervised visual representation learning.
\newblock In {\em Proceedings of the IEEE/CVF Conference on Computer Vision and
  Pattern Recognition}, pages 9729--9738, 2020.

\bibitem{he2016deep}
Kaiming He, Xiangyu Zhang, Shaoqing Ren, and Jian Sun.
\newblock Deep residual learning for image recognition.
\newblock In {\em Proceedings of the IEEE conference on computer vision and
  pattern recognition}, pages 770--778, 2016.

\bibitem{Hinton2007ToRS}
Geoffrey~E. Hinton.
\newblock To recognize shapes, first learn to generate images.
\newblock {\em Progress in brain research}, 165:535--47, 2007.

\bibitem{huang2017arbitrary}
Xun Huang and Serge Belongie.
\newblock Arbitrary style transfer in real-time with adaptive instance
  normalization.
\newblock In {\em Proceedings of the IEEE International Conference on Computer
  Vision}, pages 1501--1510, 2017.

\bibitem{hung2018adversarial}
Wei-Chih Hung, Yi-Hsuan Tsai, Yan-Ting Liou, Yen-Yu Lin, and Ming-Hsuan Yang.
\newblock Adversarial learning for semi-supervised semantic segmentation.
\newblock {\em arXiv preprint arXiv:1802.07934}, 2018.

\bibitem{ji2019invariant}
Xu~Ji, Jo{\~a}o~F Henriques, and Andrea Vedaldi.
\newblock Invariant information clustering for unsupervised image
  classification and segmentation.
\newblock In {\em Proceedings of the IEEE International Conference on Computer
  Vision}, pages 9865--9874, 2019.

\bibitem{Metasim19}
Amlan Kar, Aayush Prakash, Ming-Yu Liu, Eric Cameracci, Justin Yuan, Matt
  Rusiniak, David Acuna, Antonio Torralba, and Sanja Fidler.
\newblock Meta-sim: Learning to generate synthetic datasets.
\newblock In {\em ICCV}, 2019.

\bibitem{karras2017progressive}
Tero Karras, Timo Aila, Samuli Laine, and Jaakko Lehtinen.
\newblock Progressive growing of gans for improved quality, stability, and
  variation.
\newblock {\em arXiv preprint arXiv:1710.10196}, 2017.

\bibitem{stylegan}
Tero Karras, Samuli Laine, and Timo Aila.
\newblock A style-based generator architecture for generative adversarial
  networks.
\newblock In {\em CVPR}, 2019.

\bibitem{ke2020guided}
Zhanghan Ke, Di~Qiu, Kaican Li, Qiong Yan, and Rynson~WH Lau.
\newblock Guided collaborative training for pixel-wise semi-supervised
  learning.
\newblock {\em arXiv preprint arXiv:2008.05258}, 2020.

\bibitem{KrauseStarkDengFei-Fei_3DRR2013}
Jonathan Krause, Michael Stark, Jia Deng, and Li~Fei-Fei.
\newblock 3d object representations for fine-grained categorization.
\newblock In {\em 4th International IEEE Workshop on 3D Representation and
  Recognition (3dRR-13)}, Sydney, Australia, 2013.

\bibitem{Kuo2018CostSensitiveAL}
Weicheng Kuo, Christian H{\"a}ne, E.~Yuh, P.~Mukherjee, and Jitendra Malik.
\newblock Cost-sensitive active learning for intracranial hemorrhage detection.
\newblock In {\em MICCAI}, 2018.

\bibitem{fedsim20}
Daiqing Li, Amlan Kar, Nishant Ravikumar, Alejandro~F Frangi, and Sanja Fidler.
\newblock Federated simulation for medical imaging.
\newblock In {\em MICCAI}, 2020.

\bibitem{segGAN21}
Daiqing Li, Junlin Yang, Karsten Kreis, Antonio Torralba, and Sanja Fidler.
\newblock Semantic segmentation with generative models: Semi-supervised
  learning and strong out-of-domain generalization.
\newblock In {\em CVPR}, 2021.

\bibitem{Lin2014MicrosoftCC}
Tsung-Yi Lin, M.~Maire, Serge~J. Belongie, James Hays, P.~Perona, D.~Ramanan,
  Piotr Doll{\'a}r, and C.~L. Zitnick.
\newblock Microsoft coco: Common objects in context.
\newblock {\em ArXiv}, abs/1405.0312, 2014.

\bibitem{lin2014microsoft}
Tsung-Yi Lin, Michael Maire, Serge Belongie, James Hays, Pietro Perona, Deva
  Ramanan, Piotr Doll{\'a}r, and C~Lawrence Zitnick.
\newblock Microsoft coco: Common objects in context.
\newblock In {\em European conference on computer vision}, pages 740--755.
  Springer, 2014.

\bibitem{amodalVAE20}
Huan Ling, David Acuna, Karsten Kreis, Seung Kim, and Sanja Fidler.
\newblock Variational amodal object completion for interactive scene editing.
\newblock In {\em NeurIPS}, 2020.

\bibitem{liu2015faceattributes}
Ziwei Liu, Ping Luo, Xiaogang Wang, and Xiaoou Tang.
\newblock Deep learning face attributes in the wild.
\newblock In {\em Proceedings of International Conference on Computer Vision
  (ICCV)}, December 2015.

\bibitem{long2017zero}
Yang Long, Li~Liu, Fumin Shen, Ling Shao, and Xuelong Li.
\newblock Zero-shot learning using synthesised unseen visual data with
  diffusion regularisation.
\newblock {\em IEEE transactions on pattern analysis and machine intelligence},
  40(10):2498--2512, 2017.

\bibitem{luc2016semantic}
Pauline Luc, Camille Couprie, Soumith Chintala, and Jakob Verbeek.
\newblock Semantic segmentation using adversarial networks.
\newblock In {\em NIPS Workshop on Adversarial Training}, 2016.

\bibitem{JSAL}
Prem Melville, Stewart~M. Yang, Maytal Saar-Tsechansky, and Raymond Mooney.
\newblock Active learning for probability estimation using jensen-shannon
  divergence.
\newblock In Jo{\~a}o Gama, Rui Camacho, Pavel~B. Brazdil, Al{\'i}pio~M{\'a}rio
  Jorge, and Lu{\'i}s Torgo, editors, {\em Machine Learning: ECML 2005}, pages
  268--279, Berlin, Heidelberg, 2005. Springer Berlin Heidelberg.

\bibitem{misra2020self}
Ishan Misra and Laurens van~der Maaten.
\newblock Self-supervised learning of pretext-invariant representations.
\newblock In {\em Proceedings of the IEEE/CVF Conference on Computer Vision and
  Pattern Recognition}, pages 6707--6717, 2020.

\bibitem{mittal2019semi}
Sudhanshu Mittal, Maxim Tatarchenko, and Thomas Brox.
\newblock Semi-supervised semantic segmentation with high-and low-level
  consistency.
\newblock {\em IEEE Transactions on Pattern Analysis and Machine Intelligence},
  2019.

\bibitem{murez2018image}
Zak Murez, Soheil Kolouri, David Kriegman, Ravi Ramamoorthi, and Kyungnam Kim.
\newblock Image to image translation for domain adaptation.
\newblock In {\em Proceedings of the IEEE Conference on Computer Vision and
  Pattern Recognition}, pages 4500--4509, 2018.

\bibitem{oord2018representation}
Aaron van~den Oord, Yazhe Li, and Oriol Vinyals.
\newblock Representation learning with contrastive predictive coding.
\newblock {\em arXiv preprint arXiv:1807.03748}, 2018.

\bibitem{labelme}
Bryan Russell, Antonio Torralba, Kevin Murphy, and William Freeman.
\newblock Labelme: A database and web-based tool for image annotation.
\newblock {\em International Journal of Computer Vision}, 77(05), 2008.

\bibitem{sariyildiz2019gradient}
Mert~Bulent Sariyildiz and Ramazan~Gokberk Cinbis.
\newblock Gradient matching generative networks for zero-shot learning.
\newblock In {\em Proceedings of the IEEE Conference on Computer Vision and
  Pattern Recognition}, pages 2168--2178, 2019.

\bibitem{Sener2017AGA}
O.~Sener and S.~Savarese.
\newblock A geometric approach to active learning for convolutional neural
  networks.
\newblock {\em ArXiv}, abs/1708.00489, 2017.

\bibitem{sohn2020fixmatch}
Kihyuk Sohn, David Berthelot, Chun-Liang Li, Zizhao Zhang, Nicholas Carlini,
  Ekin~D. Cubuk, Alex Kurakin, Han Zhang, and Colin Raffel.
\newblock Fixmatch: Simplifying semi-supervised learning with consistency and
  confidence.
\newblock In {\em NeurIPS}, 2020.

\bibitem{souly2017semi}
Nasim Souly, Concetto Spampinato, and Mubarak Shah.
\newblock Semi supervised semantic segmentation using generative adversarial
  network.
\newblock In {\em Proceedings of the IEEE International Conference on Computer
  Vision}, pages 5688--5696, 2017.

\bibitem{tarvainen2017mean}
Antti Tarvainen and Harri Valpola.
\newblock Mean teachers are better role models: Weight-averaged consistency
  targets improve semi-supervised deep learning results.
\newblock In {\em Advances in neural information processing systems}, pages
  1195--1204, 2017.

\bibitem{tian2019contrastive}
Yonglong Tian, Dilip Krishnan, and Phillip Isola.
\newblock Contrastive multiview coding.
\newblock {\em arXiv preprint arXiv:1906.05849}, 2019.

\bibitem{7298658}
G.~{Van Horn}, S.~{Branson}, R.~{Farrell}, S.~{Haber}, J.~{Barry},
  P.~{Ipeirotis}, P.~{Perona}, and S.~{Belongie}.
\newblock Building a bird recognition app and large scale dataset with citizen
  scientists: The fine print in fine-grained dataset collection.
\newblock In {\em Conference on Computer Vision and Pattern Recognition
  (CVPR)}, pages 595--604, 2015.

\bibitem{vu2019advent}
Tuan-Hung Vu, Himalaya Jain, Maxime Bucher, Matthieu Cord, and Patrick
  P{\'e}rez.
\newblock Advent: Adversarial entropy minimization for domain adaptation in
  semantic segmentation.
\newblock In {\em Proceedings of the IEEE conference on computer vision and
  pattern recognition}, pages 2517--2526, 2019.

\bibitem{cub}
P.~Welinder, S.~Branson, T.~Mita, C.~Wah, F.~Schroff, S.~Belongie, and
  P.~Perona.
\newblock {Caltech-UCSD Birds 200}.
\newblock Technical Report CNS-TR-2010-001, California Institute of Technology,
  2010.

\bibitem{kaggle_cat}
Weiwei Zhang, Jian Sun, and Xiaoou Tang.
\newblock Cat head detection - how to effectively exploit shape and texture
  features.
\newblock In David Forsyth, Philip Torr, and Andrew Zisserman, editors, {\em
  Computer Vision -- ECCV 2008}, pages 802--816, Berlin, Heidelberg, 2008.
  Springer Berlin Heidelberg.

\bibitem{zhang2020image}
Yuxuan Zhang, Wenzheng Chen, Huan Ling, Jun Gao, Yinan Zhang, Antonio Torralba,
  and Sanja Fidler.
\newblock Image gans meet differentiable rendering for inverse graphics and
  interpretable 3d neural rendering.
\newblock {\em arXiv preprint arXiv:2010.09125}, 2020.

\bibitem{zhou2016semantic}
Bolei Zhou, Hang Zhao, Xavier Puig, Sanja Fidler, Adela Barriuso, and Antonio
  Torralba.
\newblock Semantic understanding of scenes through the ade20k dataset.
\newblock {\em arXiv preprint arXiv:1608.05442}, 2016.

\bibitem{zhou2017scene}
Bolei Zhou, Hang Zhao, Xavier Puig, Sanja Fidler, Adela Barriuso, and Antonio
  Torralba.
\newblock Scene parsing through ade20k dataset.
\newblock In {\em Proceedings of the IEEE Conference on Computer Vision and
  Pattern Recognition}, 2017.

\bibitem{zou2018unsupervised}
Yang Zou, Zhiding Yu, BVK Vijaya~Kumar, and Jinsong Wang.
\newblock Unsupervised domain adaptation for semantic segmentation via
  class-balanced self-training.
\newblock In {\em Proceedings of the European conference on computer vision
  (ECCV)}, pages 289--305, 2018.

\end{thebibliography}
